\documentclass{article}

\makeatletter
\def\input@path{{./}}
\makeatother
\usepackage{iclr2027_conference,times}
\usepackage{array}
\usepackage{enumitem}
\usepackage{amsmath,amssymb}
\usepackage{graphicx}
\newcommand{\bodyCI}[3]{{#1}^{\scalebox{0.8}{$\scriptstyle #2$}}_{\scalebox{0.8}{$\scriptstyle #3$}}}
\graphicspath{{./}}
\usepackage{booktabs}
\usepackage{placeins}
\usepackage{float}
\usepackage{flafter}
\usepackage{xcolor}
\usepackage[hidelinks]{hyperref}
\usepackage{url}
\usepackage{tikz}
\usepackage{titletoc}
\usetikzlibrary{arrows.meta,positioning,shapes.geometric,fit}

\definecolor{armU}{HTML}{0077BB}
\definecolor{armS}{HTML}{AA3377}
\definecolor{armT}{HTML}{EE7733}

\newcommand{\pihalf}{$\pi_{0.5}$}

\newcommand{\MERGE}{\textsc{Merge}}

\newcommand{\pp}{\,\mathrm{pp}}

\title{Beyond Success Rate: Interventional Evaluation of Memory in Vision--Language--Action Systems}

\author{%
\begingroup
\normalfont\fontsize{9}{10}\selectfont
\renewcommand{\arraystretch}{1}
\begin{tabular}{@{}c@{}}
{\bfseries\fontsize{9}{10}\selectfont Keyuan Luo$^{1}$ \quad Yi Xie$^{2}$ \quad Wanyun Ling$^{2}$ \quad
Yuanzhe Li$^{3}$ \quad Ziyue Li$^{2,\dagger}$ \quad Zhuoqi Zeng$^{1,*}$}\\[2pt]
$^{1}$Hainan Bielefeld University of Applied Sciences (BiUH)\\
$^{2}$Technical University of Munich \quad $^{3}$Technische Universit\"at Berlin\\[2pt]
{\fontsize{8}{9.5}\selectfont\ttfamily
\href{mailto:keyuan.luo@hibiuh.edu.cn}{keyuan.luo@hibiuh.edu.cn} \quad
\href{mailto:aaron.xie@tum.de}{aaron.xie@tum.de} \quad
\href{mailto:wanyun.ling@tum.de}{wanyun.ling@tum.de}}\\
{\fontsize{8}{9.5}\selectfont\ttfamily
\href{mailto:yuanzhe.li@campus.tu-berlin.de}{yuanzhe.li@campus.tu-berlin.de} \quad
\href{mailto:ziyue.li@tum.de}{ziyue.li@tum.de} \quad
\href{mailto:zhuoqi.zeng@hibiuh.edu.cn}{zhuoqi.zeng@hibiuh.edu.cn}}\\[1pt]
$^{*}$Primary corresponding author \quad $^{\dagger}$Secondary corresponding author
\end{tabular}
\endgroup}

\makeatletter
\renewcommand{\@maketitle}{\vbox{\hsize\textwidth\parskip=0pt
  {\LARGE\scshape\@title\par}
  \vskip 4pt
  {\centering\@author\par}
  \vskip 0pt}}
\makeatother

\iclrfinalcopy
\hypersetup{
  unicode=true,
  pdftitle={Beyond Success Rate: Interventional Evaluation of Memory in Vision-Language-Action Systems},
  pdfauthor={Keyuan Luo, Yi Xie, Wanyun Ling, Yuanzhe Li, Ziyue Li, Zhuoqi Zeng}
}

\begin{document}
\maketitle
\lhead{Preprint}
\raggedbottom

\begin{abstract}
When we say that a robot uses memory, does that memory actually play a role?
Success-rate ablations alone cannot establish the use of task-relevant facts: they
change information and computation together, while aggregate rates can hide
offsetting gains and losses. We propose an Interventional Memory Evaluation
Protocol (IMEP) for vision--language--action (VLA) systems to address four
questions: (1) Does memory provide a net benefit? (2) Under what conditions does
memory help? (3) Do changes to memory alter task outcomes? (4) Are these effects
specific to task-relevant facts? At the retrieval interface, we
compare target-fact substitutions with fact-preserving replacements, minimal
numerical perturbations, and exact replicas, while assessing decision correctness
separately from motor execution success. We evaluate within-episode,
cross-episode, and agentic memory. In cross-episode learned-bank experiments,
a one-unit-in-the-last-place (ULP) intervention, including its installation
procedure, changes 11.9\% of seed-paired outcomes, while changing key facts produces no clear additional effect
over fact-preserving replacements. To examine when memory helps, we introduce
ChainBench, where correct decisions require historical information unavailable in
current observations. The full memory configuration raises task
success from 39.1\% to 74.3\%, matching the oracle-decision condition under
shared recorded executions. These results show that correct history can support
robot decisions, but establishing whether a memory system realizes this benefit
requires fact-specific evidence beyond aggregate success rates.
\end{abstract}

\section{Introduction}
\label{sec:intro}

\begin{figure}[t]
\centering
\includegraphics[width=\textwidth]{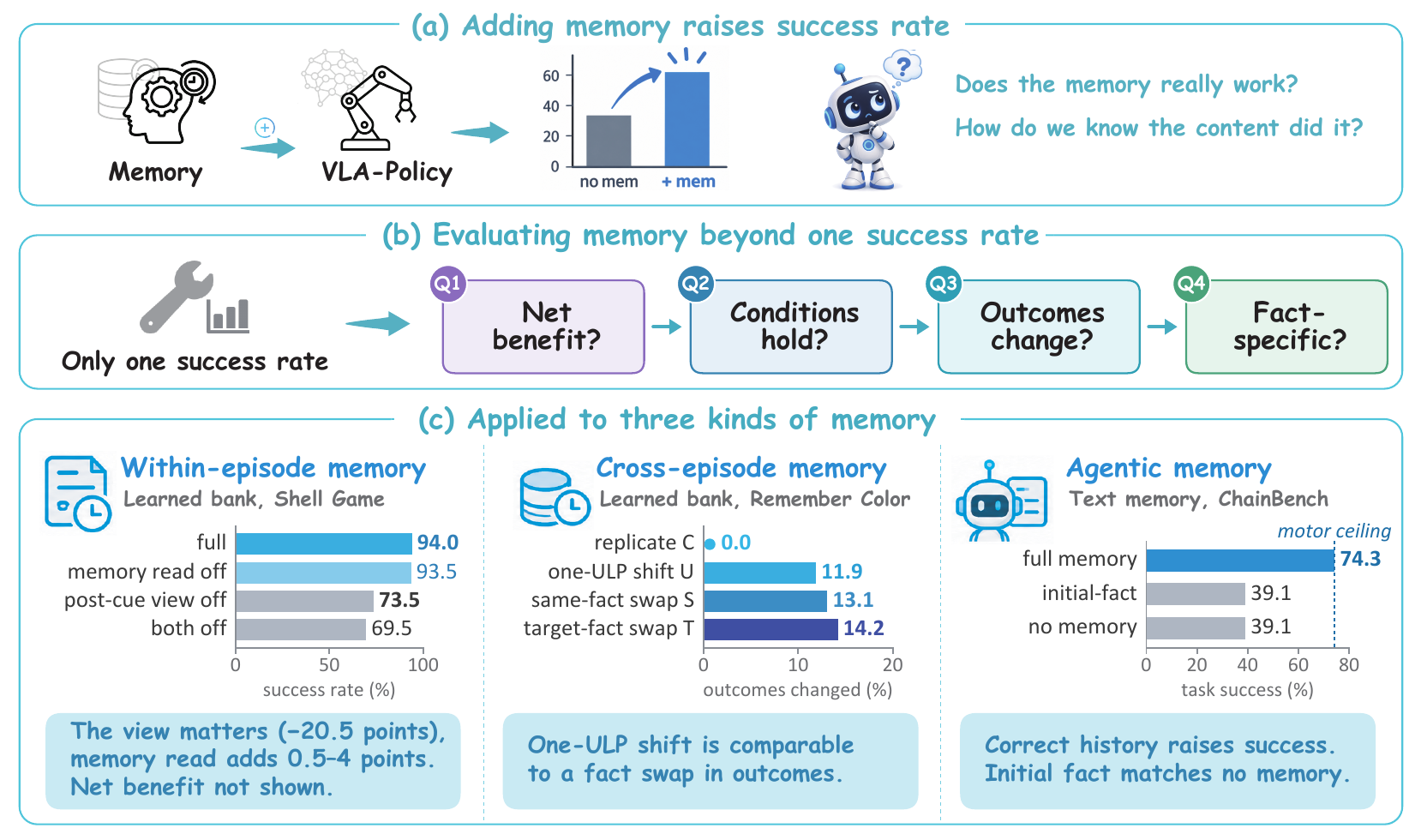}
\caption{\textbf{Success rate alone cannot show that memory content mattered.}
(a) Higher task success alone does not establish use of stored content.
(b) IMEP separates four questions. (c) Shell Game ($200$ pairs per contrast),
Remember Color ($388$ pairs), and ChainBench
($202$ chains). Shell Game's memory-read gain has an upper bound below the
registered 10-pp criterion; Remember Color operates near chance. ChainBench
uses no decision-time LLM; shared recorded executions yield equal success
when full memory and the oracle choose correctly. The initial-fact baseline
ignores updates. Equal $U/T$ rates and initial-fact/no-memory equivalence are
not established (Sections~\ref{sec:within}--\ref{sec:agentic}).}
\label{fig:teaser}
\end{figure}

A robot's current observation may omit information needed to act: an earlier
visual cue, a previous episode's experience, or a fact retained across tasks.
Vision--language--action (VLA) systems therefore augment policies with memory
banks, event keyframes, or temporal representations
\citep{memoryvla,kemo,hamlet}, experience retrieval \citep{memer}, and external
text records \citep{harnessvla}. These mechanisms aim to turn history into
useful evidence for current decisions. Their value is often evaluated through
memory ablations and task success \citep{memoryvla,kemo}. Such gains show
improved performance, but do they establish use of task-relevant stored facts
(Figure~\ref{fig:teaser}a)?

Three distinctions prevent that inference. First, enabling memory can change
both information and computation. Second, an unchanged success rate can hide
gains on some trials and losses on others \citep{predictionchurn}. Third, changing memory can alter
behavior even when a replacement preserves the target fact or introduces only
a small numerical perturbation. History can also support shortcuts, as shown
by causal confusion and copycat behavior in imitation learning
\citep{causalconfusion,copycat}. Behavioral sensitivity alone does not
establish use of task-relevant facts. Conversely, an undetected net gain does not imply
that memory has no effect. Interpretation also depends on whether current
observations already reveal the answer, whether an intervention reaches the
reader, and whether stored facts remain valid.

We introduce the \textbf{Interventional Memory Evaluation Protocol (IMEP)} to
separate net benefit (Q1), conditions for memory to help (Q2), outcome
sensitivity (Q3), and fact specificity (Q4). At~a~fixed retrieval interface,
IMEP compares replacements that change the target fact with those that preserve
it. On matched units, it records outcome flips alongside aggregate success
and scores decisions separately from execution where observable.
Section~\ref{sec:method} defines the protocol; Section~\ref{sec:related-main}
discusses related work.

In the evaluated learned banks, memory interventions change outcomes without
establishing fact specificity. Remember Color's one-unit-in-the-last-place
(ULP) control, including installation, changes $11.9\%$ of paired outcomes;
different-fact replacements, as delivered, produce no clear additional effect
over same-fact replacements. No detected learned-bank gain meets a registered net-benefit
criterion. On ChainBench, full memory achieves perfect decision accuracy on
202 chains and raises task success from $39.1\%$ to $74.3\%$, matching the
oracle under shared recorded executions. A separate
experiment supports fact specificity for a scripted reader on curated records.

Our contributions are:
\begin{itemize}[nosep]
\item \textbf{IMEP}, a four-question protocol using fact-changing and
fact-preserving substitutions, exact repeats, and one-ULP controls, with
separate decision and execution readouts.
\item \textbf{ChainBench}, a LIBERO benchmark of multi-episode chains requiring
historical facts absent from current observations, with decisions scored
separately from motor execution.
\item \textbf{An evaluation of within-episode, cross-episode and agentic
memory}, finding outcome sensitivity without established fact specificity in
learned banks, and task success at the motor-execution ceiling for the evaluated
ChainBench configuration under shared recorded executions.
\end{itemize}

\section{Related Work}
\label{sec:related-main}

VLA memory methods augment policies with learned banks, temporal representations,
or retrieved experience \citep{memoryvla,hamlet,memer}.
Robotic benchmarks test history dependence \citep{mikasarobo,robomme},
while MemoryAgentBench tests fact updating \citep{memoryagentbench}.
ChainBench separates decision and execution readouts when query observations
omit the deciding fact.

Causal scrubbing and activation patching probe behavioral dependence on internal
variables \citep{causalscrubbing,zhangnanda}. PoPE uses placebo controls to
distinguish useful error content from presentation effects in code-model
self-repair \citep{pope}. IMEP adapts these principles to embodied memory through
fact-changing and fact-preserving interventions at a specified retrieval interface,
paired outcome flips, and delivery checks. Appendix~\ref{sec:related} expands
these connections and their limits.

\section{Interventional Memory Evaluation Protocol}
\label{sec:method}\label{sec:protocol}

\subsection{Evaluation Target and Paired Design}
\label{sec:overview}\label{sec:formal}\label{sec:metric}

\begin{figure}[!t]
\centering
\includegraphics[width=\linewidth]{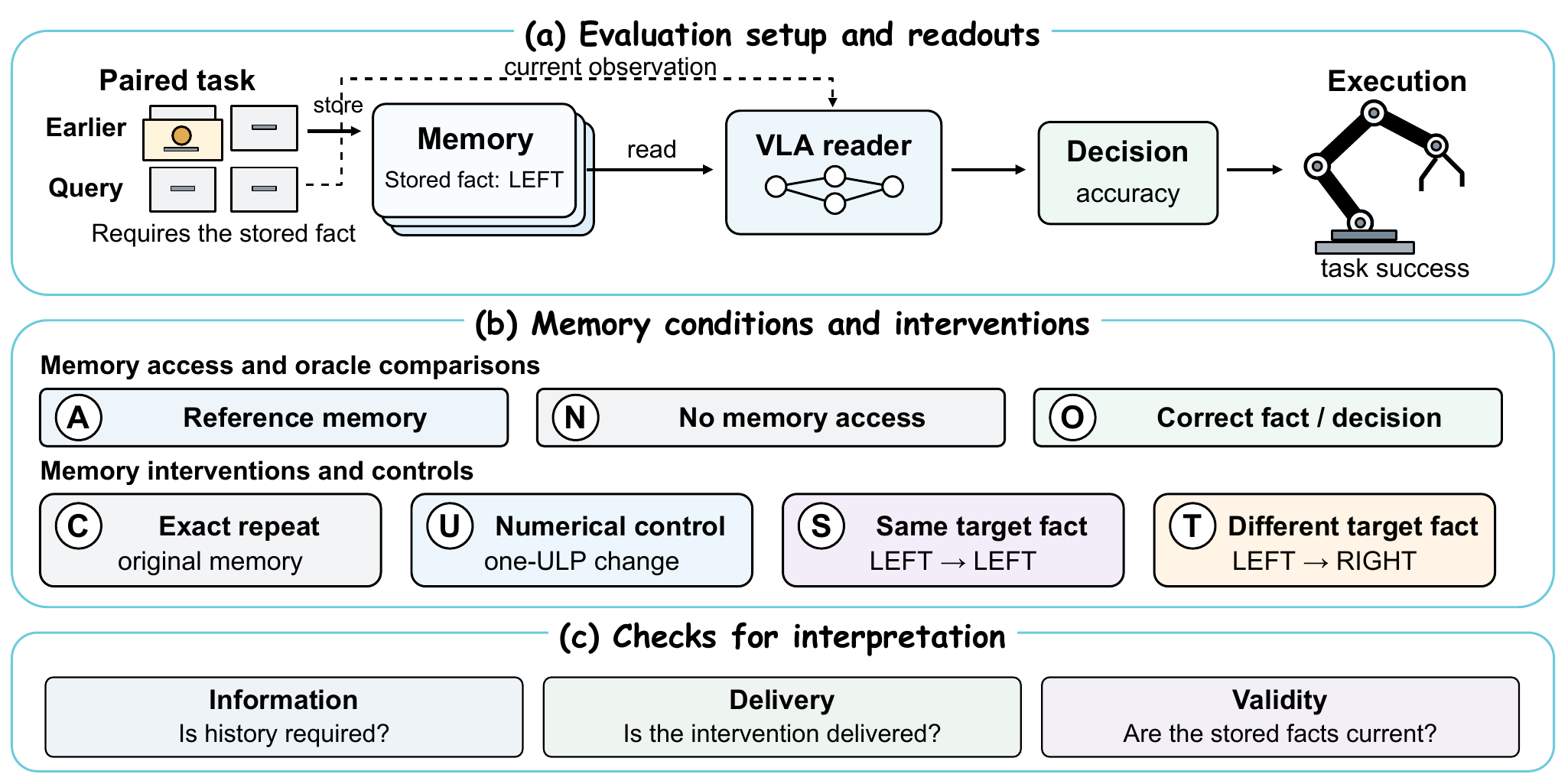}
\caption{\textbf{Overview of IMEP.}
(a) Paired evaluation follows the memory--decision--execution workflow, with
separate decision (where observable) and task-success readouts. (b) Access and oracle comparisons
($A,N,O$), interventions and controls ($C,U,S,T$). (c) Information, delivery,
and record-validity checks guide interpretation. The drawer task is illustrative.
Implementation limits, including installation effects and unverified semantic
edits, are detailed in Section~\ref{sec:arms}.}
\label{fig:imep-overview}
\end{figure}

IMEP evaluates whether memory benefits VLA systems and whether their responses
to memory interventions are specific to task-relevant facts. Using paired
comparisons and control interventions (Figure~\ref{fig:imep-overview}), it
addresses four questions:
\begin{itemize}[nosep]
\item[\textbf{Q1.}] \textbf{Net benefit:} Does memory provide a net benefit?
\item[\textbf{Q2.}] \textbf{Conditions:} Under what conditions does memory help?
\item[\textbf{Q3.}] \textbf{Outcome sensitivity:} Do changes to memory alter task outcomes?
\item[\textbf{Q4.}] \textbf{Fact specificity:} Are these effects specific to task-relevant facts?
\end{itemize}
The same task--seed pairs or multi-episode chains are evaluated across
conditions. Let $A$ denote reference memory, $X$ an intervention, and
$Y_i^X\in\{0,1\}$ task success for unit $i$. We specify the memory interface,
intervention time, controlled variables, and execution randomness, and score
decisions separately where observable. Shared seeds do not guarantee identical
execution. The framework and interpretation criteria were developed after the
campaigns closed. They do not retroactively make post hoc checks confirmatory
(Appendix~\ref{app:amendments}).

\subsection{Intervention Design and Controls}
\label{sec:arms}

\textbf{Access and oracle comparisons ($A,N,O$).} $A$ retains reference
memory, $N$ blocks reads or empties the bank, and $O$ supplies the correct fact or
decision. Blocking access measures utility together with any changed computation.
\textbf{Controls ($C,U$).} $C$ repeats the reference exactly; $U$ applies a
one-ULP numerical change. \textbf{Fact substitutions ($S,T$).} With the retrieval
interface fixed at installation, $S$ substitutes a record with the same target
fact; $T$ substitutes one with a different target fact. Same-fact records can
differ in other content. In cross-episode learned banks, $U/S/T$ share an
installation routine bypassed by $C$, so comparing $U$ with $C$ does not isolate
numerical perturbation alone. Under Shell Game blackout, $S_\pi$ permutes and
$T_\mu$ resamples retrieved latent values, but these operations are not verified semantic edits
(Appendix~\ref{app:blackout-ladders}).

\subsection{Readouts: Benefit, Sensitivity, and Specificity}
\label{sec:net-benefit}\label{sec:rq4}\label{sec:sensitivity}\label{sec:rq3}

For $n$ paired units, signed benefit and outcome sensitivity are
\begin{equation}
D_X=\frac{1}{n}\sum_{i=1}^{n}(Y_i^A-Y_i^X),\qquad
\rho_X=\frac{1}{n}\sum_{i=1}^{n}\mathbf{1}[Y_i^A\neq Y_i^X].
\label{eq:decomp}
\end{equation}
Positive $D_X$ favors the reference. With $b_X$ success-to-failure and $c_X$
failure-to-success flips, these equal $(b_X-c_X)/n$ and $(b_X+c_X)/n$:
net benefit can hide offsetting changes. Q1 uses $D_N$, and Q3 compares $\rho_X$
with the exact-repeat floor. Q4 measures the additional outcome sensitivity
under different-fact versus same-fact substitutions:
\begin{equation}
\theta=\rho_T-\rho_S,\qquad
\bar\theta=\frac{1}{|\mathcal{T}|}\sum_{t\in\mathcal{T}}(\rho_{T,t}-\rho_{S,t}),
\label{eq:discordance}
\end{equation}
where $\mathcal{T}$ indexes tasks and $\bar\theta$ weights them equally.
Differences between rates are percentage-point changes. Specificity depends on
controlling non-target differences and delivering replacements to the reader.
Decision flips and motor-outcome flips are separate readouts and are not pooled.

\subsection{Validity Checks and Interpretation}
\label{sec:conditions}\label{sec:rq1}\label{sec:rq2}\label{sec:step0}

To interpret the measured effects, Q2 examines whether the task requires history,
whether interventions reach the reader, and whether stored facts remain valid.
These checks guide attribution rather than specify universal prerequisites for
memory to help.

\textbf{(a) Information:} a history-required task hides the deciding
fact from the current observation and requires one choice among $K$ options,
with no recovery by trying alternatives. Where decisions are observed, we
compare no-memory accuracy with $1/K$ and require a positive lower bound on
$G=\mathrm{Acc}_O-\mathrm{Acc}_N$. A no-memory accuracy interval containing
$1/K$ does not establish equivalence to chance or rule out relevant information
in model weights.

\textbf{(b) Delivery:} check reads in $A$, empty reads in $N$, and substitution
coverage: the fraction of units whose every decision query selects at least one
unmodified substituted record. Coverage of one verifies this operational
check, not fact use. When coverage is below one, interpretation concerns swaps
as delivered.

\textbf{(c) Validity:} vary persistence while holding other components fixed.
Whole-configuration contrasts cannot isolate the effect of stale information.

\textbf{Measurement and decision rules.} We retain campaign-specific verdicts
and magnitude criteria. Under the
retrospective rules, Q3 requires the lower bound on $\rho_X$ to exceed the
repeatability floor; Q4 requires the lower bound on $\theta$ to exceed zero
and the upper bound on $\rho_U$, subject to the checks above. Remember Color's
registered equivalence analysis requires the 90\% CI of $\bar\theta$ within
$\pm7.5$ percentage points \citep{lakens2017}. The analysis concerns swaps as delivered
and does not exclude smaller effects. Nonsignificance alone is not equivalence.
Appendix~\ref{app:protocol-details} expands these checks and their implementation
limits.

\subsection{ChainBench: A Memory-Required Evaluation Setting}
\label{sec:chainbench}\label{sec:instruments}

A write episode reveals an object's drawer. Same-scene distractor episodes
then write other records before a query with closed drawers and identical observations
across targets. The robot commits to one drawer before a frozen \pihalf{} policy
executes that choice. In the 202-chain study, the full-memory configuration uses
a stipulated latest-write set, a calibrated presence gate, two-stage retrieval,
and a leave-one-chain-out nearest-centroid readout, with no decision-time LLM
call or evaluated eviction policy. Conditions share recorded executions within
each chain, so identical correct decisions imply equal task success at the
observed oracle ceiling $\mathrm{SR}^*=\mathrm{SR}_O$, where $\mathrm{SR}_O$ is
the oracle's observed task-success rate. The separate $C/U/S/T$ study tests only
the scripted decision reader on curated records, not motor outcomes or live
memory writing and retrieval (Appendices~\ref{app:chainbench}
and~\ref{app:executor-floors}).

\section{Experiments}
\label{sec:experiments}

\subsection{Experimental Setups}
\label{sec:setup}\label{sec:settings}

\begin{table}[t]
\centering\scriptsize
\newcommand{\ciOffset}[1]{\scalebox{0.8}{$\scriptstyle #1$}}
\setlength{\tabcolsep}{2pt}
\caption{\textbf{Net effects of memory access on task success.}
$A$: memory on; $X$: listed intervention; $n$: paired outcomes.
$D=\mathrm{SR}_A-\mathrm{SR}_X$ (percentage points, pp) favors memory when positive.
Superscripts/subscripts give distances to upper/lower confidence limits:
$90\%$ CIs for unmarked rows, $95\%$ for $\dagger$.
The conservative CP--Bonferroni $90\%$ CIs are post hoc additions except for
Bridge grid. The $95\%$ CIs retain study-specific methods (Appendices A, D, and F).
Verdicts retain the original decision rules.}
\label{tab:net-benefit}
\vspace{2pt}
\begin{tabular*}{\textwidth}{@{\extracolsep{\fill}}>{\raggedright\arraybackslash}m{3.20cm}>{\raggedright\arraybackslash}m{2.10cm}rrrr>{\raggedright\arraybackslash}m{1.80cm}@{}}
\toprule
Setting & Intervention $X$ & $n$ & $\mathrm{SR}_A$ (\%) & $\mathrm{SR}_X$ (\%) & $D$ (pp) & Verdict \\
\midrule
\multicolumn{7}{@{}l}{\emph{Layer 1, within-episode learned banks}} \\
Shell Game, camera blackout & read blocked & 200 & 73.5 & 69.5 & ${+4.0}^{\ciOffset{+4.4}}_{\ciOffset{-4.7}}$ & below the 10-pp criterion \\
Shell Game, view available & read blocked & 200 & 94.0 & 93.5 & ${+0.5}^{\ciOffset{+3.7}}_{\ciOffset{-3.8}}$ & not detected \\
Shell Game blackout ladder, released & both reads blocked & 400 & 67.25 & 63.00 & ${+4.25}^{\ciOffset{+3.39}}_{\ciOffset{-3.51}}$ & detected \\
Shell Game blackout ladder, structured PCMB & both reads blocked & 400 & 66.50 & 63.00 & ${+3.50}^{\ciOffset{+3.25}}_{\ciOffset{-3.35}}$ & detected \\
SimplerEnv Fractal, 12 cells & perceptual bank removed & 300 & 91.0 & 92.7 & ${-1.7}^{\ciOffset{+4.6}}_{\ciOffset{-4.4}}$ & not detected \\
SimplerEnv Bridge grid & read blocked & 72 & 65.3 & 55.6 & ${+9.7}^{\ciOffset{+16.1}}_{\ciOffset{-17.0}}$ & unresolved \\
\midrule
\multicolumn{7}{@{}l}{\emph{Layer 2, cross-episode learned banks}} \\
Shell Game attempt 2 & attempt-1 records cleared & 200 & 86.5 & 86.5 & ${0.0}^{\ciOffset{+4.0}}_{\ciOffset{-4.0}}$ & post hoc, no verdict \\
Shell Game persistent-read ablation & persistent reads emptied & 200 & 90.5 & 90.5 & ${0.0}^{\ciOffset{+3.4}}_{\ciOffset{-3.4}}$ & not detected \\
Remember Color 3/5/9, persistence under conflict$^{\dagger}$ & stale records cleared & 690 & 23.91 & 23.62 & ${+0.29}^{\ciOffset{+3.19}}_{\ciOffset{-3.33}}$ & not detected \\
Bridge carry, token-merge (attempts 4--5)$^{\dagger}$ & fresh memory & 200 & 29.0 & 31.5 & ${-2.5}^{\ciOffset{+7.5}}_{\ciOffset{-8.0}}$ & not detected \\
Bridge carry, structured/versioned (attempts 4--5)$^{\dagger}$ & fresh memory & 200 & 31.0 & 31.5 & ${-0.5}^{\ciOffset{+6.5}}_{\ciOffset{-7.0}}$ & not detected \\
Bridge carry, hybrid (attempts 4--5)$^{\dagger}$ & fresh memory & 200 & 32.5 & 31.5 & ${+1.0}^{\ciOffset{+5.5}}_{\ciOffset{-5.0}}$ & not detected \\
Trained memory plugin$^{\dagger}$ & empty bank & 500 & 10.2 & 18.4 & ${-8.2}^{\ciOffset{+4.2}}_{\ciOffset{-4.4}}$ & detected harm \\
\bottomrule
\end{tabular*}
\end{table}

We evaluate within-episode, cross-episode, and agentic memory, reporting
Q1--Q4 in that order for each layer. Table~\ref{tab:net-benefit} summarizes
the learned-bank net-benefit comparisons, Figure~\ref{fig:sham} shows
responses to memory interventions, and Table~\ref{tab:chainbench} reports
ChainBench decision accuracy and task success. The four-question framework
and its interpretation criteria were developed after the campaigns ended.
The cross-episode survival audit was also post hoc
(Appendix~\ref{app:amendments}).

\textbf{Layer 1 (within-episode learned banks).} On MIKASA-Robo Shell Game
\citep{mikasarobo}, we evaluate episode-local memory using a community
checkpoint of the MemoryVLA architecture \citep{memoryvla} in the released
and structured-PCMB configurations. Comparisons between these configurations
do not isolate the effect of memory access. We test with the view available
and under an out-of-distribution camera blackout from elapsed
step~6 to episode end, covering 85 of 90 frames. On SimplerEnv
\citep{simplerenv}, we remove the perceptual bank across twelve Fractal
coke-can configurations (Fractal) and block episode-local memory reads on
four Bridge tasks. Each Bridge task uses 24 seed-paired episodes. Three
tasks enter the 72-pair analysis under the preregistered inclusion rule
(Bridge grid).

\textbf{Layer 2 (cross-episode learned banks).} On MIKASA-Robo, each
task--seed pair is evaluated in two episodes (attempts~1 and~2), allowing
the second episode to read records written during the first. We evaluate
cross-episode memory in Shell Game under structured PCMB. On Remember Color~3 and~5 (RC3, RC5), the preregistered
substitution study changes stored contents once at the start of attempt~2,
preserving the initial retrieval path. We also test version
conflicts on Remember Color~3, 5, and~9, memory carryover along
five-attempt Bridge chains (Bridge carry), and a trained neural memory plugin
on Remember Color and Shell Game.
Appendix~\ref{app:checkpoint-provenance} documents the checkpoint and
implementation for each study.

\textbf{Layer 3 (agentic text memory).} We evaluate agentic text memory
using ChainBench (Section~\ref{sec:chainbench}) on LIBERO \citep{libero}.
The main study uses 202 chains with $K=2$ and a frozen \pihalf{} policy
\citep{pi05} for motor execution. A supporting study uses 143 chains with
$K=3$, a different executor, and a different chain manifest. We analyze
their results separately. We compare the full memory condition
on curated records with an initial-fact baseline that ignores
later updates, alongside no-memory, shuffled-record, and oracle-decision
controls. We report decision accuracy and task success on the same chains.
A separate live-harness coordinate study is reported in
Appendix~\ref{app:coordinate}.

\textbf{Detectability.} We interpret undetected effects in light of the
controls and resolution available for each readout.
Removing the post-cue view reduces Shell Game success from $94.0\%$ to
$73.5\%$ over $200$ paired episodes, showing that the success readout detects
this large visual intervention. In the tested learned-bank simulators, exact
replicates change $0/388$ outcomes on the common Remember Color cohort and
$0/400$ on the Shell Game blackout cohort. The scripted ChainBench reader
provides a positive control for decision accuracy. These controls do not
establish sensitivity to small fact-specific effects in learned banks.
Control details and design-specific resolution estimates are reported in
Appendix~\ref{app:power}.

\subsection{Within-Episode Memory}
\label{sec:within}\label{sec:layer1}

\begin{figure}[t]
\centering
\begin{tikzpicture}[x=1cm,y=1cm,line cap=round,
  every node/.style={font=\scriptsize,inner sep=1pt}]
  \node[font=\scriptsize\bfseries] at (2.40,4.20) {(a) Shell Game blackout};
  \node[text=black!65] at (2.40,3.88) {95\% CI};
  \foreach \x/\lab in {0.55/0,1.83/4,3.11/8,4.39/12} {
    \draw[gray!12,line width=.35pt] (\x,1.35) -- (\x,3.05);
    \draw (\x,1.29) -- (\x,1.41);
    \node[anchor=north] at (\x,1.27) {\lab};
  }
  \draw[thick] (0.55,1.35) -- (4.39,1.35);
  \node[anchor=north] at (2.47,0.92) {outcomes changed (\%)};
  \foreach \y/\lab in {3.45/{$C$},2.85/{$U$},2.30/{$S_\pi$},1.75/{$T_\mu$}}
    \node[anchor=east] at (0.45,\y) {\lab};
  \foreach \y/\lo/\m/\hi/\col in {3.54/0.550/0.550/0.844/gray!75,
      2.94/0.636/0.870/1.363/armU, 2.39/1.990/2.710/3.644/armS,
      1.84/2.121/2.870/3.830/armT} {
    \draw[draw=\col,line width=.7pt] (\lo,\y) -- (\hi,\y);
    \fill[fill=\col] (\m,\y) circle (1.8pt);
  }
  \foreach \y/\lo/\m/\hi/\col in {3.36/0.550/0.550/0.844/gray!75,
      2.76/0.937/1.350/2.006/armU, 2.21/1.792/2.470/3.366/armS,
      1.66/2.054/2.790/3.737/armT} {
    \draw[draw=\col,line width=.7pt] (\lo,\y) -- (\hi,\y);
    \filldraw[fill=white,draw=\col,line width=.7pt] (\m,\y) circle (1.8pt);
  }
  \fill[black] (1.45,3.58) circle (1.8pt);
  \node[anchor=west] at (1.56,3.58) {released};
  \filldraw[fill=white,draw=black,line width=.7pt] (1.45,3.28) circle (1.8pt);
  \node[anchor=west] at (1.56,3.28) {structured};

  \node[font=\scriptsize\bfseries] at (7.80,4.20) {(b) Remember Color, as delivered};
  \node[text=black!65] at (6.55,3.88) {95\% CI};
  \node[text=black!65] at (9.75,3.88) {90\% CI};
  \foreach \x/\lab in {5.15/0,6.55/10,7.95/20} {
    \draw[gray!12,line width=.35pt] (\x,1.35) -- (\x,3.70);
    \draw (\x,1.29) -- (\x,1.41);
    \node[anchor=north] at (\x,1.27) {\lab};
  }
  \draw[thick] (5.15,1.35) -- (7.95,1.35);
  \node[anchor=north] at (6.55,0.92) {outcomes changed (\%)};
  \foreach \y/\lab in {3.45/{$C$},2.85/{$U$},2.30/{$S$},1.75/{$T$}}
    \node[anchor=east] at (5.05,\y) {\lab};
  \foreach \y/\lo/\m/\hi/\col/\cnt in {2.85/6.383/6.810/7.320/armU/{46/388},
      2.30/6.543/6.990/7.519/armS/{51/388},
      1.75/6.670/7.135/7.677/armT/{55/388}} {
    \draw[draw=\col,line width=.7pt] (\lo,\y) -- (\hi,\y);
    \draw[draw=\col,line width=.7pt] (\lo,\y-0.07) -- (\lo,\y+0.07);
    \draw[draw=\col,line width=.7pt] (\hi,\y-0.07) -- (\hi,\y+0.07);
    \fill[fill=\col] (\m,\y) circle (1.9pt);
    \node[anchor=east] at (\lo-0.06,\y) {\cnt};
  }
  \draw[gray!75,line width=.7pt] (5.150,3.45) -- (5.283,3.45);
  \filldraw[fill=white,draw=gray!75,line width=.7pt] (5.15,3.45) circle (1.9pt);
  \node[anchor=west] at (5.36,3.45) {0/388};
  \fill[gray!12] (9.00,1.60) rectangle (10.50,2.60);
  \draw[gray!55,line width=.45pt] (9.75,1.35) -- (9.75,2.60);
  \draw[thick] (8.85,1.35) -- (10.65,1.35);
  \foreach \x/\lab in {9.00/{$-7.5$},9.75/0,10.50/{$+7.5$}} {
    \draw (\x,1.29) -- (\x,1.41);
    \node[anchor=north] at (\x,1.27) {\lab};
  }
  \node[anchor=north] at (9.75,0.92) {$\bar\theta$ (pp)};
  \draw[line width=1pt] (9.516,2.10) -- (10.191,2.10);
  \draw[line width=1pt] (9.516,2.00) -- (9.516,2.20);
  \draw[line width=1pt] (10.191,2.00) -- (10.191,2.20);
  \fill[black] (9.853,2.10) circle (2.1pt);
  \node at (9.75,3.22) {$\bar\theta{=}{+}1.03^{\scalebox{0.8}{$\scriptstyle +3.38$}}_{\scalebox{0.8}{$\scriptstyle -3.37$}}\,\mathrm{pp}$};

  \node[font=\scriptsize\bfseries] at (12.60,4.20) {(c) Scripted reader};
  \foreach \x/\lab in {11.70/0,13.50/100} {
    \draw (\x,1.29) -- (\x,1.41);
    \node[anchor=north] at (\x,1.27) {\lab};
  }
  \draw[thick] (11.70,1.35) -- (13.50,1.35);
  \node[anchor=north] at (12.60,0.92) {decisions changed (\%)};
  \foreach \y/\lab/\col in {3.45/{$C$}/gray!75,2.85/{$U$}/armU,2.30/{$S$}/armS} {
    \node[anchor=east] at (11.60,\y) {\lab};
    \draw[draw=\col,line width=1.6pt] (11.70,\y-0.15) -- (11.70,\y+0.15);
    \node[anchor=west] at (11.78,\y) {0/401};
  }
  \node[anchor=east] at (11.60,1.75) {$T$};
  \fill[armT] (11.70,1.60) rectangle (13.50,1.90);
  \node[text=white] at (12.60,1.75) {401/401};
\end{tikzpicture}
\caption{\textbf{Outcome sensitivity and fact specificity across layers.}
(a) Shell Game blackout: released (filled) and structured-PCMB (hollow) markers,
with $400$ pairs per condition. (b) Remember Color: $388$ pairs with swaps evaluated as delivered
(coverage $0.70$--$0.83$). The policy is near the single-choice chance benchmark.
Outcome-change intervals in (a,b)
are exact $95\%$ Clopper--Pearson CIs. The black interval is a task-stratified
paired-bootstrap $90\%$ CI for the task-averaged different-fact minus same-fact
sensitivity $\bar\theta$. Shading denotes the preregistered $\pm7.5$-pp
equivalence band. (c) Scripted ChainBench reader on curated records ($401$
chains): only $T$ changes decisions. This single experiment provides both Q4
evidence for that reader and the decision-readout control, not independent
evidence.}
\label{fig:sham}
\end{figure}
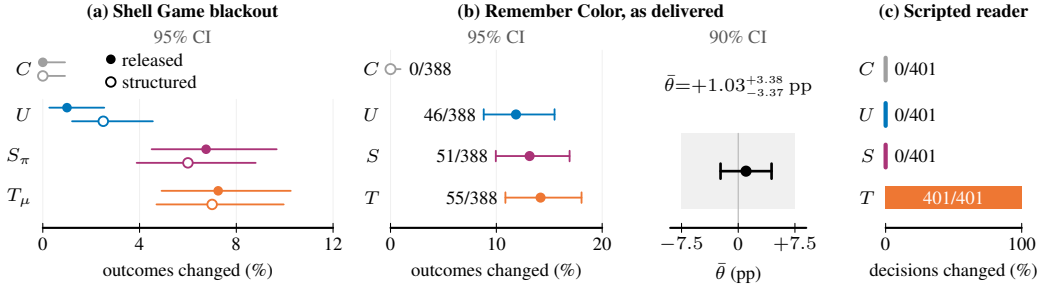

Layer~1 covers within-episode learned banks in Shell Game with the view
available or under camera blackout, the Fractal perceptual bank, and the
Bridge grid.

\textbf{Q1 (net benefit).} Under Shell Game blackout, reading memory raises
the success rate from $69.5\%$ to $73.5\%$, as shown in
Table~\ref{tab:net-benefit}. This gain is statistically detectable
($p=0.0215$), but its one-sided upper bound of $7.70\,\mathrm{pp}$ falls below the
registered $10$-pp criterion (Appendix~\ref{app:viewremoval}). The two separate
blackout ladders also show detectable gains, although neither study registered
a magnitude criterion. With the view available, the $0.5$-point gain is not
statistically detectable. The separate Shell Game within-episode
read-ablation study also shows no detectable benefit ($D=-0.5\,\mathrm{pp}$, with a post hoc upper bound of $3.14\,\mathrm{pp}$,
Appendix~\ref{app:readpath}). Removing the Fractal perceptual bank produces no
detectable change in success, while the Bridge grid remains inconclusive
(Appendix~\ref{app:resolution}). These studies do not measure decision
accuracy separately, so the decision--execution comparison in Section~\ref{sec:chainbench} cannot be
evaluated.

\textbf{Q2 (conditions).} Shell Game counts any touch of the correct cup as
success, even after an incorrect choice. Success therefore does not require
a single correct decision, and condition (a) in Section~\ref{sec:conditions}
fails both with the view available and under blackout. In a post hoc
check of the separate read-ablation study with the view available, the
first cup visited is correct in $88/177$ successful
episodes with memory reads and $88/178$ without
(Appendix~\ref{app:sg-anytouch}). First-choice accuracy under blackout was
not computed. Both blackout ladders pass the registered survival check:
substituted content is present at the decision step in every check
(Appendix~\ref{app:survival}). This satisfies condition (b) in
Section~\ref{sec:conditions}.

\textbf{Q3 (sensitivity).} Both latent permutation ($S_\pi$) and resampling
($T_\mu$) change task outcomes, as shown in Figure~\ref{fig:sham}a
($400$ pairs per condition). Their outcome-change rates are $6.75\%$ and $7.25\%$
in the released configuration, and $6.00\%$ and $7.00\%$ under structured
PCMB. The one-ULP control changes $1.00\%$ of outcomes in the released
configuration and $2.50\%$ under structured PCMB, while exact replicates
change $0/400$ outcomes in each configuration. Paired counts are reported in
Appendix~\ref{app:blackout-ladders}. In a separate 200-pair Fractal
payload-substitution study, the original and neutral banks yield different
outcomes in $21/200$ episodes ($10.5\%$). The upper
endpoint of the exact Clopper--Pearson 95\% confidence interval is $15.6\%$.
This comparison has no one-ULP or latent-matched control, so it does not
isolate a fact-specific effect (Appendix~\ref{app:footprint}).

\textbf{Q4 (specificity).} Resampling does not change detectably more outcomes
than permutation. The difference in outcome-change rates,
$\theta=\rho_{T_\mu}-\rho_{S_\pi}$, is $+0.50\,\mathrm{pp}$ in the released
configuration and $+1.00\,\mathrm{pp}$ under structured PCMB (exact one-sided McNemar
tests, $p=0.387$ and $p=0.145$, respectively). These tests do not establish
equivalence between the two interventions. The two configurations were not
statistically compared with each other. In addition, the released ladder's
sham-validity check was corrected after data collection
(Appendix~\ref{app:blackout-ladders}). Reading memory therefore provides a
modest success gain under blackout, but the interventions do not establish
that this gain reflects use of the task-relevant fact.

\subsection{Cross-Episode Memory}
\label{sec:cross}\label{sec:layer2}

Layer~2 covers Shell Game memory across episodes, Remember Color substitutions
and version conflicts, Bridge carry, and the trained plugin.

\textbf{Q1 (net benefit).} In the Shell Game attempt-2 study, clearing
attempt-1 records leaves success unchanged at $86.5\%$, as shown in
Table~\ref{tab:net-benefit}. This comparison was analyzed post hoc and is not
used for a formal verdict. In the separate Shell Game persistent-read
ablation, success likewise remains unchanged at $90.5\%$. Clearing stale
Remember Color records produces no detectable change in success, nor does
carrying Bridge memory across attempts. For the trained plugin, retaining
the memory bank lowers success from $18.4\%$ to $10.2\%$, a success-rate
difference of $\bodyCI{-8.2}{+4.2}{-4.4}$ percentage points
(95\%~CI; frozen adjudicator). This interval excludes zero, indicating harm, but its
upper endpoint does not fall below the prespecified threshold of $-5\,\mathrm{pp}$
(Appendix~\ref{app:trained}).

\textbf{Q2 (conditions).} Condition (a) in Section~\ref{sec:conditions} is not
satisfied by these task designs. Remember Color shows the cue again in
attempt~2, so the relevant fact is available without recalling the earlier
attempt. Under the released configuration, success is $34.0\%$ on RC3 and
$23.5\%$ on RC5, close to the respective single-choice chance benchmarks
of $33.3\%$ and $20.0\%$
(Appendix~\ref{app:rc-floors}). Shell Game counts any touch of the correct cup
as success, while Bridge changes the target object's position across attempts.
The trained plugin uses the same Remember Color and Shell Game tasks.
Condition (b) is not established either. Post hoc coverage is
$0.70$\,--\,$0.83$ for Remember Color, $0.61$\,--\,$0.68$ for the Shell Game full-vision
ladder, and $0.51$\,--\,$0.55$ for the plugin's RC5 substitutions
(Appendix~\ref{app:survival}). These values fall below the required coverage
of~1. However, coverage counts only unmodified records, and merged records
retain six of the eight substituted fields. Coverage below~1 therefore does
not mean that substituted content is entirely absent. The outdated-record
check for condition (c) was conducted only in the Remember Color
version-conflict study and remains inconclusive (Appendix~\ref{app:version}).

\textbf{Q3 (sensitivity).} Across 388 donor-eligible Remember Color pairs
(193 on RC3 and 195 on RC5), the one-ULP control $U$, same-fact swaps $S$,
and different-fact swaps $T$ change $\bodyCI{11.9}{+3.6}{-3.1}\%$
(95\%~CI for~$U$), $13.1\%$, and $14.2\%$ of outcomes, respectively,
as shown in Figure~\ref{fig:sham}b. Exact replicates change none.
Unlike exact replication, $U$ passes through the memory-installation routine,
so the comparison does not isolate numerical perturbation alone. It tests
only one perturbation magnitude and modifies stored payloads rather than
retrieved latents, limiting comparison with the Shell Game blackout controls.
Under $T$, outcome-change rates are $16.06\%$ on RC3 and $12.31\%$ on RC5,
but net success-rate differences are only $+3.63\,\mathrm{pp}$ and $+2.05\,\mathrm{pp}$
(Appendix~\ref{app:payload}). Small net differences therefore conceal changes
in individual outcomes.

\textbf{Q4 (specificity).} Different-fact swaps show no clear additional
effect over same-fact swaps. The task-averaged difference is
$\bar{\theta}=\bodyCI{+1.03}{+3.38}{-3.37}\,\mathrm{pp}$ (paired-bootstrap 90\%~CI;
Eq.~\eqref{eq:discordance}, Figure~\ref{fig:sham}b).
This meets the preregistered $\pm7.5$-pp equivalence criterion for swaps as
delivered, but does not exclude smaller fact-specific effects or establish
complete delivery (Appendices~\ref{app:resolution} and~\ref{app:survival}).
Comparisons of $U$ with $S$ and $T$ detect no differences but do not establish
equivalence (Appendix~\ref{app:payload}). The trained-plugin result remains
inconclusive ($\theta=\bodyCI{+1.53}{+3.06}{-3.06}\,\mathrm{pp}$, 95\% CI, nine discordant
pairs, Appendix~\ref{app:plugin-sham}). The Shell Game full-vision ladder was
withdrawn from Q3/Q4 evidence (Appendix~\ref{app:withdrawn}). Thus, sensitivity
to memory interventions is observed, but fact specificity is not established.

\subsection{Agentic Memory}
\label{sec:agentic}\label{sec:layer3}

\begin{table}[t]
\centering\small
\newcommand{\ciOffset}[1]{\scalebox{0.8}{$\scriptstyle #1$}}
\renewcommand{\arraystretch}{1.15}
\setlength{\tabcolsep}{10pt}
\caption{\textbf{ChainBench results.}
$202$ chains, $K=2$ (chance accuracy $50.0\%$). Superscripts/subscripts give
distances to upper/lower exact $95\%$ Clopper--Pearson confidence limits.
Full memory uses Section~\ref{sec:chainbench}'s configuration. The initial-fact
baseline ignores updates. Conditions share two recorded physical executions
per chain. Full memory and the oracle choose correctly on every chain, making
their task success rates equal by construction.}
\label{tab:chainbench}
\vspace{2pt}
\begin{tabular}{@{}lcc@{}}
\toprule
Condition & Decision accuracy (\%) & Task success (\%) \\
\midrule
No memory & ${53.0}^{\ciOffset{+7.0}}_{\ciOffset{-7.2}}$ & ${39.1}^{\ciOffset{+7.1}}_{\ciOffset{-6.8}}$ \\
Shuffled records & ${51.5}^{\ciOffset{+7.1}}_{\ciOffset{-7.1}}$ & ${38.6}^{\ciOffset{+7.1}}_{\ciOffset{-6.7}}$ \\
Initial-fact baseline & ${52.0}^{\ciOffset{+7.0}}_{\ciOffset{-7.1}}$ & ${39.1}^{\ciOffset{+7.1}}_{\ciOffset{-6.8}}$ \\
Full memory & ${100.0}^{\ciOffset{+0.0}}_{\ciOffset{-1.8}}$ & ${74.3}^{\ciOffset{+5.8}}_{\ciOffset{-6.6}}$ \\
Oracle decision & ${100.0}^{\ciOffset{+0.0}}_{\ciOffset{-1.8}}$ & ${74.3}^{\ciOffset{+5.8}}_{\ciOffset{-6.6}}$ \\
\bottomrule
\end{tabular}
\end{table}

\begin{figure}[htbp]
\centering
\begin{tikzpicture}[x=1cm,y=1cm]
\foreach \x/\file/\title in {
  0/panel1_write/Write,
  3.62/panel2_distractor/Distractor,
  7.24/panel3_distractor/Distractor,
  10.86/panel4_query/Query} {
  \node[anchor=south west,inner sep=0pt,outer sep=0pt] at (\x,0)
    {\includegraphics[width=2.97cm]{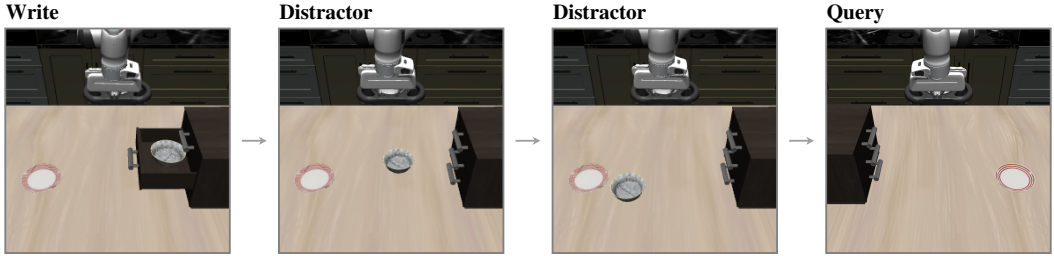}};
  \draw[draw={rgb,255:red,128;green,128;blue,128},line width=0.7pt] (\x,0) rectangle (\x+2.97,2.97);
  \node[anchor=base west,inner sep=0pt,font=\fontsize{8}{9.5}\selectfont\bfseries]
    at (\x,3.10) {\title};
}
\foreach \x in {2.97,6.59,10.21} {
  \draw[-{Stealth[length=3pt,width=3pt]},gray!75,line width=0.7pt]
    (\x+0.16,1.485) -- (\x+0.49,1.485);
}
\end{tikzpicture}
\caption{\textbf{ChainBench chain construction.}
A write episode reveals the target drawer, and same-scene distractors then write
other records. At query time, closed drawers give identical observations across
target locations, requiring recall of the earlier object--drawer relation.
Simulator frames illustrate the construction, not a rollout from the
202-chain study.}
\label{fig:chain}
\end{figure}

Layer~3 is agentic text memory, evaluated on ChainBench
(Figure~\ref{fig:chain}) with a 202-chain study ($K=2$) and a supporting
143-chain study ($K=3$) (Appendix~\ref{app:e1b}). Intervals are exact Clopper--Pearson 95\%
CIs unless marked.

\textbf{Q1 (net benefit).} The full memory configuration selects the correct
answer on all 202 chains, and task success rises from
$39.1\%$ without memory to $74.3\%$ (Table~\ref{tab:chainbench}; $b_N{=}71$,
$c_N{=}0$, \mbox{$D_N=\bodyCI{+35.15}{+6.45}{-6.45}\,\mathrm{pp}$}, chain-bootstrap 95\%~CI). The $74.3\%$ task-success rate matches the oracle-decision condition,
the observed motor-execution ceiling for these trials. With full memory, ChainBench is execution-limited in this study. All conditions of a
chain look up the same two physical executions. Given identical correct
decisions, full memory and the oracle necessarily have the same task success.
This equality does not make perfect decision accuracy a constructed result. In a post hoc descriptive check that assumes equal execution of
correct decisions across conditions, each prediction $\mathrm{Acc}_X\cdot\mathrm{SR}^*$
lies inside the observed success rate's CI: $39.3\%$ without memory, $38.6\%$ with
the initial-fact baseline and $38.2\%$ with shuffled records. In the 202-chain study, the
initial-fact baseline has the same observed success rate as no memory (no
equivalence test). In the supporting study with 143 chains and $K=3$, curated records
raise task success from $32.9\%$ without memory to $95.1\%$.
The initial-fact and shuffled-record baselines both reach $49.0\%$ there,
despite their different decision rules. These supporting results concern a
rebuilt manifest, for which potential selection bias remains unresolved
(Appendix~\ref{app:e1b}).

\textbf{Q2 (conditions).} Condition (a) holds by design: the query hides the
deciding fact, and the robot commits to one drawer before execution, with no
recovery by trying alternatives. Without memory, decision accuracy is $53.0\%$,
whose CI contains $1/K=50.0\%$ (non-rejection, not equivalence), and
$G=\bodyCI{+47.0}{+7.2}{-7.0}\,\mathrm{pp}$ (95\%~CI). The all-$K{=}2$ floor and $G$ come from a
post-closeout check. In a separate, non-preregistered retrieval diagnostic on 211 chains,
appearance top-1 retrieval selects the informative write in $0/211$ chains
(Appendix~\ref{app:chainbench}). Full memory matches oracle decision accuracy
($\mathrm{Acc}_A=\mathrm{Acc}_O$), providing behavioral evidence of successful
readout, but this equality alone does not verify condition (b). The stale-system
condition decides correctly on $0/97$ rewritten chains, demonstrating failure
under that stale-record configuration (Appendix~\ref{app:e1b}).

\textbf{Q3 (sensitivity).} Shuffling curated records reduces decision accuracy
to $51.5\%$ and task success to $38.6\%$, showing that decisions and outcomes
are sensitive to changes in the records. We did not apply the $C/U/S/T$ ladder
to motor outcomes because, in a separate execution-repeatability study,
replaying identical instructions changes about $27\%$ of outcomes
(Appendix~\ref{app:executor-floors}). This execution
variability would make the effects of memory interventions difficult to
distinguish.

\textbf{Q4 (specificity).} In a separate decision-level evaluation on 401 chains,
the nearest-centroid readout changes no decisions under $U$ or $S$
($0/401$ each), whereas
different-fact swaps $T$ change all 401 decisions (Figure~\ref{fig:sham}c;
Appendix~\ref{app:poscal}). These results support fact specificity for this
readout on curated records. The same experiment also serves as the positive
control for the decision readout, so these two uses do not provide independent
evidence. This ladder does not test memory writing or retrieval by the live
harness.

\textbf{Across layers.} Does memory work? No detected learned-bank gain meets a
registered net-benefit criterion, and the trained plugin's bank reduces task
success. On ChainBench, curated records in the 202-chain study raise task
success to the observed motor-execution ceiling. When can memory help? No learned-bank setting establishes both the
history requirement and delivery. ChainBench requires history, and full memory
matches oracle accuracy, although this equality alone does not verify delivery.
Are the effects fact-specific? Learned-bank outcomes respond to memory
interventions, including the one-ULP control and its installation procedure,
but fact specificity is not established. The separate
ChainBench decision-level experiment supports fact specificity for the
nearest-centroid readout on curated records (Tables~\ref{tab:net-benefit}
and~\ref{tab:chainbench}; Figure~\ref{fig:sham}).

\section{Conclusion}
\label{sec:conclusion}
\label{sec:limitations}

IMEP shows why memory-access gains and intervention sensitivity should be reported separately from evidence of fact-specific use. Across the evaluated learned banks, interventions alter outcomes, but the available controls do not establish fact specificity. ChainBench provides complementary evidence that historical records can support correct decisions, with specificity demonstrated for the evaluated reader on curated records. The main limitations concern measurement resolution and the scope of the evaluated systems. The learned-bank studies lack decision readouts and content positive controls, leaving small fact-specific effects unresolved. The ChainBench findings concern the evaluated memory configuration and readout, with task success measured using shared recorded executions. Generalization across checkpoints and implementations remains unverified. Future work should evaluate IMEP prospectively on tasks that require memory and trace task-relevant facts through writing, retrieval, and decision-making to identify where memory use fails.

\clearpage
\flushbottom

\subsection*{AI use statement}

Generative AI tools assisted with refining research ideas and experimental designs, developing and checking evaluation code and statistical analyses, identifying relevant literature, interpreting results, and revising the manuscript, including tables and figures. AI tools also supported planning and memory writing in selected ChainBench studies, as described in the methods and appendices. The authors take responsibility for the final text, claims, results, code, and artifacts.

\subsection*{Ethics statement}

All experiments were conducted in simulation using SimplerEnv, MIKASA-Robo,
and LIBERO. They involved no physical robots, human participants, or
personal data. We used public benchmarks and community checkpoints under
their respective licenses.

\subsection*{Reproducibility statement}

An anonymized package of available code, protocols, archived results, and
offline verification scripts is provided at
\url{https://anonymous.4open.science/r/IMEP-ChainBench-DC83/}.
The README documents verification scope and remaining gaps. These checks do
not constitute full experimental reruns. The corresponding non-anonymized
materials will be released upon publication.

Appendices~\ref{app:ledger}, \ref{app:checkpoint-provenance}, and
\ref{app:reproducibility} document evidence status, checkpoint and code
bindings, and archived artifacts. The protocols are
hash-bound but lack third-party timestamps. The MIKASA checkpoint is
MemoryVLA-compatible, not an official release. Equivalence between its
evaluation implementation and the sealed release runtime has not been
established (Appendix~\ref{app:checkpoint-provenance}). Archived decisions
from the 143- and 202-chain ChainBench studies support replay, while the
closed-weight planner sessions cannot be regenerated bit-for-bit.
Appendix~\ref{app:chainbench} describes the 202-chain memory configuration
and its retrieval and nearest-centroid readout. The exact historical source
version of that study's decision evaluator was not recorded. Writer-study
checksums were established at audit time, not at launch
(Appendix~\ref{app:writer-audit}). Appendix~\ref{app:amendments} records
post hoc analyses, including the IMEP framework and its interpretation
criteria, and post-data amendments. Appendix~\ref{app:excluded} separates
exploratory, superseded, excluded, and withdrawn studies from confirmatory
evidence.

\bibliography{references}

\begin{thebibliography}{40}
\providecommand{\natexlab}[1]{#1}
\providecommand{\url}[1]{\texttt{#1}}
\expandafter\ifx\csname urlstyle\endcsname\relax
  \providecommand{\doi}[1]{doi: #1}\else
  \providecommand{\doi}{doi: \begingroup \urlstyle{rm}\Url}\fi

\bibitem[Adebayo et~al.(2018)Adebayo, Gilmer, Muelly, Goodfellow, Hardt, and
  Kim]{sanitychecks}
Julius Adebayo, Justin Gilmer, Michael Muelly, Ian Goodfellow, Moritz Hardt,
  and Been Kim.
\newblock Sanity checks for saliency maps.
\newblock In \emph{Advances in Neural Information Processing Systems},
  volume~31, 2018.

\bibitem[Black et~al.(2025{\natexlab{a}})Black, Brown, Darpinian, Dhabalia,
  Driess, Esmail, Equi, Finn, Fusai, Galliker, Ghosh, Groom, Hausman, Ichter,
  Jakubczak, Jones, Ke, LeBlanc, Levine, Li-Bell, Mothukuri, Nair, Pertsch,
  Ren, Shi, Smith, Springenberg, Stachowicz, Tanner, Vuong, Walke, Walling,
  Wang, Yu, and Zhilinsky]{pi05}
Kevin Black, Noah Brown, James Darpinian, Karan Dhabalia, Danny Driess, Adnan
  Esmail, Michael~Robert Equi, Chelsea Finn, Niccolo Fusai, Manuel~Y. Galliker,
  Dibya Ghosh, Lachy Groom, Karol Hausman, Brian Ichter, Szymon Jakubczak, Tim
  Jones, Liyiming Ke, Devin LeBlanc, Sergey Levine, Adrian Li-Bell, Mohith
  Mothukuri, Suraj Nair, Karl Pertsch, Allen~Z. Ren, Lucy~Xiaoyang Shi, Laura
  Smith, Jost~Tobias Springenberg, Kyle Stachowicz, James Tanner, Quan Vuong,
  Homer Walke, Anna Walling, Haohuan Wang, Lili Yu, and Ury Zhilinsky.
\newblock {$\pi_{0.5}$}: A vision-language-action model with open-world
  generalization.
\newblock In \emph{Proceedings of The 9th Conference on Robot Learning}, volume
  305 of \emph{Proceedings of Machine Learning Research}, pp.\  17--40. PMLR,
  2025{\natexlab{a}}.

\bibitem[Black et~al.(2025{\natexlab{b}})Black, Brown, Driess, Esmail, Equi,
  Finn, Fusai, Groom, Hausman, Ichter, Jakubczak, Jones, Ke, Levine, Li-Bell,
  Mothukuri, Nair, Pertsch, Shi, Smith, Tanner, Vuong, Walling, Wang, and
  Zhilinsky]{pi0}
Kevin Black, Noah Brown, Danny Driess, Adnan Esmail, Michael~Robert Equi,
  Chelsea Finn, Niccolo Fusai, Lachy Groom, Karol Hausman, Brian Ichter, Szymon
  Jakubczak, Tim Jones, Liyiming Ke, Sergey Levine, Adrian Li-Bell, Mohith
  Mothukuri, Suraj Nair, Karl Pertsch, Lucy~Xiaoyang Shi, Laura Smith, James
  Tanner, Quan Vuong, Anna Walling, Haohuan Wang, and Ury Zhilinsky.
\newblock {$\pi_0$}: A vision-language-action flow model for general robot
  control.
\newblock In \emph{Proceedings of Robotics: Science and Systems}, Los Angeles,
  CA, USA, June 2025{\natexlab{b}}.

\bibitem[Chan et~al.(2022)Chan, Garriga-Alonso, Goldowsky-Dill, Greenblatt,
  Nitishinskaya, Radhakrishnan, Shlegeris, and Thomas]{causalscrubbing}
Lawrence Chan, Adri\`a Garriga-Alonso, Nicholas Goldowsky-Dill, Ryan
  Greenblatt, Jenny Nitishinskaya, Ansh Radhakrishnan, Buck Shlegeris, and Nate
  Thomas.
\newblock Causal scrubbing: A method for rigorously testing interpretability
  hypotheses.
\newblock \emph{AI Alignment Forum}, 2022.
\newblock URL
  \url{https://www.alignmentforum.org/posts/JvZhhzycHu2Yd57RN/causal-scrubbing-a-method-for-rigorously-testing}.

\bibitem[Chen et~al.(2026)Chen, Wang, Li, Qin, Shi, Li, Hu, Zhang, Wang, Chen,
  Wang, Wang, Yang, Xu, Wu, Mu, Yang, Dong, and Luo]{rmbench}
Tianxing Chen, Yuran Wang, Mingleyang Li, Yan Qin, Hao Shi, Zixuan Li, Yifan
  Hu, Yingsheng Zhang, Kaixuan Wang, Yue Chen, Hongcheng Wang, Junjie Wang,
  Tianhang Yang, Renjing Xu, Ruihai Wu, Yao Mu, Yaodong Yang, Hao Dong, and
  Ping Luo.
\newblock {RMBench}: Memory-dependent robotic manipulation benchmark with
  insights into policy design.
\newblock \emph{arXiv preprint arXiv:2603.01229}, 2026.

\bibitem[Cherepanov et~al.(2026)Cherepanov, Kachaev, Kovalev, and
  Panov]{mikasarobo}
Egor Cherepanov, Nikita Kachaev, Alexey~K. Kovalev, and Aleksandr~I. Panov.
\newblock Memory, benchmark \& robots: A benchmark for solving complex tasks
  with reinforcement learning.
\newblock In \emph{The Fourteenth International Conference on Learning
  Representations}, pp.\  9759--9815, 2026.

\bibitem[Dai et~al.(2026)Dai, Fu, Lee, Liu, Zhang, Yang, Finn, Fazeli, and
  Chai]{robomme}
Yinpei Dai, Hongze Fu, Jayjun Lee, Yuejiang Liu, Haoran Zhang, Jianing Yang,
  Chelsea Finn, Nima Fazeli, and Joyce Chai.
\newblock {RoboMME}: Benchmarking and understanding memory for robotic
  generalist policies.
\newblock In \emph{Proceedings of the 43rd International Conference on Machine
  Learning}, 2026.

\bibitem[de~Haan et~al.(2019)de~Haan, Jayaraman, and Levine]{causalconfusion}
Pim de~Haan, Dinesh Jayaraman, and Sergey Levine.
\newblock Causal confusion in imitation learning.
\newblock In \emph{Advances in Neural Information Processing Systems},
  volume~32, 2019.

\bibitem[{Figure AI}(2025)]{helix}
{Figure AI}.
\newblock Helix: A vision-language-action model for generalist humanoid
  control.
\newblock \url{https://www.figure.ai/news/helix}, 2025.
\newblock Company technical blog; accessed July 9, 2026.

\bibitem[Geiger et~al.(2021)Geiger, Lu, Icard, and Potts]{causalabstractions}
Atticus Geiger, Hanson Lu, Thomas Icard, and Christopher Potts.
\newblock Causal abstractions of neural networks.
\newblock In \emph{Advances in Neural Information Processing Systems},
  volume~34, pp.\  9574--9586, 2021.

\bibitem[Hu et~al.(2026)Hu, Wang, and McAuley]{memoryagentbench}
Yuanzhe Hu, Yu~Wang, and Julian McAuley.
\newblock Evaluating memory in {LLM} agents via incremental multi-turn
  interactions.
\newblock In \emph{The Fourteenth International Conference on Learning
  Representations}, pp.\  156259--156291, 2026.

\bibitem[Iscan(2026)]{pope}
Mehmet Iscan.
\newblock Form, not content? a preregistered, placebo-controlled evaluation of
  learned error-conditioned self-repair through prompts and weights in frozen
  small code models.
\newblock \emph{arXiv preprint arXiv:2607.12962}, 2026.

\bibitem[Kim et~al.(2025)Kim, Pertsch, Karamcheti, Xiao, Balakrishna, Nair,
  Rafailov, Foster, Sanketi, Vuong, Kollar, Burchfiel, Tedrake, Sadigh, Levine,
  Liang, and Finn]{openvla}
Moo~Jin Kim, Karl Pertsch, Siddharth Karamcheti, Ted Xiao, Ashwin Balakrishna,
  Suraj Nair, Rafael Rafailov, Ethan~P. Foster, Pannag~R. Sanketi, Quan Vuong,
  Thomas Kollar, Benjamin Burchfiel, Russ Tedrake, Dorsa Sadigh, Sergey Levine,
  Percy Liang, and Chelsea Finn.
\newblock {OpenVLA}: An open-source vision-language-action model.
\newblock In \emph{Proceedings of The 8th Conference on Robot Learning}, volume
  270 of \emph{Proceedings of Machine Learning Research}, pp.\  2679--2713.
  PMLR, 2025.

\bibitem[Koo et~al.(2026)Koo, Choi, Kim, Lee, Kim, Seo, and Shin]{hamlet}
Myungkyu Koo, Daewon Choi, Taeyoung Kim, Kyungmin Lee, Changyeon Kim, Younggyo
  Seo, and Jinwoo Shin.
\newblock {HAMLET}: Switch your vision-language-action model into a
  history-aware policy.
\newblock In \emph{The Fourteenth International Conference on Learning
  Representations}, pp.\  101537--101558, 2026.

\bibitem[Lakens(2017)]{lakens2017}
Dani{\"e}l Lakens.
\newblock Equivalence tests: A practical primer for $t$ tests, correlations,
  and meta-analyses.
\newblock \emph{Social Psychological and Personality Science}, 8\penalty0
  (4):\penalty0 355--362, 2017.

\bibitem[Li et~al.(2025)Li, Hsu, Gu, Mees, Pertsch, Walke, Fu, Lunawat, Sieh,
  Kirmani, Levine, Wu, Finn, Su, Vuong, and Xiao]{simplerenv}
Xuanlin Li, Kyle Hsu, Jiayuan Gu, Oier Mees, Karl Pertsch, Homer~Rich Walke,
  Chuyuan Fu, Ishikaa Lunawat, Isabel Sieh, Sean Kirmani, Sergey Levine, Jiajun
  Wu, Chelsea Finn, Hao Su, Quan Vuong, and Ted Xiao.
\newblock Evaluating real-world robot manipulation policies in simulation.
\newblock In \emph{Proceedings of The 8th Conference on Robot Learning}, volume
  270 of \emph{Proceedings of Machine Learning Research}, pp.\  3705--3728.
  PMLR, 2025.

\bibitem[Liao \& Cao(2026)Liao and Cao]{presentnotremembered}
Chih-Ting Liao and Xin Cao.
\newblock Present but not remembered: Auditing how frozen {VLA}s encode,
  deploy, and steer visual history.
\newblock \emph{arXiv preprint arXiv:2607.03372}, 2026.

\bibitem[Lipsitch et~al.(2010)Lipsitch, Tchetgen~Tchetgen, and
  Cohen]{negativecontrols}
Marc Lipsitch, Eric Tchetgen~Tchetgen, and Ted Cohen.
\newblock Negative controls: A tool for detecting confounding and bias in
  observational studies.
\newblock \emph{Epidemiology}, 21\penalty0 (3):\penalty0 383--388, 2010.

\bibitem[Liu et~al.(2023)Liu, Zhu, Gao, Feng, Liu, Zhu, and Stone]{libero}
Bo~Liu, Yifeng Zhu, Chongkai Gao, Yihao Feng, Qiang Liu, Yuke Zhu, and Peter
  Stone.
\newblock {LIBERO}: Benchmarking knowledge transfer for lifelong robot
  learning.
\newblock In \emph{Advances in Neural Information Processing Systems},
  volume~36, pp.\  44776--44791, 2023.

\bibitem[Maharana et~al.(2024)Maharana, Lee, Tulyakov, Bansal, Barbieri, and
  Fang]{locomo}
Adyasha Maharana, Dong-Ho Lee, Sergey Tulyakov, Mohit Bansal, Francesco
  Barbieri, and Yuwei Fang.
\newblock Evaluating very long-term conversational memory of {LLM} agents.
\newblock In \emph{Proceedings of the 62nd Annual Meeting of the Association
  for Computational Linguistics (Volume 1: Long Papers)}, pp.\  13851--13870,
  Bangkok, Thailand, August 2024. Association for Computational Linguistics.

\bibitem[Meng et~al.(2022)Meng, Bau, Andonian, and Belinkov]{rome}
Kevin Meng, David Bau, Alex Andonian, and Yonatan Belinkov.
\newblock Locating and editing factual associations in {GPT}.
\newblock In \emph{Advances in Neural Information Processing Systems},
  volume~35, pp.\  17359--17372, 2022.

\bibitem[Milani~Fard et~al.(2016)Milani~Fard, Cormier, Canini, and
  Gupta]{predictionchurn}
Mahdi Milani~Fard, Quentin Cormier, Kevin Canini, and Maya Gupta.
\newblock Launch and iterate: Reducing prediction churn.
\newblock In \emph{Advances in Neural Information Processing Systems},
  volume~29, 2016.

\bibitem[{NVIDIA} et~al.(2025){NVIDIA}, Bjorck, Casta{\~n}eda, Cherniadev, Da,
  Ding, Fan, Fang, Fox, Hu, Huang, Jang, Jiang, Kautz, Kundalia, Lao, Li, Lin,
  Lin, Liu, Llontop, Magne, Mandlekar, Narayan, Nasiriany, Reed, Tan, Wang,
  Wang, Wang, Wang, Xiang, Xie, Xu, Xu, Ye, Yu, Zhang, Zhang, Zhao, Zheng, and
  Zhu]{groot}
{NVIDIA}, Johan Bjorck, Fernando Casta{\~n}eda, Nikita Cherniadev, Xingye Da,
  Runyu Ding, Linxi Fan, Yu~Fang, Dieter Fox, Fengyuan Hu, Spencer Huang, Joel
  Jang, Zhenyu Jiang, Jan Kautz, Kaushil Kundalia, Lawrence Lao, Zhiqi Li,
  Zongyu Lin, Kevin Lin, Guilin Liu, Edith Llontop, Loic Magne, Ajay Mandlekar,
  Avnish Narayan, Soroush Nasiriany, Scott Reed, {You Liang} Tan, Guanzhi Wang,
  Zu~Wang, Jing Wang, Qi~Wang, Jiannan Xiang, Yuqi Xie, Yinzhen Xu, Zhenjia Xu,
  Seonghyeon Ye, Zhiding Yu, Ao~Zhang, Hao Zhang, Yizhou Zhao, Ruijie Zheng,
  and Yuke Zhu.
\newblock {GR00T N1}: An open foundation model for generalist humanoid robots.
\newblock \emph{arXiv preprint arXiv:2503.14734}, 2025.

\bibitem[{Octo Model Team} et~al.(2024){Octo Model Team}, Ghosh, Walke,
  Pertsch, Black, Mees, Dasari, Hejna, Kreiman, Xu, Luo, Tan, Chen, Sanketi,
  Vuong, Xiao, Sadigh, Finn, and Levine]{octo}
{Octo Model Team}, Dibya Ghosh, Homer Walke, Karl Pertsch, Kevin Black, Oier
  Mees, Sudeep Dasari, Joey Hejna, Tobias Kreiman, Charles Xu, Jianlan Luo,
  {You Liang} Tan, Lawrence~Yunliang Chen, Pannag Sanketi, Quan Vuong, Ted
  Xiao, Dorsa Sadigh, Chelsea Finn, and Sergey Levine.
\newblock Octo: An open-source generalist robot policy.
\newblock In \emph{Proceedings of Robotics: Science and Systems}, Delft,
  Netherlands, July 2024.

\bibitem[Rathi(2026)]{robommeinterference}
Soumil Rathi.
\newblock {RoboMME-Interference}: Benchmarking robot memory under interference.
\newblock \emph{arXiv preprint arXiv:2606.22338}, 2026.

\bibitem[Ross et~al.(2011)Ross, Gordon, and Bagnell]{dagger}
St{\'e}phane Ross, Geoffrey Gordon, and Drew Bagnell.
\newblock A reduction of imitation learning and structured prediction to
  no-regret online learning.
\newblock In \emph{Proceedings of the Fourteenth International Conference on
  Artificial Intelligence and Statistics (AISTATS)}, volume~15 of
  \emph{Proceedings of Machine Learning Research}, pp.\  627--635. PMLR, 2011.

\bibitem[Shi et~al.(2026)Shi, Xie, Liu, Sun, Liu, Wang, Zhou, Fan, Zhang, and
  Huang]{memoryvla}
Hao Shi, Bin Xie, Yingfei Liu, Lin Sun, Fengrong Liu, Tiancai Wang, Erjin Zhou,
  Haoqiang Fan, Xiangyu Zhang, and Gao Huang.
\newblock {MemoryVLA}: Perceptual-cognitive memory in vision-language-action
  models for robotic manipulation.
\newblock In \emph{The Fourteenth International Conference on Learning
  Representations}, pp.\  18567--18602, 2026.

\bibitem[Shi et~al.(2025)Shi, Ichter, Equi, Ke, Pertsch, Vuong, Tanner,
  Walling, Wang, Fusai, Li-Bell, Driess, Groom, Levine, and Finn]{hirobot}
Lucy~Xiaoyang Shi, Brian Ichter, Michael~Robert Equi, Liyiming Ke, Karl
  Pertsch, Quan Vuong, James Tanner, Anna Walling, Haohuan Wang, Niccolo Fusai,
  Adrian Li-Bell, Danny Driess, Lachy Groom, Sergey Levine, and Chelsea Finn.
\newblock Hi robot: Open-ended instruction following with hierarchical
  vision-language-action models.
\newblock In \emph{Proceedings of the 42nd International Conference on Machine
  Learning}, volume 267 of \emph{Proceedings of Machine Learning Research},
  pp.\  54919--54933. PMLR, 2025.

\bibitem[Shinn et~al.(2023)Shinn, Cassano, Gopinath, Narasimhan, and
  Yao]{reflexion}
Noah Shinn, Federico Cassano, Ashwin Gopinath, Karthik Narasimhan, and Shunyu
  Yao.
\newblock Reflexion: Language agents with verbal reinforcement learning.
\newblock In \emph{Advances in Neural Information Processing Systems},
  volume~36, pp.\  8634--8652, 2023.

\bibitem[Sridhar et~al.(2026)Sridhar, Pan, Sharma, and Finn]{memer}
Ajay Sridhar, Jennifer Pan, Satvik Sharma, and Chelsea Finn.
\newblock Scaling up memory for robotic control via experience retrieval.
\newblock In \emph{The Fourteenth International Conference on Learning
  Representations}, pp.\  97142--97166, 2026.

\bibitem[Tango(1998)]{tango1998}
Toshiro Tango.
\newblock Equivalence test and confidence interval for the difference in
  proportions for the paired-sample design.
\newblock \emph{Statistics in Medicine}, 17\penalty0 (8):\penalty0 891--908,
  1998.

\bibitem[Wen et~al.(2020)Wen, Lin, Darrell, Jayaraman, and Gao]{copycat}
Chuan Wen, Jierui Lin, Trevor Darrell, Dinesh Jayaraman, and Yang Gao.
\newblock Fighting copycat agents in behavioral cloning from observation
  histories.
\newblock In \emph{Advances in Neural Information Processing Systems},
  volume~33, pp.\  2564--2575, 2020.

\bibitem[Wu et~al.(2025)Wu, Wang, Yu, Zhang, Chang, and Yu]{longmemeval}
Di~Wu, Hongwei Wang, Wenhao Yu, Yuwei Zhang, Kai-Wei Chang, and Dong Yu.
\newblock {LongMemEval}: Benchmarking chat assistants on long-term interactive
  memory.
\newblock In \emph{International Conference on Learning Representations}, pp.\
  86809--86836, 2025.

\bibitem[Yu et~al.(2026)Yu, Yao, Xie, Tan, Feng, Li, and Wu]{agemem}
Yi~Yu, Liuyi Yao, Yuexiang Xie, Qingquan Tan, Jiaqi Feng, Yaliang Li, and
  Libing Wu.
\newblock Agentic memory: Learning unified long-term and short-term memory
  management for large language model agents.
\newblock In \emph{Proceedings of the 64th Annual Meeting of the Association
  for Computational Linguistics (Volume 1: Long Papers)}, pp.\  21457--21483,
  San Diego, California, United States, July 2026. Association for
  Computational Linguistics.

\bibitem[Zeng et~al.(2026)Zeng, Ye, Chen, Shentu, Wu, Yan, and Li]{kemo}
Yihan Zeng, Minghao Ye, Yiyuan Chen, Yide Shentu, Philipp Wu, Zike Yan, and
  Zhongyu Li.
\newblock {KEMO}: Event-driven keyframe memory for long-horizon robot
  manipulation with {VLA} policies.
\newblock \emph{arXiv preprint arXiv:2606.23589}, 2026.

\bibitem[Zhang \& Nanda(2024)Zhang and Nanda]{zhangnanda}
Fred Zhang and Neel Nanda.
\newblock Towards best practices of activation patching in language models:
  Metrics and methods.
\newblock In \emph{The Twelfth International Conference on Learning
  Representations}, pp.\  1651--1678, 2024.

\bibitem[Zhang et~al.(2026{\natexlab{a}})Zhang, Xu, Dhafer, Yue, Dong, and
  Hao]{embodiedinterpretability}
Hanxin Zhang, Mingshuo Xu, Abdulqader Dhafer, Shigang Yue, Hongbiao Dong, and
  Zhou~Daniel Hao.
\newblock Embodied interpretability: Linking causal understanding to
  generalization in vision-language-action models.
\newblock In \emph{Proceedings of the 43rd International Conference on Machine
  Learning}, 2026{\natexlab{a}}.

\bibitem[Zhang et~al.(2026{\natexlab{b}})Zhang, Zhang, Gao, Li, Liu, Nie, Zhu,
  Qiu, Yan, Liu, Tang, Rao, Fang, Wei, Wang, Ding, and Yu]{harnessvla}
Yixian Zhang, Huanming Zhang, Feng Gao, Xiao Li, Zhihao Liu, Yi~Nie, Chunyang
  Zhu, Jiaxing Qiu, Yuchen Yan, Jiyuan Liu, Wenhao Tang, Jiaji Rao, Zhengru
  Fang, Changxu Wei, Yu~Wang, Wenbo Ding, and Chao Yu.
\newblock Harness {VLA}: Steering frozen {VLA}s into reliable manipulation
  primitives via memory-guided agents.
\newblock \emph{arXiv preprint arXiv:2607.08448v5}, 2026{\natexlab{b}}.

\bibitem[Zhao et~al.(2024)Zhao, Huang, Xu, Lin, Liu, and Huang]{expel}
Andrew Zhao, Daniel Huang, Quentin Xu, Matthieu Lin, Yong-Jin Liu, and Gao
  Huang.
\newblock {ExpeL}: {LLM} agents are experiential learners.
\newblock \emph{Proceedings of the AAAI Conference on Artificial Intelligence},
  38\penalty0 (17):\penalty0 19632--19642, March 2024.

\bibitem[Zitkovich et~al.(2023)Zitkovich, Yu, Xu, Xu, Xiao, Xia, Wu, Wohlhart,
  Welker, Wahid, Vuong, Vanhoucke, Tran, Soricut, Singh, Singh, Sermanet,
  Sanketi, Salazar, Ryoo, Reymann, Rao, Pertsch, Mordatch, Michalewski, Lu,
  Levine, Lee, Lee, Leal, Kuang, Kalashnikov, Julian, Joshi, Irpan, Ichter,
  Hsu, Herzog, Hausman, Gopalakrishnan, Fu, Florence, Finn, Dubey, Driess,
  Ding, Choromanski, Chen, Chebotar, Carbajal, Brown, Brohan, Arenas, and
  Han]{rt2}
Brianna Zitkovich, Tianhe Yu, Sichun Xu, Peng Xu, Ted Xiao, Fei Xia, Jialin Wu,
  Paul Wohlhart, Stefan Welker, Ayzaan Wahid, Quan Vuong, Vincent Vanhoucke,
  Huong Tran, Radu Soricut, Anikait Singh, Jaspiar Singh, Pierre Sermanet,
  Pannag~R. Sanketi, Grecia Salazar, Michael~S. Ryoo, Krista Reymann, Kanishka
  Rao, Karl Pertsch, Igor Mordatch, Henryk Michalewski, Yao Lu, Sergey Levine,
  Lisa Lee, Tsang-Wei~Edward Lee, Isabel Leal, Yuheng Kuang, Dmitry
  Kalashnikov, Ryan Julian, Nikhil~J. Joshi, Alex Irpan, Brian Ichter, Jasmine
  Hsu, Alexander Herzog, Karol Hausman, Keerthana Gopalakrishnan, Chuyuan Fu,
  Pete Florence, Chelsea Finn, Kumar~Avinava Dubey, Danny Driess, Tianli Ding,
  Krzysztof~Marcin Choromanski, Xi~Chen, Yevgen Chebotar, Justice Carbajal,
  Noah Brown, Anthony Brohan, Montserrat~Gonzalez Arenas, and Kehang Han.
\newblock {RT-2}: Vision-language-action models transfer web knowledge to
  robotic control.
\newblock In \emph{Proceedings of The 7th Conference on Robot Learning}, volume
  229 of \emph{Proceedings of Machine Learning Research}, pp.\  2165--2183.
  PMLR, 2023.

\end{thebibliography}
\bibliographystyle{iclr2027_conference}

\clearpage
\appendix
\startcontents[appendix]
\section*{Appendix}
{\setlength{\parskip}{0pt}\printcontents[appendix]{}{1}{\setcounter{tocdepth}{2}}}
\clearpage

\section{Protocol Details, Conventions and Provenance}
\label{app:scope}

\textbf{Earlier terms.} Frozen protocols and run records use an earlier framing.
There, ``Step~0'' is the detectability precondition (Section~\ref{sec:step0}); RQ1
and RQ2 are the conditions of Q2, that the deciding information is only in
memory and that it is delivered and valid at the decision
(Section~\ref{sec:conditions}); RQ3 is Q3 and Q4 (Section~\ref{sec:sensitivity}); and RQ4
is Q1 (Section~\ref{sec:net-benefit}). Campaign names such as Q2-XA and Q3-VC predate
the four questions and do not refer to them.

In the appendix, E1B denotes the 202-chain ChainBench study ($K=2$), and E1
denotes the supporting 143-chain study ($K=3$). These identifiers are
retained to match the archived experimental records.

The appendix expands the experiments summarized in the main text and records
the negative, withdrawn and superseded studies that affected our conclusions.
It does not catalog every exploratory run. The verdicts on the four
questions apply criteria formulated after the campaigns closed
(Appendix~\ref{app:amendments}).
Descriptive and post hoc analyses are labeled as such, and excluded or
withdrawn studies support no paper claim (Appendix~\ref{app:excluded}).
\textbf{Scope.} Transfer of the fixed-path calibration beyond Remember Color is
untested, and the one-ULP discordance rate may be specific to this simulator
medium and to the implementation that produced it.

\subsection{Evidence Ledger by Question}
\label{app:ledger}

\begin{table}[!ht]
\centering
\scriptsize
\setlength{\tabcolsep}{3pt}
\caption{\textbf{Evidence ledger, ordered by question} (detectability, then Q2
(only in memory), Q2 (delivery/validity), Q3/Q4 and Q1; criteria in
Sections~\ref{sec:net-benefit} and~\ref{sec:conditions}). The status column uses the verdict words of
Appendix~\ref{app:statistics}. Withdrawn and excluded rows support no claim.
``App.'' is the section that reports the row. $^{a}$The scripted reader is both
the detectability control for the decision readout and the Q4 evidence for curated records,
so these rows are not independent evidence. $^{b}$These analyses are post hoc. No survival gate
was registered for these campaigns. $^{c}$Threshold gate against the default
token merge, 250 seed-paired units, attempt~1: $b{=}3$, $c{=}1$ (exact McNemar
$p=0.625$). Remember Color~5 is near chance (Appendix~\ref{app:rc-floors}). Whether the rule acts within or across attempts was not verified. The study has no closeout record, although its sealed root holds the
frozen design note and the launch record.}
\label{tab:app-ledger}\label{tab:link-profile}
\vspace{2pt}
\begin{tabular}{@{}>{\raggedright\arraybackslash}m{6.9cm}>{\raggedright\arraybackslash}m{5.6cm}l@{}}
\toprule
Study & Status & App. \\
\midrule
\multicolumn{3}{@{}l}{\emph{Detectability: what effect can each readout detect?}} \\
Post-cue view removal (Shell Game, released banks) & preregistered; holds ($D=+20.5\pp$) & \ref{app:viewremoval} \\
Scripted-reader positive control (ChainBench)$^{a}$ & holds & \ref{app:poscal} \\
Graded synthetic wrapper (R-1, RC3/RC5) & holds (synthetic post-policy injection) & \ref{app:r1ladder} \\
Grid resolution (GRID-1a, SimplerEnv-Bridge) & preregistered; unresolved (below announceable effect) & \ref{app:resolution} \\
Repeat checks (exact replicates) & holds (determinism only) & \ref{app:executor-floors} \\
Instruction manipulation check (E1CL) & check only; replay floor $\approx27\%$ & \ref{app:executor-floors} \\
\midrule
\multicolumn{3}{@{}l}{\emph{Q2 (only in memory): is the deciding information only in memory?}} \\
ChainBench v1 retrieval diagnostic & holds & \ref{app:chainbench} \\
E1B no-memory accuracy against $1/K$ (all $K{=}2$) & holds (floor from post-closeout check) & \ref{app:chainbench} \\
Remember Color retry semantics & fails (cue re-shown; near chance) & \ref{app:rc-floors} \\
Shell Game success rule; E0 screen & fails (any-touch success rule) & \ref{app:sg-anytouch} \\
Bridge carry chains & fails (by construction: position rotates) & \ref{app:learned-outcomes} \\
Fractal; Bridge grid & not measured & --- \\
\midrule
\multicolumn{3}{@{}l}{\emph{Q2 (delivery/validity): is it still there at the decision?}} \\
Decision-time survival audit (RC; SG full vision; plugin)$^{b}$ & fails (coverage 0.51--0.83, a lower bound) & \ref{app:survival} \\
Blackout-ladder delivery & holds (2,400/2,400 each) & \ref{app:survival} \\
Read-path gates (Q2-XA, E2-W0) & holds & \ref{app:readpath} \\
Version-conflict configuration contrast (RC3/5/9) & harm criterion met; validity unresolved & \ref{app:version} \\
Persistence-only replication & not replicated (power $\approx.42$) & \ref{app:version} \\
Threshold-gated retention (MIKASA RC5)$^{c}$ & preregistered; not detected & \ref{app:ledger} \\
Writer closed loop & retrospective audit; supporting only & \ref{app:writer-audit} \\
\midrule
\multicolumn{3}{@{}l}{\emph{Q3/Q4: does behavior follow the task-relevant content?}} \\
Shell Game blackout content ladders & not detected; sham criterion corrected after data & \ref{app:blackout-ladders} \\
Fixed-path matched-control study (RC3/RC5) & not detected beyond $\pm7.5\pp$, as delivered & \ref{app:payload} \\
Trained-plugin sham; offline gate & sham unresolved; offline gate passed & \ref{app:plugin-sham} \\
Payload footprints (Fractal, Bridge) & bounds only; specificity not measured & \ref{app:footprint} \\
Coordinate de-prescription & unresolved & \ref{app:coordinate} \\
ChainBench shuffled records & holds (outcome sensitivity) & \ref{app:e1b} \\
Scripted-reader content ladder (ChainBench)$^{a}$ & holds (fact specificity for this readout) & \ref{app:poscal} \\
\midrule
\multicolumn{3}{@{}l}{\emph{Q1: does success change, and where is the bottleneck?}} \\
ChainBench E1B (public \pihalf{} executor) & preregistered; holds (execution-limited) & \ref{app:e1b} \\
ChainBench E1 (rebuilt manifest) & supporting; lineage-limited & \ref{app:e1b} \\
Perceptual-bank ablation (Fractal) & preregistered; not detected & \ref{app:learned-outcomes} \\
Cross-episode carry (Bridge) & preregistered; not detected & \ref{app:learned-outcomes} \\
Released vs.\ structured PCMB (MIKASA, 5 tasks) & not detected (configuration contrast) & \ref{app:mikasa} \\
Trained memory plugin ($2\times2$ factorial) & preregistered; fails (success lower) & \ref{app:trained} \\
\midrule
\multicolumn{3}{@{}l}{\emph{Withdrawn and excluded}} \\
Shell Game full-vision $C/U/S/T$ ladder & withdrawn & \ref{app:withdrawn} \\
Twin testbed, WriterBench-X, pilots & excluded; no paper claim & \ref{app:twins} \\
\bottomrule
\end{tabular}
\end{table}

\subsection{Paired Estimands, Bounds and Verdict Words}
\label{app:statistics}

\textbf{Terminology.} An \emph{episode} is one simulator rollout. An
\emph{attempt} is an episode at a fixed position in a repeated-task sequence. A
\emph{chain} is the ordered sequence that may carry memory across attempts. A
\emph{cell} is one policy $\times$ task $\times$ memory interface at $n$
seed-paired episodes. We use \emph{preregistered} to mean fixed in a written
protocol before the corresponding outcomes were collected.

\textbf{Sign conventions.} Absolute differences between rates and their bounds are reported in percentage points (pp), while rates are percentages or proportions as labeled. Besides $D=(b-c)/n$ of Section~\ref{sec:net-benefit}, whose
reference is the memory-on condition $A$, Appendices~\ref{app:version}
and~\ref{app:rq4} report $\Delta=(c-b)/n$, a named condition minus a named
baseline, with $b$ counting pairs in which that baseline succeeds, and declare
it at each use. $\Delta=-D$ when the baseline is the memory-on condition (Fractal);
when it is the memory-off condition (ChainBench E1/E1B, Bridge carry, the plugin bank
channel, the persistence-only $2\times2$), $b$ and $c$ are swapped relative
to Section~\ref{sec:net-benefit} and $\Delta=D$. We never subtract across the two
conventions. Where a $\Delta$-convention number is quoted, both
conditions are named. For example, the version-conflict contrast is the
structured-PCMB condition minus its released comparator, $-3.91\pp$
(Appendix~\ref{app:version}).

\textbf{What the content estimand does and does not measure.} We
operationalize a memory contribution as the target-fact-specific increment
under a retrieval and computation path fixed at installation. By construction
$\bar\theta$ misses diffuse benefits that bypass the audited fact and effects
that change retrieval, gating or consolidation timing: fact specificity is the
unit of our attribution question, not a complete account of memory value.

\textbf{Tests and bounds.} Unless stated otherwise, $p$-values are two-sided
exact McNemar tests conditional on the $b+c$ discordant pairs, with Holm
correction for families frozen in advance. Campaign decision bounds are built
from exact Clopper--Pearson bounds on the two discordance proportions,
$\mathrm{UB}=\mathrm{CP}_{+}(b;\alpha)-\mathrm{CP}_{-}(c;\alpha)$. Each
\emph{component} bound is one-sided $95\%$ ($\alpha=0.05$), but a bound on the
\emph{difference} spends one tail on each proportion, so by Bonferroni it has
coverage at least $90\%$, and a two-sided interval formed from two such bounds
spends four tails and has coverage at least $80\%$. We label each bound at the
coverage of the quantity being bounded, never at the coverage of a component.
One design uses a different setting and says so where it is reported: the
Bridge grid (GRID-1a) uses $\alpha=0.025$ per component, so each one-sided
endpoint has coverage at least $95\%$ and the two-sided interval at least
$90\%$. It is not a joint $95\%$ interval. Campaign decision bounds retain
the $\alpha$ used by their frozen adjudicators.

\textbf{Additional intervals in Table~\ref{tab:net-benefit}.} For seven
comparisons, we additionally compute post hoc two-sided intervals from the
reported paired counts, using $\alpha=0.025$ for each Clopper--Pearson component
tail. If $L_b,U_b$ and $L_c,U_c$ are the component limits based on $b$ and $c$
among $n$ paired units, the endpoints are $100(L_b-U_c)$ and $100(U_b-L_c)$ percentage points. Under the
paired-unit binomial model, the four-tail Bonferroni construction gives coverage
of at least $90\%$ for each contrast, not simultaneous coverage across the table.
Bridge grid retains its previously reported $90\%$ interval; rows marked
$\dagger$ retain their reported $95\%$ intervals. These additional summaries do
not replace the original tests, decision bounds, or verdicts.

Other interval methods are specified with their study results: the Tango paired-binomial score
interval \citep{tango1998}, the task-stratified paired bootstrap, the chain
bootstrap and, on the blackout ladder, a conditional exact bound. For small
discordance counts we also report the exact bound, because a paired bootstrap
over episodes is anti-conservative exactly where the decision is made: with
$b=c=0$ it collapses to $[0,0]$, where the exact bound from a one-sided $95\%$
component at $n=200$ is $1.49\pp$.

\textbf{Verdict words.} For the bound-based magnitude decisions described here,
\emph{not detected} means that the bound excludes the registered criterion but
no equivalence test was frozen, and it is always stated with its bound.
\emph{Unresolved} means that the interval straddles the registered criterion,
or that the registered contrasts disagree at insufficient power.

\textbf{Legacy gates.} The frozen protocols name gates and readouts that the
paper now reports under the detectability precondition and the four questions,
as mapped in Table~\ref{tab:legacy-gates}. A gate that passed does not by itself
license a verdict on any other question.

\begin{table}[ht]
\centering
\scriptsize
\setlength{\tabcolsep}{3pt}
\caption{\textbf{Legacy gates and readouts.} Each gate or readout named in the
frozen protocols, the requirement it froze, and the question under which the
paper now reports it.}
\label{tab:legacy-gates}
\vspace{4pt}
\begin{tabular}{@{}>{\raggedright\arraybackslash}m{2.4cm}>{\raggedright\arraybackslash}m{6.7cm}>{\raggedright\arraybackslash}m{4.4cm}@{}}
\toprule
Legacy name & Frozen requirement & Question under which it is now reported \\
\midrule
wiring gate & $C$ matches $A$ bitwise; two adjudicators agree; cross-machine recompute & Detectability: exact replicate $C$ (App.~\ref{app:executor-floors}) \\
path gate & probe: $A$ reads memory; $N$'s forced-empty reads equal its reads & Q2 (delivery): forced reads over reads (App.~\ref{app:readpath}) \\
competence gate & Wilson $95\%$ interval of $\mathrm{SR}(A)$ inside $(\text{floor}+0.05,\,1-0.05)$ & Q2 (only in memory) floor and ceiling (App.~\ref{app:rc-floors}) \\
survival gate ($G_{\rm surv}$; blackout ladders only) & substituted content present at the decision step in every substituting-condition trial & Q2 delivery (App.~\ref{app:survival}); elsewhere post hoc coverage \\
\midrule
utility readout & $D$ against a criterion registered before launch & Q1 (Section~\ref{sec:rq4}) \\
sensitivity readout & $\rho$ vs.\ frozen 5\% activity threshold; $U$ as reference & Q3 numerical control $\rho_U$ (Section~\ref{sec:rq3}) \\
specificity readout & $\bar\theta$ against the preregistered $\pm7.5\pp$ equivalence band & Q4 content estimand (Section~\ref{sec:rq3}) \\
\bottomrule
\end{tabular}
\end{table}

\noindent Three rules of the frozen protocols still apply. (i) A cell failing
competence is \emph{excluded}; one failing path or wiring is \emph{invalid},
and its number is withheld. (ii) Verdicts route on the bound, never on the
point estimate, against a criterion registered before launch. (iii) The
prescribed outer sequence for a formal campaign is precommit, probe, smoke,
formal run, seal, adjudication, cross-check and closeout, with criteria frozen
at the precommit, failed runs renamed rather than deleted, and adjudication
triggered by a session other than the one that ran the conditions
(Appendix~\ref{app:adjudication}).

\subsection{Checkpoint, Implementation and Campaign Binding}
\label{app:checkpoint-provenance}

The three benchmark families use distinct public community assets, robot
embodiments and action interfaces. Thus, ``the community checkpoint'' does not
refer to the same asset across families. Table~\ref{tab:checkpoint-lineage} binds each
paper-facing family to the exact evaluated asset. Within MIKASA, one checkpoint
serves every Shell Game and Remember Color campaign: the checkpoint checksum
recorded in the view-removal, released blackout-ladder, fixed-path,
full-vision-ladder and plugin-factorial roots is the same value. The same
weights are therefore competent on Shell Game and near chance on Remember Color
(Appendix~\ref{app:rc-floors}). The MIKASA adjacent configuration records
OpenVLA-7B initialization, the official MIKASA five-task training mix (1,250
trajectories; 90,000 transitions), seed 42, and a 20,000-step maximum. Its
memory settings are 16 slots, ToMe consolidation, retrieval aggregation, and
gated fusion. None of these assets contains a complete optimizer-state or
checkpoint-selection history, so we do not infer more than configuration-bound
facts.

The evaluating implementation is not a cross-family identity either, and
within a family it is bound per campaign. The MIKASA campaigns whose records
bind it run an extended $3322$-line MemoryVLA implementation
(the MIKASA build) under the released configuration. The
SimplerEnv-Fractal perceptual-bank results were produced by a $3251$-line
overlay lineage (the Fractal overlay) whose identity at the time of the run
cannot be established from the surviving records
(Appendix~\ref{app:learned-outcomes}). The $903$-line sealed release file
is the implementation of the Bridge grid: that
campaign's frozen lineage gate checked this checksum in both conditions and passed,
and no other number in this paper comes from it. The sealed manifest of the
Bridge five-attempt carry study checksums an overlay MemoryVLA implementation
(the carry overlay) in its source snapshot. That snapshot is not part of
the archived evidence unit, so the file's line count and its relation to the
other lineages are not established. Whether these lineages are behaviorally
equivalent under the released configuration is not established, and no result
is transferred across them. No surviving record binds the implementation that
ran the trained-plugin closed loop and factorial (Appendix~\ref{app:trained})
or the retention head-to-head on RC5 (Table~\ref{tab:app-ledger}). Both
identities are disclosed as not established. The plugin's RC5 sham is the
exception (Table~\ref{tab:campaign-binding}).

\begin{table}[!ht]
\centering
\scriptsize
\setlength{\tabcolsep}{3pt}%
\caption{\textbf{Per-family checkpoint lineage.} Checkpoint files, action interfaces and
implementation versions. SHA-256 checksums of all weights, implementations and
asset snapshots are listed in the archived manifests.
$^{\ast\dagger\ddagger}$See the text.}
\label{tab:checkpoint-lineage}
\vspace{2pt}
\begin{tabular}{@{}>{\raggedright\arraybackslash}m{1.9cm}>{\raggedright\arraybackslash}m{3.9cm}>{\raggedright\arraybackslash}m{3.1cm}>{\raggedright\arraybackslash}m{4.4cm}@{}}
\toprule
Benchmark family & checkpoint & embodiment; action interface & implementation / lines \\
\midrule
SimplerEnv--Fractal & MemoryVLA Fractal checkpoints for visual matching and variant aggregation & Google Robot (static); 7-D EEF delta & Fractal overlay / 3251$^{\ast}$ \\
SimplerEnv--Bridge & MemoryVLA Bridge checkpoint & WidowX, sink camera; 7-D EEF delta & grid: sealed release / 903;\newline carry: carry overlay / not established$^{\dagger}$ \\
MIKASA--Robo & MemoryVLA MIKASA checkpoint & ManiSkill Panda, wrist camera; 7-D end-effector delta pose & MIKASA build / 3322$^{\ddagger}$ \\
\bottomrule
\end{tabular}
\end{table}

\textbf{Campaign, method and evaluator.} On MIKASA the retry method sets both
the memory architecture and persistence: released MemoryVLA runs the
two episode-local banks and passes no scope to the method, and every other
method runs structured PCMB, so a contrast between a released and a structured
condition is a configuration contrast, not an ablation. Only the record-retaining
methods, persistent PCMB and persistent PCMB with execution memory, carry a scope's
records into attempt~2; the two no-cross-episode variants delete the scope
before attempt~2, and the released method's reset clears both of its banks. The
MIKASA paired-retry evaluator exists in several
revisions, labeled $e_1$--$e_4$ where recorded, so absolute rates are never compared across campaigns
(Table~\ref{tab:campaign-binding}).

\begin{table}[!ht]
\centering
\scriptsize
\setlength{\tabcolsep}{3pt}
\caption{\textbf{Campaign binding.} Method, evaluator revision and MemoryVLA
implementation version carried by each campaign's archived records. $e_1$--$e_4$ label distinct
evaluator checksums without implying order, and implementation names follow
Table~\ref{tab:checkpoint-lineage}. ``Not recorded'' means the archived evidence
unit carries no such checksum. Full checksums are in the archived manifests.}
\label{tab:campaign-binding}
\vspace{2pt}
\begin{tabular}{@{}>{\raggedright\arraybackslash}m{3.7cm}>{\raggedright\arraybackslash}m{4.9cm}>{\raggedright\arraybackslash}m{4.9cm}@{}}
\toprule
Setting (Section~\ref{sec:settings}) & method & evaluator / implementation \\
\midrule
\multicolumn{3}{@{}l}{\emph{Within-episode learned banks}} \\
Shell Game view removal & released MemoryVLA & $e_1$ / MIKASA build \\
Shell Game read, Q2-XA & released MemoryVLA & not recorded \\
SG-blackout ladder, released & released MemoryVLA & ladder runner / MIKASA build$^{\ast}$ \\
SG-blackout ladder, structured & episode-local PCMB & ladder runner / MIKASA build$^{\ast}$ \\
Fractal (completion and eight-cell runs) & perceptual bank ablated & Fractal overlay, dated after the run \\
Bridge grid (GRID-1a) & released MemoryVLA, read on/off & sealed release \\
\midrule
\multicolumn{3}{@{}l}{\emph{Cross-episode learned banks}} \\
SG-attempts: E0 screen & structured PCMB & not recorded \\
SG-attempts: E2-W0 path ladder & structured PCMB & not recorded \\
SG-attempts: full vision (withdrawn) & structured PCMB & $e_1$ / MIKASA build \\
RC-swap fixed path & persistent PCMB & $e_2$ / MIKASA build in snapshot, not preregistered \\
R-1 graded wrapper & persistent PCMB & $e_1$ / MIKASA build \\
RC-VC conflict-on & persistent PCMB vs.\ released MemoryVLA & $e_3$ (checksum only) / not recorded \\
RC-VC no-conflict twin & same two methods & $e_1$ / not recorded \\
RC-VC persistence-only $2\times2$ & persistent PCMB, four conditions & $e_1$ / MIKASA build \\
Five-task configuration contrast & released MemoryVLA vs.\ persistent PCMB (with execution memory identical) & $e_4$ / not recorded \\
Retention head-to-head, RC5 & token merge (default) vs.\ threshold gate & not recorded \\
Bridge carry & fresh, token-merge, structured and hybrid carry & carry overlay (manifest only) \\
Plugin closed loop & trained memory components & not recorded \\
Plugin factorial and RC5 sham & recipe $\times$ bank; sham conditions on the trained checkpoint & sham: $e_1$ / MIKASA build; factorial: not recorded \\
\bottomrule
\multicolumn{3}{@{}p{13.5cm}@{}}{$^{\ast}$Stated in the structured ladder's preregistration as shared by both ladders; not re-verified from run records. ChainBench runs on LIBERO with the \pihalf{} executor and uses none of these files (Appendix~\ref{app:chainbench}).}
\end{tabular}
\end{table}

Every evaluator revision whose source survives contains the allowlist overlap
check and the two-ended reset (Appendix~\ref{app:payload}), and the scope and
episode hash functions are byte-identical across the protocol revisions. For
the same-architecture persistence-only replication and the $2\times2$ Shell
Game blackout study, sampled scope and episode identifiers recompute from
(task, seed) and (scope, attempt), and sampled persistent-condition records reach
attempt~2 but not the next scope. Evaluator revision $e_3$
of the conflict-on version-merge campaign (Appendix~\ref{app:version}) survives
only as a checksum, so these properties are not verified for that run.

\textbf{Bridge reproduction.} MemoryVLA reports 71.9\% aggregate
SimplerEnv-Bridge success \citep{memoryvla}. With our evaluation code, the released
checkpoint reaches $71/96=73.96\%$ aggregate, with a per-task spread that
does not match the published per-task numbers (cube $7/24=29.2\%$ against a
published $37.5\%$). Our 29--31\% Bridge attempt-1 rates come from a different
repeated-chain construction and are not a reproduction or zero-shot comparison
of that published benchmark result.

\section{Detectability, Power and Resolution}
\label{app:power}

For each readout of the detectability precondition (Section~\ref{sec:step0}), this chapter records an
in-design control showing that the readout can move, the smallest effect each
design could have announced, and the repeat checks that set its noise floor.

\subsection{Shell Game View Removal and Read Contrasts}
\label{app:viewremoval}

\textbf{Design.} R27-A is a preregistered $2\times2$ study on MIKASA Shell
Game. It crosses the current view (visible, or a camera blackout from elapsed
step~6 to the end of the episode, 85 of 90 frames) with the episode-local
memory reads (both reads on, or both blocked). Each condition has $n=200$
seed-paired scopes on seeds 8701--8900, a band disjoint from every other band
in the seed registry. The run uses the released configuration (checkpoint and
implementation: Appendix~\ref{app:checkpoint-provenance}). In every condition both
episode-local banks were empty at the first memory query of each attempt for
all 200 seeds, and under blackout every blocked read was counted ($4{,}600$
cognitive and $4{,}600$ perceptual calls). After elapsed step~5 the cups are
closed and identical, so the withheld view carries no cue information. Any
touch of the correct cup counts as success (Appendix~\ref{app:sg-anytouch}).
Attempt~1 is the registered estimand. Attempt~2 is recorded and enters no
criterion. Within one process, it is not a bitwise replay of attempt~1.
Under blackout with reads on, first-observation digests and first-query bank sizes
match in 200/200 seeds, yet action-sequence digests differ in 200/200.
The cause is not established.

\begin{table}[ht]
\centering
\scriptsize
\setlength{\tabcolsep}{4pt}
\caption{\textbf{Shell Game view removal and read contrasts, attempt~1.}
$n=200$ seed-paired scopes per condition. SR base and SR int.\ are success without
and with the intervention, and $D=(b-c)/n$ is positive when the intervention
lowers success. LB and UB are exact bounds on $D$, and $p$ values are exact
McNemar tests. The interaction lies below the minimum detectable effect (MDE)
and supports no claim.}
\label{tab:app-viewremoval}
\vspace{4pt}
\begin{tabular}{@{}lrrcrrr@{}}
\toprule
Contrast (attempt~1) & SR base & SR int. & $b/c$ & $D$ (pp) & bound & McNemar $p$ \\
\midrule
view removed, reads on & 0.940 & 0.735 & 45/4 & $+20.5$ & LB $+13.21$ & $8.2\times10^{-10}$ \\
reads blocked, view visible & 0.940 & 0.935 & 3/2 & $+0.5$ & UB $3.65$ & $1.0$ \\
reads blocked, under blackout & 0.735 & 0.695 & 9/1 & $+4.0$ & UB $7.70$ & $0.0215$ \\
interaction (row 3 $-$ row 2) & & & & $+3.5$ & below MDE $5.78$ & --- \\
\bottomrule
\end{tabular}
\end{table}

\noindent In Table~\ref{tab:app-viewremoval}, conventions follow
Appendix~\ref{app:statistics}: $D=(b-c)/n$, so
positive $D$ means the intervention lowers success. The $p$ values come from exact
McNemar tests. We use exact Clopper--Pearson bounds on the two discordance
proportions, one-sided 95\% per component, so each bound on the difference has
coverage $\geq90\%$. Wilson 95\% intervals for attempt-1 success are
$[0.898,0.965]$ (view visible, reads on), $[0.892,0.962]$ (view visible, reads
blocked), $[0.670,0.791]$ (blackout, reads on) and $[0.628,0.755]$ (blackout,
reads blocked).

\textbf{Registered rule and verdicts.} The frozen rule required a lower bound
for view removal above the registered $10\pp$ criterion, an upper bound for the
reads under blackout below that criterion, and blackout success above the
analytic floor of $1/3$. All three were met ($13.21\pp>10\pp$,
$7.70\pp<10\pp$, Wilson lower bound $0.670>1/3$). Removing the post-cue view
lowers success (holds). Under blackout, the gain from enabling memory reads falls below the
registered 10-pp criterion, although the paired success difference is
statistically detectable.
Effects smaller than $7.70\pp$ are not excluded ($b{=}9$, $c{=}1$), and no
claim is made that the reads do nothing. With the view visible, blocking the
reads changes success by $+0.5\pp$ (UB $3.65\pp$): not detected. The
interaction is $+3.5\pp$. The frozen MDE table gives $5.78\pp$ at the observed
non-zero rate of $15/200$, so the interaction supports no claim. $D=+20.5\pp$
measures how much execution needs post-cue visual input, together with any cue
the policy has externalized into its own pose. This design does not separate
those two channels or locate the cue. First-visit correctness was
not computed in this run.

\textbf{Scope.} We evaluated one backbone on one task using a fixed checkpoint
under the released configuration. The interventions were applied at inference
time without retraining. Only within-episode history is removed. These banks
are distinct from the cross-episode persistent records of the structured-PCMB
studies (Appendix~\ref{app:readpath}). The blackout is an unnatural,
out-of-distribution perturbation, so the results do not transfer to partial or
textured occlusions, other onsets or other tasks. Compensation under blackout
would not imply that reading the banks changes success when the view is
available. Bank contents and the write path are not instrumented, so the
study says nothing about what the banks store.

\subsection{Scripted-Reader Positive Control}
\label{app:poscal}

The $C/U/S/T$ ladder (Section~\ref{sec:arms}) was applied to the ChainBench v3
retrieval reader: curated records read by a fixed script (leave-one-chain-out
nearest-centroid selection over pre-LM patch features, a symmetric
maximum-similarity matcher, presence gating and two-stage selection), with
zero decision-time LLM calls. Centroids and the selection path are identical
across conditions; only the evaluated event feature changes. On the full 401-chain
decision cohort, $C$ reproduces every archived decision (401/401 rows identical
to the sealed full-system rows). $U$, which perturbs every element by one ULP,
changes $0/401$ decisions, as does the same-fact cross-chain $S$. The
different-fact $T$ changes $401/401$ ($b{=}401$, $c{=}0$). Hence
$\rho_T-\rho_S=1.00$, with a chain-level paired-bootstrap 95\% CI of
$[1.00,1.00]$. This degenerate interval reflects unanimous changes in this
cohort and does not imply zero population uncertainty. The exact Clopper--Pearson 95\% interval for $\rho_U$ is
$[0,0.92\%]$. The preregistered 202-chain E1B subset gives the same rates
($0/202$, $0/202$, $202/202$), and a second implementation reproduces every
condition decision. The same measurement serves as the detectability control for the
decision readout and as the Q4 evidence for CB-curated (Section~\ref{sec:rq3},
Table~\ref{tab:link-profile}). The two entries are not independent evidence.

This result calibrates the ladder on the decision readout only. It is not a
robot closed-loop result, and it is not a positive control on learned-bank
content. No such control exists in this paper (Section~\ref{sec:limitations}). The
earlier 143-chain E1 lineage uses an evaluator association table of a
different type, so the reader result is not described as replicated there.

\subsection{Graded Synthetic Detectability Ladder}
\label{app:r1ladder}

The preregistered R-1 wrapper supplies the detectability entry for the discordance
readout on Remember Color (Table~\ref{tab:link-profile}). On RC3/RC5 it
injects a deterministic, fact-conditioned action difference after policy
prediction and before the action queue, in attempt~2 only. Attempt~1 receives
zero injections. The injection gate found zero table mismatches and zero
mixed-injection episodes. The wrapper ran 15{,}200 rollouts over 38 units,
reusing the sealed substitution band's seeds whole, with no per-seed
selection. At each amplitude the common cohort is $n_{\mathrm{joint}}=388$,
and $\theta=\rho_T-\rho_S$ is bounded by a chain-level paired bootstrap
(10{,}000 replicates, seed 20260824; Table~\ref{tab:app-r1-ladder}).

\begin{table}[ht]
\centering
\small
\caption{\textbf{Graded detectability ladder.} R-1 wrapper on the common
$n_{\mathrm{joint}}=388$ cohort at each injected amplitude $\delta$. LB is the
one-sided 95\% lower bound on $\theta$ (chain-level paired bootstrap).
$\delta^*$ is the smallest detected amplitude.}
\label{tab:app-r1-ladder}
\vspace{4pt}
\begin{tabular}{lrrl}
\toprule
$\delta$ ($L_2$/step) & $\theta=\rho_T-\rho_S$ & One-sided 95\% LB & Verdict \\
\midrule
$10^{-4}$ & $-2.6\pp$ & $-5.9\pp$ & not detected \\
$10^{-3}$ & $+2.3\pp$ & $-0.8\pp$ & not detected \\
$10^{-2}$ & $+2.1\pp$ & $-1.6\pp$ & not detected \\
$2\!\times\!10^{-2}$ & $+7.2\pp$ & $+3.4\pp$ & detected ($\delta^*$) \\
$5\!\times\!10^{-2}$ & $+8.8\pp$ & $+4.9\pp$ & detected \\
\bottomrule
\end{tabular}
\end{table}

The lower three levels are not detected. The frozen MDE analysis gives
insufficient power below $5\pp$, so these levels are not evidence of absence.
At $\delta=0.02$, a direction not conditioned on the fact gives
$\theta=0.0\pp$ (90\% chain-level paired-bootstrap CI $[-3.6,+3.6]\pp$), so
amplitude alone does not create a $\theta$ increment. The replicate anchor
changes $0/400$ outcomes, completion is at least $0.97$ in every unit, and two
implementations agree. Because the injection happens after the policy, this is
an outcome-detectability result only: it does not test whether content
survives in the bank until the decision (Q2, delivery), and it is not a resolution
guarantee for the Remember Color substitution result.

\subsection{Operating Characteristics and Grid Resolution}
\label{app:resolution}

\textbf{Frozen power simulation.} The fixed-path Remember Color campaign froze
a power simulation before launch (2{,}000 Monte Carlo repetitions per setting,
seed 20260824, task-stratified paired bootstrap, assumed sham floor
$\rho_S=0.05$), which motivated $\pm7.5\pp$ as the feasible confirmatory
equivalence margin for the difference of discordance rates. The realized floor
was $0.12$--$0.145$, close to the one-ULP rate ($46/388=11.9\%$). A difference
of two binomial rates has larger variance at a higher base rate, so the
realized operating characteristics are worse than the frozen grid describes,
and the margin is roughly half of $\rho_T$. A retrospective rerun at
$\rho_S=0.13$, the actual $n=193+195$ and $\phi=0.3$ gives
Table~\ref{tab:app-realized-oc}.

\begin{table}[ht]
\centering
\small
\caption{\textbf{Retrospective operating characteristics of the realized-floor
decision grid.} Settings: $\rho_S=0.13$, $n=193+195$, $\phi=0.3$. Each entry
is the fraction of simulations reaching the column's decision at the given true
increment $\bar\theta$. This is a post-outcome diagnostic.}
\label{tab:app-realized-oc}
\vspace{4pt}
\begin{tabular}{lrrrr}
\toprule
True $\bar\theta$ & TOST $\pm7.5$ accept & 95\% LB $>0$ & positive & inconclusive \\
\midrule
$0\pp$ & 0.913 & 0.023 & 0.011 & 0.075 \\
$2.5\pp$ & 0.703 & 0.191 & 0.139 & 0.159 \\
$5.0\pp$ & 0.292 & 0.568 & 0.504 & 0.203 \\
$7.5\pp$ & 0.052 & 0.866 & 0.839 & 0.110 \\
\bottomrule
\end{tabular}
\end{table}

The realized-floor grid accepts a true zero in 91.3\% of simulations, the
boundary in 5.2\% and a true $2.5\pp$ in 70.3\%, so its discrimination at
$2.5\pp$ is limited. It is a post-outcome diagnostic that does not replace the
frozen power analysis, and it characterizes the increment produced by the swap
as delivered.

\textbf{Bridge grid (GRID-1a).} The same success readout ran on
SimplerEnv-Bridge across four tasks at $n=24$ seed-paired episodes per cell,
one attempt per episode. The eggplant cell, at ceiling in both conditions (24/24),
was excluded by the competence rule registered before launch. On the three
competent cells, $D=\mathrm{SR}(A)-\mathrm{SR}(N_{\rm within})=+9.72\pp$
($n_{\rm joint}=72$, $b{=}13$, $c{=}6$; success $47/72$ versus $40/72$; exact
McNemar $p=0.1671$). The exact interval is $[-7.28,+25.77]\pp$ at
$\alpha=0.025$ per component, so each one-sided endpoint has coverage
$\geq95\%$ and the two-sided interval $\geq90\%$. The verdict is unresolved:
the registered rule can declare an effect only at $|b-c|\geq8$, i.e.\
$|D|\geq11.11\pp$ at $n_{\rm joint}=72$, and the observed $b-c=7$ falls one
discordant pair short. The design has 80\% power only at $|D|=13.90$--$22.10\pp$
pooled, and a single cell needs $|D|\geq33.33\pp$ at the strictest Holm tier. The
exclusion does not produce the verdict: eggplant has 0/0 discordant pairs, so
restoring it changes only the denominator ($D=+7.29\pp$ over four cells,
$[-5.70,+19.72]\pp$), and $b-c=7$ either way. Actions diverged on 72/72
competent pairs, but the replicate gate did not pass
(Appendix~\ref{app:executor-floors}), so that divergence is not attributed to
the ablation. Post-data amendments to this campaign are listed in
Appendix~\ref{app:amendments}.

\subsection{Repeat Checks and Executor Floors}
\label{app:executor-floors}

\textbf{Exact replicate checks.} Zero outcome discordance under exact repeats
holds only on the learned-bank simulators, and only where a replicate condition was
run. It does not hold at the public \pihalf{} executor, whose
identical-instruction re-execution floor is about 27\% (below). These checks
show repeatability only: they are not sham interventions, and a deterministic
replicate ($C$ or $C_{\rm det}$) does not certify that the substitution
machinery leaves no footprint. The one-ULP condition $U$ checks for that footprint.

On Remember Color (RC3/RC5), the exact replicate $C$ changes $0/400$ outcomes
relative to the reference $A$ ($0/388$ on the common donor-eligible cohort;
Appendix~\ref{app:payload}). The R-1 replicate anchor changes $0/400$
(Appendix~\ref{app:r1ladder}), and the Shell Game blackout ladder's $C_{\rm det}$
changes $0/400$ (Appendix~\ref{app:blackout-ladders}). The independent
SimplerEnv-Fractal replicate gives $0/300$ flips over 12 cells, with lineage
not established (Appendix~\ref{app:learned-outcomes}). No replicate condition ran
for R27-A, whose only determinism check is a cross-process smoke test ($0/16$
(seed, attempt) action sequences differ between $C$ and the view-visible condition),
or for Q2-XA (Appendix~\ref{app:readpath}). The GRID-1a replicate gate failed,
because the companion's argv cannot match the formal run's by construction.
Its $32/32$ shared pairs (8 of 24 episodes per cell) are bit-identical, an
observation rather than a passed gate.

\textbf{Executor floor at the public \pihalf{} executor.} ChainBench outcomes
are re-executed from sealed decisions through the public
RLinf \pihalf{} LIBERO-130 executor, which is not
bitwise repeatable per episode: re-executing the same sealed full-memory
decisions gives task success $0.7723$ (E1CL run~1), $0.7574$ (run~2) and
$0.7426$ (E1B, Appendix~\ref{app:e1b}). A separate preregistered motor-level
ladder executes 202 chains under four instruction conditions (808/808
executions complete): sealed reference $C$, one-byte edit $U$,
identical-instruction replay $R$ and wrong-fact $T$. We write $R$, not $S$:
$R$ re-issues the \emph{same} instruction and measures an order-conditional
replay floor, whereas $S$ in Section~\ref{sec:arms} is a different scope carrying
the same task latent.

\begin{table}[ht]
\centering
\small
\caption{\textbf{Instruction-level manipulation check on the public executor}
(first of two full $n=202$ runs). The identical-instruction replay floor is
$26.7$--$28.2\%$ across the two runs, and the one-byte increment ($-1.49\pp$,
one-sided 95\% lower bound $-7.92\pp$, chain-level paired bootstrap) is not
detected above it. Because $T$ scores exactly $0.0000$, $\rho_T$ is
identically $\mathrm{SR}_C$: the wrong-instruction contrast only restates that
an incorrect instruction fails deterministically, is not a measurement of
content sensitivity, and is not the latent-matched $S$/$T$ contrast of
Section~\ref{sec:arms}.}
\label{tab:e1cl-ladder}
\vspace{4pt}
\begin{tabular}{lrrrr}
\toprule
Condition (baseline $C$) & Task success & $b$ & $c$ & $\rho$ \\
\midrule
$C$ sealed reference & 0.7723 & --- & --- & --- \\
$U$ one-byte edit & 0.8267 & 20 & 31 & 0.2525 \\
$R$ identical-instruction replay floor & 0.7426 & 30 & 24 & 0.2673 \\
$T$ wrong fact & 0.0000 & 156 & 0 & 0.7723 \\
\bottomrule
\end{tabular}
\end{table}

Unlike the exact replicates on the learned-bank simulators, this floor is
stochastic (motor-level reproducibility depends on RNG-stream alignment and
order, not on per-episode bitwise identity). The integrity audit reports zero
missing, duplicate or extra rows, zero byte-diff failures across all 808
instructions, and agreement between two implementations. A preregistered
40-chain, $K{=}8$ replay profile attributes most of this floor to a small
fragile subset of chains. Its preregistered split was revised after the data
(Appendix~\ref{app:amendments}).

\section{Information Demand}
\label{app:rq1}

This chapter supports the first condition of Q2, that the deciding information
is only in memory (Section~\ref{sec:rq1}): the condition holds on ChainBench
(\ref{app:chainbench}) and fails on Remember Color, where attempt~2 shows the
cue again (\ref{app:rc-floors}), and on the Shell Game success readout, where
any touch of the correct cup counts as success (\ref{app:sg-anytouch}). The
trained plugin is evaluated on these Remember Color and Shell Game tasks
(Appendix~\ref{app:trained}), where both failures apply.

\subsection{ChainBench Construction and Retrieval Diagnostics}
\label{app:chainbench}

The three ChainBench generations are never pooled: v1 is a 211-chain $K{=}2$
retrieval diagnostic, v2 is E1 (143 chains, $K{=}3$, supporting) and v3 is E1B
(202 chains, $K{=}2$, primary; Appendix~\ref{app:e1b}).

\textbf{Harness, memory configuration, and baselines.} The live
\emph{agentic memory harness} uses a closed-weight LLM planner (Claude Code /
Opus 4.8), reads and writes text memory, and dispatches subgoals to a frozen
\pihalf{} policy in LIBERO \citep{libero}. E1/E1B replay sealed decisions
through their motor bridges without re-invoking the planner. The E1B full
memory configuration combines a memory set retaining the latest write, a
calibrated presence gate, two-stage retrieval, and a leave-one-chain-out
nearest-centroid readout. The query chain's events are excluded from the
centroids. The latest-write memory set is selected by episode role, not the
fact label, and is a stipulated state rather than the output of an evaluated
eviction policy. These details describe E1B and are not assumed for E1.
The initial-fact baseline instead selects the answer specified by the chain's
initial fact, ignoring updates. It does not invoke the harness memory path
(Table~\ref{tab:chainbench-full-baselines}). The separate $C/U/S/T$ experiment
in Appendix~\ref{app:poscal} tests the nearest-centroid readout on curated
records, not writing or retrieval by the live harness. Its positive-control
and fact-specificity interpretations are two uses of the same experiment.

\textbf{E1B decision provenance.} The frozen 202-chain manifest is a subset
of the 403-chain v3 evaluation. Its decision entries match the corresponding
evaluation rows for all ten conditions: 202 chains for conditions defined on
the full cohort and 97 for the three restricted to rewritten chains. Initial
facts, final facts, and rewrite flags also agree. The task-success rates in
this paper come from the August 25 rerun of these sealed decisions, not the
earlier execution export. The archived run artifacts do not record the
source digest of the decision evaluator. Matching decisions and data digests
therefore establishes the cohort correspondence but not the exact historical
evaluator code version. Additional non-confirmatory memory-state and
retrieval comparisons are reported in Appendix~\ref{app:memory-maintenance}.

\textbf{Decision-level retrieval construct (v1).} In each chain exactly one
earlier LIBERO episode writes the deciding object--drawer relation, later
episodes add distractor writes, and the query frame is the same for every
hidden fact (Figure~\ref{fig:chain}). In v1 the hidden fact is balanced 108:103
and the query observation is pixel-identical across possible answers. The
oracle scores 1.000 and the no-memory decision rule 0.498, near the $K{=}2$
fallback floor of $1/2$. Appearance top-1 retrieval selects the informative
own-chain write in 0/211 cases, always choosing a visually similar same-scene
distractor. These are decision-level outcomes, not robot task success rates,
and v1 has no local closeout or preregistration.

\textbf{Q2 (only in memory) reading on E1B.} Without memory, decision accuracy is
$107/202=0.530$ (exact Clopper--Pearson 95\% CI $[0.458,0.600]$), an interval
that contains the analytic floor $1/K=0.500$. With curated records, accuracy is
$202/202$ (exact Clopper--Pearson 95\% CI $[0.982,1.000]$). The paired decision
gain is $+0.470$ ($b{=}95$, $c{=}0$, with curated records as the reference condition of Section~\ref{sec:net-benefit}). Because $c=0$, the gain equals $b/n$, so its
exact Clopper--Pearson 95\% CI is $[+0.400,+0.542]$. The preregistration and
the closeout score the no-memory value against a mixed-$K$ fallback floor
(per-$K$ chance rates taken from the manifest structure). The finding that
every E1B chain has $K{=}2$, and with it the $0.500$ floor and the $+0.470$
gain, comes from a check made after the E1B closeout and is post hoc. Motor
outcomes and all baselines are in Appendix~\ref{app:e1b}.
\emph{Verdict:} Q2 (only in memory) \textbf{holds} on ChainBench.

\subsection{Remember Color Retry Semantics and Chance Floors}
\label{app:rc-floors}

\textbf{Attempt~2 reruns the same seed and shows the cue again.} Both attempts
of a task--seed scope start from the same seeded simulator reset. On the
RC3/RC5 substitution campaign, the post hoc read-only survival audit finds the
attempt-2 seed and initial-observation hash equal to those of attempt~1 in
1,976/1,976 completed attempt pairs. Byte identity is established only for this
reset frame. Later frames depend on the policy's actions. The environment
source shows the
true-color cube at elapsed steps 0--4, hides all cubes at steps 5--9 and shows
the candidate cubes from step~10 on. The schedule runs on the elapsed-step
counter, which the reset sets to zero, so attempt~2 replays it and shows the
cue again: the deciding color is visible in attempt~2's own observation, and
Q2 (only in memory) \textbf{fails} for cross-episode records on Remember Color. Within an
attempt the cue is hidden before the candidates appear, so the task still
demands recall within the episode.

\textbf{Chance floors.} Under the released configuration, where no memory
content carries across attempts (Appendix~\ref{app:mikasa}), success (released
condition of Table~\ref{tab:app-mikasa}; $n=200$ per task, attempts 1--2 pooled) is
34.0\% on RC3 against the analytic floor $1/3=33.3\%$ (margin $+0.7\pp$) and
23.5\% on RC5 against $1/5=20.0\%$ ($+3.5\pp$). On Remember Color the policy is
therefore near chance, whereas the same checkpoint succeeds on Shell Game
(Appendix~\ref{app:checkpoint-provenance}). The floor $1/N$ is the success rate
of one uniform choice among $N$ cubes, but within-attempt retry semantics are
unverified: no local record states whether touching a wrong cube ends the
episode or whether success is scored at any step, as on Shell Game
(Appendix~\ref{app:sg-anytouch}). If several touches are allowed within an
attempt, the chance rate exceeds $1/N$ and these margins shrink or change sign.

\subsection{Shell Game Any-Touch Rule and Cue-Blind Screen}
\label{app:sg-anytouch}

\textbf{Success rule.} Shell Game success is scored
once reached: an episode succeeds if the success condition holds at
any step. Any touch of the correct cup therefore counts as success, including a
touch that follows a touch of a wrong cup (frozen E0 precommit, Section 0.4, repeated
in the later E2-W0 and Q2-XA precommits). The three-cup floor of $1/3$ is the
rate of a single guess, not the chance rate under this rule. Success does not
require identifying the cup, so Q2 (only in memory) \textbf{fails} on the Shell Game success
readout.

\textbf{First-visit diagnostic.} The first-visited cup is the first cup whose
tool-center-point distance falls below $0.08$. The frozen E0 runner did not log
the true cup, so it was recovered offline from the reset seed by a side script
that loads no policy, accepting a seed only when two reads agreed (E0
amendment 4, made before any formal result; 200/200 formal seeds determined).
In the full-view conditions of E0 and E2-W0, the first-visited cup is correct in
0.448--0.500 of attempt-2 successes (for $A_{W0}$, $78/172=0.453$, exact
Clopper--Pearson 95\% CI $[0.378,0.531]$), far below the success rates of
0.865--0.905, and the true cup is among the visited cups in every success. In
Q2-XA, turning the within-episode reads on or off leaves the attempt-1
first-visit count unchanged: $88/177$ versus $88/178$ (exact Clopper--Pearson
95\% CI $[0.421,0.573]$ and $[0.419,0.570]$). These counts were computed after
adjudication and are post hoc. The counts are consistent with success reached
by touching cups in turn, but this remains a hypothesis rather than an established
explanation. First-choice accuracy under the camera blackout was not computed:
the frozen R27-A template required cup positions and tool trajectories, neither
its sidecar nor its pair records contain them, and the blackout ladder
(Appendix~\ref{app:blackout-ladders}) reports no first-visit diagnostic. We
therefore make no statement about how success under blackout is reached
(Appendix~\ref{app:viewremoval}).

\textbf{Cue-blind screen (E0).} E0 is a preregistered screen: its precommit was
frozen before data, and all four amendments were made before any formal result.
It searched for a Shell Game reading condition in which the policy stays
competent and the answer can come only from persistent records written in
attempt~1. Such a condition would qualify a later content ladder, but the
screen itself is not content evidence. E0 runs persistent PCMB
with five conditions paired on seeds 7001--7200
($n_{\rm joint}=200$): records kept ($A$) or cleared at the attempt-2 boundary
with the scope clock preserved ($N$), crossed with the full view ($W0$) or a
zero camera frame for attempt-2 elapsed steps $\leq5$ ($W1$), plus a positive
control $P_{W1}$ that keeps the cue visible past the blanked window.

\textbf{E0 results.} With the full view, success is $173/200$ whether records
are kept or cleared ($b{=}3$, $c{=}3$, $D=0.0\pp$; exact McNemar $p=1.0$), with
an exact upper bound of $3.42\pp$ (one-sided 95\% Clopper--Pearson components,
difference bound $\geq90\%$). This difference is derived post hoc from
preregistered components, rather than being a preregistered estimand. Its paired-bootstrap 95\%
CI under the frozen protocol, $[-2.5,+2.5]\pp$, is anti-conservative at six
discordant pairs. The contrast concerns memory presence, not memory content.
Blanking the cue lowers success from $0.865$ ($A_{W0}$) to $0.140$ ($A_{W1}$;
$b{=}145$, $c{=}0$), and the Wilson 95\% upper bound for $A_{W1}$, $0.195$, is
below the $1/3$ single-guess rate. The positive control reaches only $0.420$,
below the preregistered competence level of $0.60$. The preregistered primary
estimand $D_N^{W1}=\mathrm{SR}(A_{W1})-\mathrm{SR}(N_{W1})=-0.5\pp$ ($b{=}5$,
$c{=}6$; paired bootstrap 95\% CI $[-4.0,+2.5]\pp$) therefore cannot be
interpreted, because blanking itself breaks the policy. The adjudicated cell is
``blanking breaks the policy'': no reading condition qualified, and no content
ladder was run on E0. A defective first runner was replaced before any formal
unit ran (Appendix~\ref{app:twins}).

\section{Delivery, Read Path and Validity}
\label{app:rq2}

This chapter supports the delivery and validity conditions of Q2
(Section~\ref{sec:rq2}). The survival audit is post hoc, whereas
the read-path and version-conflict studies were preregistered.

\subsection{Decision-Time Survival Audit}
\label{app:survival}

\textbf{Definition.} For a substituting condition, an attempt-2 retry is
\emph{covered} when every decision query selects at least one unmodified
substituted record. Coverage is the fraction of covered retries, and the Q2
delivery criterion is coverage $=1$ (Section~\ref{sec:conditions}). A selected record counts as
unmodified when its (slot, timestep, version) tuple equals that slot's tuple at
the first attempt-2 query $q_0$ and the slot was not stored into or evicted at
an earlier query. In the three audited campaigns every policy query is a decision
step: the action hash recorded at each query equals the episode's action-hash
sequence at that index in $59{,}280/59{,}280$ query records. Read-side and
write-side classifications of a query disagree in no query of any campaign or
condition. Coverage is a lower bound on delivery (see below).

\textbf{Status.} The audit is post hoc and read-only. It was run on 2026-09-14
on sealed rows after the campaigns had closed, and none of the cross-episode
campaigns in Table~\ref{tab:coverage} registered a survival gate. The two
Shell Game blackout ladders registered one ($G_{\text{surv}}$) as a hard gate
before launch.

\textbf{Installation.} In the cross-episode campaigns, substitution is installed
once on the whole bank at the start of attempt~2 (Appendix~\ref{app:payload}).
When a donor scope holds fewer records than the recipient, the swap fails
closed, which removes 7 scopes from each of $S$ and $T$ on Remember Color~3 and
5 on Remember Color~5 (denominators 193 and 195).

\begin{table}[ht]
\centering
\scriptsize
\caption{\textbf{Decision-time survival by campaign.} Top: post hoc audit of
the cross-episode campaigns, which registered no survival gate. A retry counts
when every attempt-2 decision query selects at least one unmodified substituted
record, so the rate is a lower bound on delivery. The audit covered the
donor-swap conditions $S$ and $T$. Bottom: the registered survival gate of the Shell
Game blackout ladders, which checks that each decision step of a substituting
trial carries substituted content, over 800 trials per condition. $^{a}$Not reported
by the audit. $^{b}$Withdrawn as Q3/Q4 evidence (Appendix~\ref{app:withdrawn}).}
\label{tab:coverage}
\vspace{4pt}
\begin{tabular}{@{}llrr@{}}
\toprule
Campaign & Condition & Passing / total & Rate \\
\midrule
\multicolumn{4}{@{}l}{\emph{Cross-episode substitution, installed once at attempt-2 start (post hoc audit)}} \\
Remember Color 3, fixed-path substitution & $S$ & $135/193$ & 0.70 \\
 & $T$ & $150/193$ & 0.78 \\
Remember Color 5, fixed-path substitution & $S$ & $161/195$ & 0.83 \\
 & $T$ & $154/195$ & 0.79 \\
Shell Game full-vision ladder, band A$^{b}$ & $S$ & $134/198$ & 0.68 \\
 & $T$ & $133/198$ & 0.67 \\
Shell Game full-vision ladder, band B$^{b}$ & $S$ & $121/198$ & 0.61 \\
 & $T$ & ---$^{a}$ & ---$^{a}$ \\
Trained plugin, Remember Color 5 substitution & $S$ & $100/196$ & 0.51 \\
 & $T$ & $107/196$ & 0.55 \\
\midrule
\multicolumn{4}{@{}l}{\emph{Shell Game blackout ladders, substituted at every read with history (registered gate)}} \\
Released configuration & $U,S,T$ & $2{,}400/2{,}400$ & 1.00 \\
Structured-PCMB configuration & $U,S,T$ & $2{,}400/2{,}400$ & 1.00 \\
\bottomrule
\end{tabular}
\end{table}

\textbf{Why coverage is a lower bound.} Selection and merge identity read only
fields that the swap preserves: the selector scores the cognitive and perceptual
keys and the state signature, so the swap changes neither which records a query
selects nor which stored record an attempt-2 write merges into. A merge
(\MERGE) keeps the slot, sets the timestep to the current clock and advances the
version. It overwrites the cognitive and perceptual values and the state
signature with the current observation but copies the other six swapped fields
(action chunk, pre- and post-action state, interaction signature, outcome
embedding and failure signature), so a merged record counts as modified yet
still carries six of the eight swapped fields. Coverage counts only unmodified
records, so delivery of substituted content is at least as high as coverage.

\textbf{Early records.} The record timestep is a per-scope query counter that
continues from attempt~1 into attempt~2, and in every audited campaign and condition
the slots with timestep 0 or 1 are overwritten at $q_0$ or $q_1$ in every retry
(for example $193/193$ on Remember Color~3 $S$ and $196/196$ on the plugin $S$
condition).

\textbf{Verdict.} By the coverage criterion, Q2 delivery fails in all three audited
cross-episode campaigns (post hoc; coverage $0.51$--$0.83$), so Q3/Q4 results on the
Remember Color and plugin shams hold only for the swap as delivered
(Appendices~\ref{app:payload} and~\ref{app:plugin-sham}). Both blackout ladders
pass their registered survival gate: in each, $2{,}400/2{,}400$
substituting-condition trials carry a decision-step substitution, and every read with
history is substituted with a changed content digest (counts and the released
ladder's post-data amendment of its sham-validity gate in
Appendix~\ref{app:blackout-ladders}). In these ladders $S_\pi$ and $T_\mu$ act
only on floating-point leaves inside each retrieved latent, so no condition changes
which record is selected.

\subsection{Read-Path Gates}
\label{app:readpath}

These studies check, before any outcome is read, that an intervention reaches
the read path it targets. Each is a presence statement about one task and one
configuration, not a statement about record content.

\textbf{Within-episode read (Q2-XA).} A preregistered study on Shell Game
(seeds $7601$--$7800$, $n=200$) uses the released configuration, whose
episode-local perceptual and cognitive banks are on the inference path and under
which no memory content carries across attempts. It disables both reads in both
attempts and leaves writes untouched: the perceptual read by the harness
no-memory override, the cognitive read by forcing history selection empty.
Every disabled call was forced ($9{,}200/9{,}200$ per bank), and action
sequences differed between conditions in every episode ($200/200$ per attempt). No
replicate condition was run, so bitwise determinism is not established. First-attempt
success moves from $0.885$ to $0.890$ ($D=-0.5\pp$; $b{=}3$, $c{=}4$;
exact Clopper--Pearson difference upper bound $3.14\pp$), and second-attempt
success from $0.910$ to $0.905$ ($D=+0.5\pp$; $b{=}3$, $c{=}2$; upper bound
$3.65\pp$). Each bound combines two one-sided 95\% Clopper--Pearson components,
so it holds with probability $\geq90\%$. The exact McNemar test gives $p=1.0$ in both attempts.
Both exact bounds were added after data beside the recorded paired-bootstrap
bounds ($1.5$ and $2.5\pp$), which are anti-conservative at so few discordant
pairs. The verdict is unchanged (Appendix~\ref{app:amendments}).
The read-path gate holds, but an outcome effect is not detected at this design's
resolution.

\textbf{Persistent-path ladder (E2-W0).} The contract was frozen before any
data. The ladder runs on Shell Game with the view available under the
structured-PCMB configuration, on the seed band of Q2-XA ($7601$--$7800$,
$n=200$ per condition) but with a different memory architecture, so the two are not
pooled. Under this configuration the persistent record path is the only memory
path (see the probe below): $N_{\rm persist}$ clears the attempt-1 persistent
records but keeps the query clock, and $N_{\rm allread}$ forces every attempt-2
persistent read to return empty. Reads were forced empty in $4{,}600/4{,}600$
attempt-2 selection calls, removing $18{,}137$ records that would otherwise have
been selected. At the first attempt-2 query, $N_{\rm persist}$ has zero
candidates while the reference condition and $N_{\rm allread}$ have candidates, so
$N_{\rm allread}$ removes the read, not the bank. Attempt-1 action sequences are
bitwise identical across the three conditions in $200/200$ seeds, and the clock gate
passes $600/600$. Success is $0.905$ in all three conditions ($181/200$ each). Each
contrast has $b{=}c{=}2$ and $D=0.0\pp$, with exact Clopper--Pearson difference
upper bound $2.94\pp$ (one-sided 95\% components, $\geq90\%$; added after data beside the recorded paired-bootstrap bound, $1.5\pp$, anti-conservative at $b{=}c{=}2$), Tango 95\% score
interval $[-2.69,+2.69]\pp$ and exact McNemar $p=1.0$. The path gate holds, but an
outcome effect is not detected at this design's resolution. Record survival was
not audited in this campaign.

\textbf{Path-scope probe (development only).} A development probe showed that
under persistent PCMB the episode-local cognitive and perceptual
banks are instantiated but their history-selection path does not execute (call
counts 0 while 63--82 persistent records were selected), so the persistent path
is the only memory path. Two earlier versions of the contract and runner were
voided before formal data, and the probe supports no other claim.

\textbf{Remember Color first-query reads.} In the fixed-path substitution
campaign (the five conditions of Remember Color~3 and~5), the attempt-2 first-query
candidate count has the same distribution as the attempt-1 end-of-episode record
count (4--13), and every attempt-2 first query selects exactly four records
($1{,}976/1{,}976$). The installed bank is therefore read at $q_0$. Whether
substituted records are still selected later is the coverage question of
Appendix~\ref{app:survival}.

\subsection{Version Conflict}
\label{app:version}

In this subsection
$\Delta=\mathrm{SR}(\text{record-retaining condition})-\mathrm{SR}(\text{comparator})$,
so a negative $\Delta$ means the retaining condition succeeds less. For each contrast,
we give the two discordant counts. Remember Color shows the cue again at the
start of every attempt (Appendix~\ref{app:rc-floors}), so these designs ask
whether an outdated attempt-1 record lowers success when the current fact is
observable, not whether memory supplies it. Decision-time survival of attempt-1
records was not audited in these campaigns.

\textbf{Frozen 690-pair test (configuration contrast).} Before collecting new
outcomes, we froze a paired TOST with a $\pm5\pp$ equivalence band on Remember
Color~3, 5 and~9, with 230 pairs per task (690 pairs and 2,760 episodes). The
two conditions are two memory methods run by the same harness on the same conflict-on
chains: the version-merge condition, persistent PCMB (on
whose path attempt-1 records persist into attempt~2) and the seed-paired
comparator, released MemoryVLA, under which no memory content carries
across attempts (Appendix~\ref{app:payload}). This is a configuration contrast
(architecture and persistence change together), not a comparison of retaining
versus clearing outdated records. The structured condition is lower by
$\Delta=-3.91\pp$ (91 pairs in which only the comparator succeeds, 64 in which
only the structured condition succeeds; exact McNemar $p=0.036$). The TOST uses the
90\% Tango score interval, $[-6.91,-0.95]\pp$ \citep{tango1998}. The 95\% Tango
interval is $[-7.48,-0.38]\pp$. The 90\% interval lies wholly below zero and
crosses the $-5\pp$ boundary, so the preregistered TOST harm criterion is met
for this configuration contrast, and the difference is not shown to be small.
By task (comparator-only pairs, structured-only pairs, $\Delta$): Remember
Color~3 $(44,29,-6.5\pp)$, 5 $(24,21,-1.3\pp)$, 9 $(23,14,-3.9\pp)$. This
campaign ran evaluator revision $e_3$, whose source survives only
as a checksum (Appendix~\ref{app:checkpoint-provenance}).

\textbf{No-conflict twin.} A separately preregistered twin keeps the checkpoint,
environment and seed pairing of the sealed run and sets the attempt-2 conflict
shift to $0$. It ran a later evaluator revision ($e_1$), so
evaluator identity across the two runs is not established. Over the same $690$
quadruples, $\Delta_{\rm twin}=-1.59\pp$ (64 comparator-only, 53 structured-only
pairs), with a 90\% TOST interval (method not named in the closeout) of
$[-4.20,+1.01]\pp$ inside the $\pm5\pp$ band. The seed-paired
difference-in-differences is $+2.32\pp$ (seed-level paired-bootstrap 90\%
interval, 1,000 replicates, $[-1.59,+6.23]\pp$) and is not resolved on its own.
The preregistered rule (twin interval inside the band) classifies the gap as
conflict-specific, but the frozen power analysis gave that rule a $0.139$
probability of passing if the gap were a general lag of the structured condition, so
we make no conflict-specificity claim.

\textbf{Same-architecture $2\times2$ (persistence only; Q3-VC).} A subsequent
preregistered $2\times2$ runs all four conditions under persistent PCMB with
the same harness, checkpoint and seeds ($8001$--$8230$ per task), crossing
conflict (attempt-2 shift $1$ or $0$) with whether the attempt-1 records
persist. The cleared conditions remove the scope's persistent records and pending
executions but keep its query clock, and the candidate and clock gates pass
($n_{\rm joint}=690$ per contrast). The persistence effect under conflict is
$\Delta=+0.29\pp$ (68 pairs with only the persisted condition succeeding, 66 with
only the cleared condition; task--seed paired-bootstrap 95\% interval, 10,000
replicates, $[-3.04,+3.48]\pp$), and the 90\% task--seed paired-bootstrap TOST
interval $[-2.46,+3.04]\pp$ lies inside the $\pm5\pp$ band. The per-task
effects ($+5.22$, $-0.87$ and $-3.48\pp$ on Remember Color~3, 5 and~9) disagree
in direction. From the frozen power analysis (pooled pairwise discordance
$\approx0.323$), the design had only $\approx0.42$ probability of a wholly
negative 95\% interval at a true $-3.91\pp$, so this is a failure to reproduce,
not a demonstration that the effect is absent. Validity remains unresolved, so no claim
is made that outdated records themselves lower success.

\section{Content Ladders}
\label{app:rq3}

This chapter reports the content ladders behind Section~\ref{sec:rq3}, each scoped by
its delivery status (Table~\ref{tab:link-profile}).

\subsection{Shell Game Blackout Ladders}
\label{app:blackout-ladders}

\textbf{Design.} Two preregistered ladders run inside the post-cue camera
blackout of the Shell Game view-removal study (cameras blanked from elapsed
step~6 to episode end; Appendix~\ref{app:viewremoval}), on one checkpoint and
one task. The released ladder reads the released configuration's two
episode-local banks; the structured ladder runs structured PCMB without
cross-episode carry. The two differ in memory architecture, are not an
ablation, and are not tested against each other. In both, every registered bank
is empty at the first memory query of every episode, so the ladders act on
within-episode history only. Both use seeds $610000$--$610399$, $n=400$ per
condition, paired on (seed, attempt~1). Attempt~2 is recorded and enters no
criterion. Wave~1 runs $A$, $S_\pi$, $T_\mu$ and $U$; wave~2 runs $A$,
$N_{\rm blk}$ and $C_{\rm det}$ ($C_{\rm det}$ is labeled $C$ in
Figure~\ref{fig:sham}a). Each contrast is taken against its own wave's anchor,
and on the released ladder the two anchors agree on all 400 paired outcomes.
The released ladder's precommit had planned $U$ at $n=200$ and $C_{\rm det}$ at
$n=50$, both in wave~2, with $N_{\rm blk}$ in wave~1. The launched manifests run
every condition at 400 seeds. The view-removal $2\times2$ used the disjoint seeds
$8701$--$8900$, so success levels are never differenced across the two
campaigns. Any touch of the correct cup counts as success. First-visit accuracy
under blackout was not computed.

\textbf{Conditions.} Each history entry read at the hook is an (episode identifier,
latent) pair. Every substituting condition edits only the floating-point leaves
inside each retrieved latent and never an identifier, so no condition changes which
record is selected. $A$ is the unablated anchor. $U$ steps every float leaf by
one ULP of its own dtype. $S_\pi$ permutes the scalar components within each
latent, preserving its multiset (norm, mean, variance and every order
statistic) bitwise while destroying the arrangement. $T_\mu$ resamples each
latent from its own empirical mean and standard deviation, matching scale and
dtype but not content. $C_{\rm det}$ replays the anchor with no substitution
applied, and $N_{\rm blk}$ blocks the read of both banks. The precommit named
the two content conditions ``same-latent sham'' and ``different-latent swap''.
The implemented operations are a within-latent permutation and a moment-matched
resample, not the cross-scope donor swap of Appendix~\ref{app:payload}, so
$\theta=\rho_{T_\mu}-\rho_{S_\pi}$ compares resampling with permuting. Here, $C_{\rm det}$ checks exact repeatability. The $U$ condition
measures sensitivity to a minimal numerical change applied through the
substitution machinery.

\textbf{Delivery and sham validity.} On both ladders the registered survival
check passes: all $2{,}400$ substituting-condition trials carry substituted content
at the decision step, and every read with retrieved history is substituted with
a changed content digest ($35{,}200$ reads per substituting condition on the released
ladder, $33{,}600$ on the structured ladder; Appendix~\ref{app:survival}). On
the released ladder the registered sham-validity check required $S_\pi$ to flip
at least one outcome (it flips 27) and a substitution-applied fraction of
$1.0$. The frozen implementation divided substitutions applied ($35{,}200$) by
all reads ($36{,}800$), including each bank's read at each episode's first
query, which carries no history and cannot be substituted ($2\times800=1{,}600$
reads), so the fraction is $22/23=0.9565$ whatever the outcomes: \emph{as
registered, the check fails}. After the data existed, an amendment
(ladder amendment 3) replaced the denominator with the reads that
carry history and merged the check with the registered substitution-coverage
gate, whose two conjuncts it now also requires: every applied substitution
changes content, and the empty reads equal their prediction ($1{,}600$ observed,
$1{,}600$ predicted). \emph{As corrected, the check passes}
($35{,}200/35{,}200=1.0$). The sealed data were re-adjudicated without a rerun
(the runner was byte-identical). The first two adjudication artifacts also
split the $\theta$ block incorrectly ($b'=c'=6$, one-sided $p=0.613$). The
reported split tabulates $b'$ and $c'$ on the flip indicators, as the precommit
defines $\theta$, and $\hat\theta=+0.50\pp$ in both. The structured ladder's
rule was set by a later amendment written after its smoke run and before its
formal launch: its empty-read count is data-dependent ($3{,}200$ empty reads, at
the same four within-episode positions in every condition), so the rule requires
identical empty-read positions across conditions instead of a predicted count.
The mechanism behind the second pair of empty reads was not determined. That check
passes (24 flips; $33{,}600/33{,}600=1.0$). Both amendments are listed in
Appendix~\ref{app:amendments}.

\begin{table}[ht]
\centering
\scriptsize
\setlength{\tabcolsep}{3pt}
\caption{\textbf{Shell Game blackout ladders.} Seeds $610000$--$610399$,
$n=400$ per condition, paired on (seed, attempt~1). $b$ counts pairs in which the
anchor $A$ succeeds and the condition fails and $c$ the reverse, so $\rho=(b+c)/n$
and $D=(b-c)/n$, positive when the condition lowers success. $C_{\rm det}$ and
$N_{\rm blk}$ are scored against the wave-2 anchor, the other conditions against the
wave-1 anchor. $b'$ and $c'$ count pairs in which only $T_\mu$ or only $S_\pi$
changes the outcome. The conditional exact bound (each endpoint one-sided
95\%) holds given the observed $b'+c'$ and is not an unconditional interval
for $\theta$.}
\label{tab:blackout-ladders}
\vspace{4pt}
\begin{tabular}{@{}llrrrrrr@{}}
\toprule
 & & \multicolumn{3}{c}{released} & \multicolumn{3}{c}{structured PCMB} \\
\cmidrule(lr){3-5}\cmidrule(lr){6-8}
Condition & Operation & $b/c$ & $\rho$ & $D$ (pp) & $b/c$ & $\rho$ & $D$ (pp) \\
\midrule
$C_{\rm det}$ & anchor replayed, no substitution & 0/0 & 0.00\% & $0.00$ & 0/0 & 0.00\% & $0.00$ \\
$U$ & one ULP on every stored float & 3/1 & 1.00\% & $+0.50$ & 6/4 & 2.50\% & $+0.50$ \\
$S_\pi$ & permuted within each latent & 23/4 & 6.75\% & $+4.75$ & 18/6 & 6.00\% & $+3.00$ \\
$T_\mu$ & resampled from own mean and s.d. & 24/5 & 7.25\% & $+4.75$ & 23/5 & 7.00\% & $+4.50$ \\
$N_{\rm blk}$ & both reads blocked & 21/4 & 6.25\% & $+4.25$ & 18/4 & 5.50\% & $+3.50$ \\
\midrule
\multicolumn{2}{@{}l}{$\hat\theta=\rho_{T_\mu}-\rho_{S_\pi}$ ($b'/c'$)} & \multicolumn{3}{c}{$+0.50\pp$ (7/5)} & \multicolumn{3}{c}{$+1.00\pp$ (6/2)} \\
\multicolumn{2}{@{}l}{exact one-sided McNemar $p$ for $\theta>0$} & \multicolumn{3}{c}{0.387} & \multicolumn{3}{c}{0.145} \\
\multicolumn{2}{@{}l}{conditional exact bound on $\theta$} &\multicolumn{3}{c}{$[-1.11,+1.91]\pp$} & \multicolumn{3}{c}{$[-0.40,+1.81]\pp$} \\
\multicolumn{2}{@{}l}{$N_{\rm blk}$ effect: exact one-sided McNemar $p$} & \multicolumn{3}{c}{0.00046} & \multicolumn{3}{c}{0.0022} \\
\bottomrule
\end{tabular}
\end{table}

\textbf{Estimates.} Table~\ref{tab:blackout-ladders} gives all counts. The
intervals for $\hat\theta$ are conditional exact bounds: each is the exact
Clopper--Pearson bound (each endpoint one-sided 95\%) within the singly
discordant pairs, rescaled by $(b'+c')/n$ at the observed $b'+c'=12$ and $8$,
and neither is an unconditional interval for $\theta$. Both ladders meet the
registered floor $b'+c'\geq8$, so the preregistered expansion to 600 seeds was
not triggered. At $n=400$ the released ladder's precommit puts the minimum
detectable $\theta$ at $2.33\pp$ (exact one-sided test, 80\% power) under an
assumed---derived, not measured---$T$--$S$ disagreement rate of $3.2\%$, and at
$3.65\pp$ if that rate is $8\%$. The observed $T$--$S$ disagreement rates are $3.0\%$ and $2.0\%$. On
both ladders, resampling does not change detectably more outcomes than
permuting. This null result bounds $\theta$ and does not establish $\theta=0$. Both
scrambles do change outcomes, and the registered numerical-floor check
$\rho_U<\rho_{S_\pi}$ passes on both ladders. Blocking the read is the
registered in-run eligibility check (one-sided $p<0.05$), which passes on both.
No magnitude criterion was registered for it. These results hold for this
checkpoint, task and blackout condition only. They support no claim that the
two configurations' $\theta$ agree or differ, and no claim about attempt~2.

\subsection{Remember Color Fixed-Path Substitution}
\label{app:payload}

\textbf{Scope and conditions.} On Remember Color the policy is near chance
(Appendix~\ref{app:rc-floors}). The only-in-memory condition of Q2 fails because
attempt~2 shows the cue again (Section~\ref{sec:cross}), and the delivery condition
fails the coverage test. Every
estimate below therefore holds for the swap as delivered, with survival
coverage $0.70$--$0.83$ as a lower bound on delivery
(Appendix~\ref{app:survival}). The formal campaign uses five conditions: reference
payload ($A$), exact replicate ($C$), one-ULP numerical control ($U$, which
moves every stored payload value by one representable value),
task-latent-matched cross-scope control ($S$), and different-target-fact swap
($T$). At installation every condition preserves the recipient's keys, state
signature, slots, timestamps, versions, candidate count, selected count, and
first-query path. $S$ matches the task fact under test through the same
intervention interface but is not a claim of complete semantic identity,
because pose, appearance, and non-target trajectory content may differ.

\textbf{Persistence and isolation.} Every condition runs the
persistent PCMB method, so the substitution acts on the
structured-PCMB persistent bank. Over the recorded pairs the attempt-2
first-query candidate counts have the same distribution as the attempt-1
end-of-episode record counts, so attempt-1 records reach attempt~2. The MIKASA
paired-retry harness assigns each task--seed pair one scope and clears memory at
both ends of the pair (an unscoped reset before attempt~1, a reset scoped to
the pair after attempt~2), so isolation between pairs does not depend on the
final deletion succeeding. Records persist across attempts only under the
record-retaining structured-PCMB methods; under the released MemoryVLA method
the scope is withheld from the method, no memory content carries across
attempts, and both resets simply clear its two banks
(Appendix~\ref{app:mikasa}). Scope and episode identifiers are truncated
SHA-256 hashes of (task, seed) and (scope identifier, attempt), and the
evaluator refuses to launch if its method-visible allowlist (RGB, instruction,
own action, opaque scope, episode boundary) overlaps its evaluator-only
allowlist (reward, success, termination, seed, task, attempt ordinal). This
check operates on the declared lists.

\textbf{Installation and retrieval.} Substitution is installed once, at the
start of attempt~2 and before its first query: the recipient scope's records,
ordered by (timestep, slot), each receive the payload of the donor record at
the same position (cognitive and perceptual values, action chunk, pre- and
post-action state, interaction signature, outcome embedding, and failure
signature), while slot, scope, source episode, timestamp, version, cognitive
and perceptual keys, and state signature are kept. The exact replicate installs
nothing; the one-ULP condition installs the recipient's own payload, with every stored
payload value moved by one representable value, through the same routine.
Retrieval uses a parameter-free similarity selector whose scorer reads only the
two keys and the state signature, compared against the current cognitive and
perceptual tokens: records are ranked by a weighted sum of cognitive and
perceptual key similarity, a query is accepted only when the best state
similarity is at least 0.45, and at most four records are returned. The scorer
does not use content fields, versions, slots, or timestamps. Slots and timestamps
are used only to break ties. Task and seed determine which records are searched
through the opaque scope identifier. The instruction, which names the target, is
method-visible and conditions the cognitive tokens from which the cognitive keys
are formed, so it shapes the compared embeddings, although the instruction text
is not itself a compared field. Attempt-2 merges overwrite only the cognitive
and perceptual values (with the state signature), so six of the eight swapped
fields survive a merge. Survival coverage ($0.70$--$0.83$) is therefore a lower
bound on delivery, and no survival check was registered for this campaign
(Appendix~\ref{app:survival}).

\begin{table}[ht]
\centering
\scriptsize
\caption{\textbf{Seed-level joint tables.} Flip indicators are $I_T$ and
$I_S$. Here, $n_{10}$ denotes true-only discordance and $n_{01}$ sham-only
discordance. Donor availability removes 7 RC3 and 5 RC5 scopes, below the
frozen 5\% failure budget. All rates hold for the swap as delivered
(Appendix~\ref{app:survival}).}
\label{tab:app-payload}
\vspace{4pt}
{\small%
\begin{tabular}{lrrrrrrrr}
\toprule
Task & $n$ & $n_{11}$ & $n_{10}$ & $n_{01}$ & $n_{00}$ & $\rho_T$ & $\rho_S$ & $\theta_t$ \\
\midrule
RC3 & 193 & 9 & 22 & 19 & 143 & 16.06\% & 14.51\% & $+1.55\pp$ \\
RC5 & 195 & 11 & 13 & 12 & 159 & 12.31\% & 11.79\% & $+0.51\pp$ \\
\bottomrule
\end{tabular}}
\end{table}

\textbf{Estimate and equivalence reading.} Computed from the joint flip tables
in Table~\ref{tab:app-payload}, the task-macro estimate is
$\bar\theta=+1.03\pp$, with task-stratified paired-bootstrap 95\% interval
$[-3.10,+5.16]\pp$ and one-sided task-stratified permutation $p=0.318$
($H_1:\bar\theta>0$). The corresponding task-stratified paired-bootstrap 90\%
interval $[-2.34,+4.41]\pp$ lies within the preregistered $\pm7.5\pp$ equivalence
margin. For the swap as delivered, the two-one-sided-test reading is therefore
that the target increment beyond the matched control is not larger than
an absolute $7.5\pp$. On the joint pairs, the one-ULP condition does not differ significantly
from $T$ or $S$ (post hoc exact McNemar: $25$ $U$-only against $34$ $T$-only
discordant pairs, $p=0.30$; $27$ against $32$ $S$-only, $p=0.60$), so these
changes do not identify content. With about 60 discordant pairs per comparison,
this is not an equivalence test. The 90\% upper endpoint, normalized by the 14.2\% task-macro target
discordance, gives an approximate 31\% descriptive upper bound on the share of
target-condition discordance beyond the matched control. On the common donor-eligible
cohort the one-ULP condition changes $46/388$ outcomes (11.86\%), while the exact
replicate changes $0/388$. The larger determinism cohort remains $0/400$.
Producer and independent statistical implementations agree within 0.005, and
all formal integrity checks pass. The analysis uses attempt-2 success after
attempt~1 writes memory. With $D_T=\mathrm{SR}_A-\mathrm{SR}_T$
(Section~\ref{sec:net-benefit}), $D_T=+3.63\pp$ on RC3 and $+2.05\pp$ on RC5. At the first
action chunk, $U/S/T$ differ from the reference on every scope while the exact
replicate differs on none. Mean step-0 $L_2$ distances are
$0.0064/0.0236/0.0195$ for $U/S/T$, so magnitude follows the replacement bytes,
not whether the target latent changes. One ULP is a single amplitude, not a
universal floor. The realized-$\rho_S$ operating grid and the equivalence
test's operating characteristics are in Appendix~\ref{app:resolution}.

\textbf{Post hoc robustness.} The mean ordering $S>T$ in step-0 $L_2$ is
carried by a thin upper tail of $S$ donors (medians $0.0059$ for $S$ and
$0.0073$ for $T$). Re-scoring the frozen per-episode records by numerical
fragility (one-ULP flip) or by step-0 perturbation magnitude, post hoc and with
no rollouts added, moves $\bar\theta$ within $[-0.68,+2.21]\pp$, and every
stratum except the fragile one keeps its task-stratified paired-bootstrap 90\%
interval inside $\pm7.5\pp$. The fragile stratum ($n=46$) is uninformative.
Flips concentrate there ($3.1$--$3.2\times$ enrichment), but removing it moves
$\bar\theta$ toward zero ($+0.58\pp$).

\subsection{Trained-Plugin Content Sham}
\label{app:plugin-sham}

The trained plugin passes its offline mutation gate (Table~\ref{tab:mutation})
but lowers closed-loop success (Appendix~\ref{app:trained}). This section
reports its content sham.

\textbf{Content sham on the trained checkpoint.} The frozen five-condition instrument
of Appendix~\ref{app:payload} was reused unchanged except for the checkpoint,
on RC5 with 200 fresh seeds (1201--1400; 2{,}000 episodes;
$n_{\rm joint}=196$ after donor availability). The exact replicate changes
$0/196$ outcomes. The instrument gives $\rho_T=6.1\%$ ($12/196$),
$\rho_S=4.6\%$ ($9/196$) and $\rho_U=5.5\%$ (no count recorded; $0.055$ is not a multiple of $1/196$, so this rate is not on the joint cohort), with joint counts
$n_{00}=181$, $n_{10}=6$, $n_{01}=3$ and $n_{11}=6$, and
$\rho_T-\rho_S=+1.53\pp$. Its 95\% interval $[-1.53,+4.59]\pp$ is the
preregistered 10{,}000-replicate stratified bootstrap. The interval contains zero, so the preregistered cell
reads that the instrument cannot discriminate, and its frozen wording applies:
\emph{the instrument's discriminative power on a content-sensitive checkpoint
was not demonstrated.} The verdict is \emph{unresolved}: the estimate rests on
nine singly discordant pairs, and survival coverage is $100/196=0.51$ ($S$) and
$107/196=0.55$ ($T$), a lower bound on delivery (Appendix~\ref{app:survival}).
This does not revise the community-checkpoint result of
Appendix~\ref{app:payload}. The two verdict words differ because the
preregistered criteria differ (an equivalence test there, a one-sided
validation here), not because the two estimates disagree.

\subsection{Payload Footprints without Matched Controls}
\label{app:footprint}

Two three-condition studies swap in a frozen payload with matched occupancy and
compute path but without a latent-matched or one-ULP control, so they bound how
many paired outcomes depend on the episode's own payload and identify no
task-fact content. On SimplerEnv-Fractal (fresh 200-seed band, base seed
20260826; original perceptual bank against a frozen neutral bank with the same
16-entry occupancy and compute path), the original-versus-neutral discordance is
$21/200=10.5\%$ (exact Clopper--Pearson 95\% upper limit $15.6\%$). Because the
neutral bank is not latent-matched and no one-ULP condition was run, this also
contains any numerical sensitivity, and the code state at run time cannot be
established from records (Appendix~\ref{app:checkpoint-provenance}). On Bridge
(structured carry against frozen-payload footprint carry over 200 units: 50
chains, attempts 2--5), the discordance is $39/200=19.5\%$ (exact upper limit
$25.7\%$). Both are upper bounds on payload dependence without task-fact
specificity. Decision-time survival of the frozen payload was not audited in
either study. Bridge carry fails Q2 (only in memory) by construction, and that
condition was not measured for Fractal (Table~\ref{tab:link-profile}).

\subsection{Coordinate De-Prescription}
\label{app:coordinate}

The coordinate study asks whether behavior in the live agentic harness
(Section~\ref{sec:settings}) follows one specific field of harness-written text
memory: recipe-level object coordinates. It is not a $C/U/S/T$ ladder and has
no one-ULP or latent-matched control, and its cells S and T are not the
substitution conditions of Section~\ref{sec:arms}. Generic prompt shortening predicts that
de-prescription helps in both cells; stale coordinates predict that it helps
only when the object position changes (S) and not when the position is fixed
and the instruction redirected (T). Each cell contains 100 paired formal
episodes, and the adjudicator and interpretation table were frozen before
launch. In the S-cell, de-prescription changed success from $53\%$ to $57\%$
($b=11$, $c=15$, exact McNemar $p=0.557$); in the T-cell, from $85\%$ to $79\%$
($b=12$, $c=6$, exact McNemar $p=0.238$). The frozen Stage-1 adjudicator's
two-sided Wald test of the condition-by-S-cell coefficient in a
task-fixed-effects logistic regression gives $p=0.0816$, above the frozen
$0.05$ gate, so the verdict is unresolved and per-cell directions are
descriptive. No prospective MDE or power calculation was frozen for this
100+100 Stage-1 subset. It was a directional falsification screen for a
planned six-cell design, not a powered confirmation of a $10\pp$
difference-in-differences, and no post hoc power claim is made for it. The
intervention removed structured recipe-level coordinate fields only and left
narrative summaries unchanged, so this study does not close the
coordinate-specific mechanism or establish a general cost of removing
coordinate fields.

\FloatBarrier

\section{Outcome and Bottleneck}
\label{app:rq4}

This chapter supports Q1 (Section~\ref{sec:rq4}). Throughout, $b$ counts pairs on
which the named baseline succeeds and the intervention fails and $c$ the
reverse. Every success difference in this chapter is intervention minus the
named baseline, $\Delta=(c-b)/n$. For ChainBench the named baseline is the
no-memory condition, so $b$ and $c$ are swapped relative to Section~\ref{sec:rq4}, whose
baseline is the memory-on condition (for E1B, $b{=}95$, $c{=}0$ for decisions and
$b{=}71$, $c{=}0$ for task success).

\subsection{ChainBench E1B and E1}
\label{app:e1b}

E1B (202 chains, $K{=}2$) is primary and E1 (143 chains, $K{=}3$) supporting.
They use different executors, manifests and chance floors and are never
pooled. Both studies re-execute sealed decision records without a
decision-time language-model call. The E1B full memory configuration is
described in Appendix~\ref{app:chainbench}. Construction is in
Appendix~\ref{app:chainbench} and the executor's replay floor in
Appendix~\ref{app:executor-floors}.

\textbf{E1B (primary).} Curated records (the full-memory condition) obtain
0.7426 task success and no memory 0.3911. The paired task-success gain is
$\Delta=+0.3515$ ($b{=}0$, $c{=}71$, $n{=}202$), with a 10,000-replicate
chain-bootstrap 95\% CI $[+0.2871,+0.4158]$, whose lower bound clears the
frozen $+0.15$ materiality line. All 404 requested executions complete, both
executor and oracle-floor sentinels pass, and an independent reconstruction
has zero row disagreements. Curated-records task success equals that of the
oracle-decision condition, i.e.\ the motor-execution ceiling $\mathrm{SR}^*$ of this
executor.

\textbf{Decision accuracy.} Curated records decide correctly on $202/202$ chains
and no memory on $107/202=0.5297$. The all-$K{=}2$ floor, the decision gain
and their post-closeout status are in Appendix~\ref{app:chainbench}.

\textbf{Constructive pairing.} For each chain, all seven conditions look up
exactly two distinct physical executions, so conditions that reach the same
decision share one execution. All 202/202 chains are verified with zero extra
executions. The headline comparison is internal to the rerun. An earlier E1B
export, retired for provenance gaps, is compared only descriptively and is not
pooled. Across runs the executor is not bitwise repeatable
(Appendix~\ref{app:executor-floors}).

\textbf{Decision accuracy times execution (post hoc).} The
$\mathrm{Acc}\cdot\mathrm{SR}^*$ check of Section~\ref{sec:chainbench} is descriptive, was
run after closeout, and assumes rather than tests that a correct decision is
executed at the same rate in every condition. The initial-fact baseline selects
the correct answer on $105/202=0.5198$ chains, where the initial and final
facts agree, against $107/202=0.5297$ without memory. Its task success is
$0.3911$, equal to no memory, whereas curated records match the oracle-decision
condition ($0.7426/1.0000$). The initial-fact decision rate reflects the chain
construction, not the performance of a harness memory path. The observed
gain belongs to the evaluated full memory configuration.

\begin{table}[ht]
\centering
\scriptsize
\caption{\textbf{Complete E1/E1B baseline morphology.} ``Curated records''
denotes the full-memory condition, whose sealed decisions are replayed
through the motor bridge. The initial-fact baseline selects the answer
specified by the initial fact and ignores later updates. The distinct
stale-system condition is defined only on rewritten chains, with
study-specific denominators (E1B: all 97 of 202; E1: 72 of 143). The
stale-system condition answers incorrectly on every rewritten E1B chain
($0/97$ correct), providing the ChainBench validity control of
Section~\ref{sec:rq2}.}
\label{tab:chainbench-full-baselines}
\vspace{4pt}
{\footnotesize%
\begin{tabular}{llrrrr}
\toprule
Study & Condition & $n$ & Task success & Decision accuracy & Wrong-open rate \\
\midrule
E1 & curated records (full memory) & 143 & 0.9510 & 0.9580 & 0.0000 \\
E1 & no memory & 143 & 0.3287 & 0.3287 & 0.1888 \\
E1 & oracle decision & 143 & 0.9930 & 1.0000 & 0.0000 \\
E1 & wrong decision & 143 & 0.0000 & 0.0000 & 1.0000 \\
E1 & shuffled memory & 143 & 0.4895 & 0.4965 & 0.1329 \\
E1 & initial-fact baseline & 143 & 0.4895 & 0.4965 & 0.1119 \\
E1 & stale system & 72 & 0.0694 & 0.0694 & 0.2222 \\
\midrule
E1B & curated records (full memory) & 202 & 0.7426 & 1.0000 & 0.0000 \\
E1B & no memory & 202 & 0.3911 & 0.5297 & 0.3911 \\
E1B & oracle decision & 202 & 0.7426 & 1.0000 & 0.0000 \\
E1B & wrong decision & 202 & 0.0000 & 0.0000 & 0.8564 \\
E1B & shuffled memory & 202 & 0.3861 & 0.5149 & 0.4059 \\
E1B & initial-fact baseline & 202 & 0.3911 & 0.5198 & 0.4010 \\
E1B & stale system & 97 & 0.0000 & 0.0000 & 0.8351 \\
\bottomrule
\end{tabular}}
\end{table}

\textbf{E1 (supporting).} E1 re-executes the sealed decisions on a frozen
143-chain manifest ($K{=}3$, $1/3$ fallback floor). The paired task-success
gain is $\Delta=+0.6224$ ($b{=}3$, $c{=}92$, $n{=}143$; 10,000-replicate
chain-bootstrap 95\% CI $[+0.5315,+0.7063]$), whose lower bound passes the
frozen $+0.30$ materiality gate. Execution given a correct curated-records
decision is 0.9927. Execution and integrity checks pass, and an independent
reconstruction
disagrees with the producer on zero rows. The manifest is a rebuild, not an
attrition subset: the immutable manifest diff shows an earlier 150-item
manifest changed by removing 51 items (all motor-infeasible bottom-drawer
finals) and adding 44. The rebuild was not documented contemporaneously and its
criterion, motor feasibility rather than observed outcome, was not frozen in
advance, so selection bias remains unresolved and we make no population-level
claim beyond this manifest.

Post hoc sensitivity checks give $\Delta=+0.626$
without the 44 additions (95\% CI $[+0.525,+0.727]$), $+0.459$ with the 51
removals reinstated as failures in both conditions ($[+0.387,+0.531]$), and $+0.196$
($[+0.072,+0.314]$) under the physically impossible assignment of all 51 as
no-memory-only wins, below the frozen $+0.30$ gate but positive. The additions
hold 28 of the 92 favorable discordances, essentially the hypergeometric
expectation 28.31. E1B mitigates but does not erase this lineage concern. The
decision fields come from the sealed v2 cohort evaluation (raw rows, report,
independent audit). Its still-earlier store and WriterBench inputs are not
regenerated, so E1 is a re-execution, not a new decision-generation campaign.

\subsection{Learned-Bank Outcome Contrasts}
\label{app:learned-outcomes}

Two learned-bank settings contrast task success under a memory intervention
without any decision readout, so their success is not decomposable and
neither can say where a bottleneck lies: the SimplerEnv-Fractal perceptual
bank within the episode, and Bridge five-attempt carry chains across episodes.
Payload-footprint follow-ups for both are in Appendix~\ref{app:footprint}.

\textbf{Fractal perceptual-bank intervention.} The SimplerEnv-Fractal
\citep{simplerenv} coke-can evaluation covers all twelve cells defined by the
official task script: three can orientations across four URDF versions, 300
seed-paired episodes. A first eight-cell run (200 pairs) omitted the
laid-vertically orientation because the launch script hard-codes the
cell loop. On the four completing cells the unablated configuration succeeds
on $94.0\%$ against $89.5\%$ on the original eight, so the omission did not
inflate the eight-cell figure. Over all twelve cells the Fractal VM/VA
configuration succeeds on 91.0\% and the configuration without its test-time
perceptual bank on 92.7\%, so the ablation raises success by
$\Delta=+1.7\pp$ ($b{=}8$, $c{=}13$; exact McNemar $p=0.38$), against
$+3.0\pp$ over the earlier eight cells. Twenty-one pairs flip ($\rho=7.0\%$,
exact Clopper--Pearson 95\% CI $[4.4\%,10.5\%]$), while the independent
replicate control flips 0/300. Because the discordance interval crosses the
frozen 5\% activity boundary, the preregistered rule returns no activity
verdict (band-straddling), and the outcome contrast is not detected at this
design's resolution. It supports neither a benefit of ablation nor
equivalence. The intervention removes only the test-time perceptual bank,
leaving the cognitive bank untouched. It does not evaluate a model trained
without memory. The twelve cells
share one backbone and checkpoint, pooled for the frozen endpoint rather than
as independent cells. This campaign and its footprint follow-up ran against a
shared, unversioned overlay source tree whose code state at run time is not
bound, so the results are implementation-bound
(Appendix~\ref{app:checkpoint-provenance}). A re-run under the archived launch
arguments and base seed reproduces cells 000 and 001 ($50/50$ episodes each in
both conditions, zero disagreements), which licenses behavioural attributability for
those two cells only and does not establish code identity.

\textbf{Bridge five-attempt carry chains.} The confirmatory Bridge study ran
four conditions over the same 100 chains and five attempts per chain (2,000 episodes
total; zero execution failures): fresh episode memory, whole-bank token-merge
carry, structured/versioned carry, and a hybrid that resets its episode-local
token-merge bank while persisting a structured store. Each seed-paired chain
rotates the object position across depth, so no earlier attempt holds
information that only memory can supply and the setting fails Q2 (only in memory) by
construction. The four conditions share condition-independent diffusion seeds. Fresh,
token-merge and hybrid carry agree on every paired first attempt, a
determinism check rather than a sham. The mechanism-distinct structured condition
differs on 10 first-attempt outcomes, cause not diagnosed. The frozen primary
endpoint pooled attempts 4--5 and was tested with
exact McNemar tests under Holm correction (Table~\ref{tab:app-bridge}). No carry
condition changes success detectably (all Holm $p{=}1.0$), although token-merge carry
changes $30.5\%$ of paired deep-chain outcomes. Without matched shams these
contrasts remain descriptive about what the carried content does. A post hoc
chain-clustered reanalysis leaves the conclusion unchanged: for token-merge,
structured/versioned and hybrid carry the chain-bootstrap 95\% intervals are
$[-10.5,+5.0]$, $[-7.5,+6.0]$ and $[-4.0,+6.5]\pp$, chain-level sign-flip $p$
values $0.607/1.000/0.854$, and intraclass correlations $-0.00/+0.10/-0.07$.
Over all 500 outcomes, a non-primary view, token-merge carry changes success by
$-4.4\pp$ (exact McNemar $p=0.051$). This is reported for transparency and is
not promoted over the frozen endpoint.

\begin{table}[ht]
\centering
\scriptsize
\caption{\textbf{Frozen deep-chain endpoint.} Attempts 4--5 give 200 paired
outcomes per comparison. The baseline is fresh reset, $\Delta$ is condition minus
fresh, and $b$ counts pairs where fresh succeeds and the condition fails.}
\label{tab:app-bridge}
\vspace{4pt}
{\footnotesize%
\setlength{\tabcolsep}{4pt}%
\begin{tabular}{lrrrrrrl}
\toprule
Condition vs.\ fresh & Fresh SR & Condition SR & $\Delta$ & $b$ & $c$ & exact McNemar $p$ & Holm $p$ \\
\midrule
Token-merge carry & 31.5\% & 29.0\% & $-2.5\pp$ & 33 & 28 & 0.609 & 1.000 \\
Structured/versioned carry & 31.5\% & 31.0\% & $-0.5\pp$ & 22 & 21 & 1.000 & 1.000 \\
Hybrid carry & 31.5\% & 32.5\% & $+1.0\pp$ & 14 & 16 & 0.856 & 1.000 \\
\bottomrule
\end{tabular}}
\end{table}

\subsection{Released versus Structured-PCMB Configuration Contrast}
\label{app:mikasa}

Table~\ref{tab:app-mikasa} compares, in the second formal
five-task campaign, the
released MemoryVLA method \citep{memoryvla} (episode-local banks, no cross-episode persistence) with
persistent PCMB (structured PCMB, scope-keyed persistence). The
switch changes memory architecture and cross-episode persistence together, a
configuration contrast, not an ablation. Each task pools attempts 1 and 2 over
100 task--seed scopes. Attempt~1 starts from cleared memory in both conditions
(Appendix~\ref{app:payload}) and contributes only an architecture contrast. The
campaign had no frozen analysis plan, so every entry is descriptive, and
discordance cannot be localized to retrieved content or to cross-episode
carry. Bindings are in Appendix~\ref{app:checkpoint-provenance}. On the
released path, bank entries are (timestep, feature) pairs ranked by symmetric
patch similarity, with timestamps used only to break ties, so evaluator-only
fields are not even stored in bank entries. Persistent PCMB with execution memory
reproduces the structured rates on all five tasks. Two independently launched
same-function configurations agree on all 1,000 outcomes and three bank diagnostics
(Appendix~\ref{app:twins}). This is a determinism control, not a content experiment,
and provides no evidence on whether bank content is used.

\begin{table}[ht]
\centering
\small
\caption{\textbf{MIKASA released versus structured-PCMB configuration
contrast.} Each condition and task has 200 episodes (attempts 1--2 pooled). $\Delta$ is
structured minus released. No task shows a detected net effect, although four
non-ceiling tasks have double-digit paired discordance. The pooled row is
descriptive (one backbone and checkpoint, not independent replicates).}
\label{tab:app-mikasa}
\vspace{4pt}
\begin{tabular}{@{}lrrrrr@{}}
\toprule
Task & released SR & structured SR & $\Delta$ & $\rho$ & exact McNemar $p$ \\
\midrule
Intercept Memory (IM) & 20.5\% & 25.5\% & $+5.0\pp$ & 18.0\% & 0.133 \\
Remember Color 3 (RC3) & 34.0\% & 32.5\% & $-1.5\pp$ & 17.5\% & 0.736 \\
Remember Color 5 (RC5) & 23.5\% & 23.0\% & $-0.5\pp$ & 18.5\% & 1.000 \\
Remember Color 9 (RC9) & 13.0\% & 12.5\% & $-0.5\pp$ & 12.5\% & 1.000 \\
Shell Game (SG) & 89.5\% & 89.5\% & $0.0\pp$ & 3.0\% & 1.000 \\
\midrule
Pooled, 5 tasks ($n{=}1{,}000$) & 36.1\% & 36.6\% & $+0.5\pp$ & 13.9\% & 0.735 \\
\bottomrule
\end{tabular}
\end{table}

\subsection{Trained Plugin Closed Loop and Factorial}
\label{app:trained}

We trained the memory components of the MemoryVLA MIKASA checkpoint
(retrieval/fusion plus a zero-initialized execution adapter; 427.7M trainable
parameters over a frozen 7B backbone) on the benchmark's demonstrations.
Training sharply improves held-out diffusion loss and passes the offline
content-mutation gate (Table~\ref{tab:mutation}). The content sham on this
checkpoint is in Appendix~\ref{app:plugin-sham}. The preregistered
1,000-rollout closed-loop study reverses the conclusion
(Table~\ref{tab:app-trained}).

\begin{table}[ht]
\centering
\small
\caption{\textbf{Offline mutation gate} (held-out diffusion loss). The trained
components separate true, empty and mismatched memory offline, but this did not
predict closed-loop success (Table~\ref{tab:app-trained}).}
\label{tab:mutation}
\vspace{4pt}
\begin{tabular}{lccc}
\toprule
Held-out diffusion loss & true memory & empty & mismatched \\
\midrule
MemoryVLA MIKASA checkpoint & 0.658 & 0.655 & 0.656 \\
trained components & \textbf{0.123} & 0.228 & 0.148 \\
\bottomrule
\end{tabular}
\end{table}

\begin{table}[ht]
\centering
\small
\caption{\textbf{Closed-loop result for the trained memory plugin.} 200 paired
episodes per task (zero execution failures). Community is the
released MemoryVLA condition of Table~\ref{tab:app-mikasa}, a cross-launch
archived condition. $\Delta$ is trained minus community. The paired-test method is
not recorded. $^\dagger$Rates are rounded as exported. $\Delta$ uses unrounded counts.}
\label{tab:app-trained}
\vspace{4pt}
\begin{tabular}{lrrrr}
\toprule
Task & trained SR$^\dagger$ & community SR$^\dagger$ & exact $\Delta$ (pp) & paired test \\
\midrule
Shell Game & 10\% & 90\% & $-79.0\pp$ & $p<10^{-6}$ \\
Remember Color 5 & 8\% & 24\% & $-15.5\pp$ & $p=0.0003$ \\
Remember Color 3 & 24\% & 34\% & $-10.5\pp$ & $p=0.051$ \\
Intercept Memory & 24\% & 20\% & $+3.5\pp$ & n.s. \\
Remember Color 9 & 14\% & 13\% & $+1.0\pp$ & n.s. \\
\midrule
Pooled & 16.0\% & 36.1\% & $-20.1\pp$ & $p<10^{-6}$ \\
\bottomrule
\end{tabular}
\end{table}

Against the untrained structured condition of Table~\ref{tab:app-mikasa} ($36.6\%$),
archived from that same separately launched campaign and therefore also
cross-launch, the trained model scores 16.0\% ($-20.6\pp$, $p<10^{-6}$). Two
differences between training and deployment are measured: demonstrations
contain approximately 12--20 frames, whereas policy rollouts last
approximately 60--90 steps, and training banks contain expert trajectories
while deployment banks contain the policy's own trajectories. Covariate shift
between the two \citep{dagger}, in state and in the memory channel, is a
hypothesis consistent with these diagnostics, but it was not tested. The
diagnostics do not identify a unique cause.

\textbf{Factorial.} The fresh preregistered factorial uses a different seed
queue from the 1,000-rollout study (100 fresh seeds per task, 1101--1200) and
resolves the bundled comparison into four paired conditions ($n=500$ each):
trained+bank $0.102$, trained+empty $0.184$, community+bank $0.398$, and
community+empty $0.414$. The 95\% intervals below are those recorded by the
frozen adjudicator. The precommit names the interval method only for the
difference-in-differences, a task-stratified paired bootstrap over
seed-matched condition quadruples (10,000 replicates). The two recipe contrasts
(trained minus community) are $-29.6\pp$ with the bank ($[-34.8,-24.6]\pp$;
$b{=}180$, $c{=}32$) and $-23.0\pp$ with an empty bank ($[-28.0,-17.8]\pp$;
$b{=}159$, $c{=}44$). The trained-recipe bank-channel contrast (bank minus
empty) is $-8.2\pp$ ($[-12.6,-4.0]\pp$; $b{=}80$, $c{=}39$): a detected
effect of bank presence, which does not isolate content. The
difference-in-differences is $-6.6\pp$ ($[-12.4,-0.8]\pp$). The frozen rule
returns an interaction verdict: the difference-in-differences interval
excludes zero, but the channel interval's upper end ($-4.0\pp$) does not pass
the separate frozen $-5\pp$ materiality gate, so attribution is limited to the
observed interaction and no family-wise claim is made. The $-23.0\pp$ recipe
effect is a separate estimand on fresh seeds and is not combined with the
$-20.1\pp$ or $-20.6\pp$ contrasts above.

\FloatBarrier

\section{Self-Correction Ledger}
\label{app:excluded}

This chapter records what does not count as evidence for any of the four questions, and the
post-data amendments and post hoc analyses that qualify the evidence that does.
It adds no claim.

\subsection{Withdrawn Analyses}
\label{app:withdrawn}

\textbf{Shell Game full-vision content ladder.} The preregistered five-condition
$C/U/S/T$ ladder completed $396/400$ joint
scopes on two 200-scope Shell Game bands. The reference policy succeeds on
$90.91\%$ against the analytic three-cup floor of $33.33\%$. The exact
replicate $C$ changes $0/396$ outcomes, the latent-matched $S$ and
different-fact $T$ conditions change $12/396$ each, $\theta=\rho_T-\rho_S=0.00\pp$
and $D_T=\mathrm{SR}_A-\mathrm{SR}_T=+0.51\pp$. The ladder, including these
numbers, is withdrawn as Q3/Q4 evidence for three reasons. Its protocol
registered no decision-time survival gate. The post hoc survival audit finds
that every decision query selects at least one unmodified substituted record in
only $0.61$--$0.68$ of retries, a lower bound on delivery
(Appendix~\ref{app:survival}). Shell Game success counts any touch of the
correct cup, so the readout is not a first-choice readout
(Appendix~\ref{app:sg-anytouch}). In Table~\ref{tab:link-profile} this Q3/Q4 entry
therefore reads ``withdrawn''. The remaining full-vision Shell Game statements
are presence statements, not content statements
(Appendices~\ref{app:sg-anytouch} and~\ref{app:readpath}).

\subsection{Excluded and Superseded Studies}
\label{app:twins}

Table~\ref{tab:app-excluded} lists studies whose design or provenance cannot
support an inferential claim. Negative preregistered studies remain in the
results.

\begin{table}[ht]
\centering
\scriptsize
\caption{\textbf{Excluded and superseded studies.} None supports a claim in
this paper.}
\label{tab:app-excluded}
\vspace{4pt}
{\setlength{\tabcolsep}{4pt}%
\begin{tabular}{@{}>{\raggedright\arraybackslash}m{0.30\textwidth}>{\raggedright\arraybackslash}m{0.38\textwidth}>{\raggedright\arraybackslash}m{0.27\textwidth}@{}}
\toprule
Study & Why excluded & Consequence \\
\midrule
Payload-swap v1/v2 & path or admissibility gates invalid & excluded before v3 \\
Payload-swap v3 (incl.\ stage B) & no matched control; outcome-exposed seeds; third-task assignment failed closed under the donor rule & replaced by the formal five-condition RC3/RC5 campaign; not pooled \\
Five-task configuration contrast, first run & aborted: the structured persistent clock was rewound by the episode-local timestep, so newly written slots could be evicted before action attachment & partial runs preserved, none used; the reported run is the second run, launched after the clock fix passed a full-horizon smoke; no claim from v1 \\
Same-function adapter toggle (1,000 episodes) & zero-initialized path absent from checkpoint; agreement by construction & determinism control only; not evidence on whether bank content is used \\
Version-conflict precursor (1,200 rollouts) & stale condition vs.\ unmodified $+1.7\pp$, vs.\ no-conflict control $+4.0\pp$; could not resolve a few-point effect & superseded by the frozen 690-pair test; no claim \\
Earlier E1B export & provenance gaps; verdict retired & not pooled with the E1B rerun; no claim \\
Counterfactual-twin phase 2 & never reached formal evaluation & never qualified; no claim \\
9.22M-parameter additive successor & 16/16 mechanism checks, outputs bitwise unchanged before training; training and six-condition closed loop not run & no result \\
WriterBench-X & development benchmark, not closed loop & no claim \\
Development pilots (Bridge deep-chain, $n\!\approx\!24$ cells; version replacement; additive C-lite J/K/L; exposed-fold gates; single-seed LIBERO) & underpowered, not run, or development-exposed & no claim \\
Camera-onset probe and E2-mini ($n=30$/cell) & exploratory; informed the blackout onset & design input only; onset curve not cited \\
E0-S cue-substitution calibration ($n=30$) & exploratory; possible image--state mismatch & never frozen; no formal run \\
E0 version-1 runner (smoke) & attempt counter never reset; gate passed vacuously & replaced by version 2 before any formal unit \\
\bottomrule
\end{tabular}}
\end{table}

\subsection{Post-Data Amendments and Post Hoc Analyses}
\label{app:amendments}

Protocols and decision rules of the preregistered campaigns cited as evidence
were frozen before launch, except for the amendments in
Table~\ref{tab:app-amendments}, written after smoke or formal data, during a
run, or at launch. Studies without a separately frozen plan on record, among
them the ChainBench v1 diagnostic (Appendix~\ref{app:chainbench}), the
five-task configuration contrast (Appendix~\ref{app:mikasa}) and the writer
study (Appendix~\ref{app:writer-audit}), are marked where they are reported, as
is the E1 manifest-rebuild criterion, which was not frozen in advance
(Appendix~\ref{app:e1b}). Amendments made before any data of the amended
campaign, and amendments of excluded or withdrawn campaigns, are recorded only
in each campaign's precommit and closeout records.
\textbf{ChainBench baseline naming correction.} A post hoc implementation
audit identified the condition stored as \texttt{harness\_task\_memory} as a
fixed initial-fact baseline. Earlier wording incorrectly described it as
harness-written memory. We correct the label and remove the associated
memory-delivery attribution. Numerical results are unchanged
(Appendix~\ref{app:e1b}).

The four-question framework (IMEP), its criteria and the first-failing-condition
rule were formulated after every campaign had closed and are listed with the other post
hoc analyses.

\begin{table}[tp]
\centering
\scriptsize
\setlength{\tabcolsep}{3pt}
\caption{\textbf{Post hoc analyses and amendments.} Post hoc analyses ran on
frozen records after the campaign's closeout. None adds rollouts or feeds a
registered gate. ``After data'' means that outcome or gate data of the amended
campaign existed when the amendment was written. Unless a row says otherwise,
no number reported in this paper moved.}
\label{tab:app-amendments}
\vspace{4pt}
\begin{tabular}{@{}>{\raggedright\arraybackslash}m{0.23\textwidth}>{\raggedright\arraybackslash}m{0.67\textwidth}>{\raggedright\arraybackslash}m{0.05\textwidth}@{}}
\toprule
Item & What was done; effect on reported numbers & App. \\
\midrule
\multicolumn{3}{@{}l}{\emph{Post hoc analyses}} \\
Four-question framework and criteria &
The detectability precondition, Q1--Q4, their criteria and the first-failing-condition rule were written after every campaign had closed; a criterion that no campaign registered is post hoc. & \ref{sec:metric} \\
Decision-time survival audit &
Coverage per cross-episode content-substitution campaign; none registered a survival gate (the blackout ladders did), so coverage is a lower bound on delivery. & \ref{app:survival} \\
All-$K{=}2$ floor (E1B) &
The closeout used a mixed-$K$ floor; all 202 chains have $K{=}2$, so the floor is exactly $0.500$ and the gain $+0.470$ is reported against it. & \ref{app:chainbench} \\
Stratified re-scoring (RC fixed path) &
$\bar\theta$ re-scored by one-ULP fragility and by step-0 perturbation magnitude; task-stratified paired bootstrap. & \ref{app:payload} \\
First-step $L_2$ (RC fixed path) &
First-step $L_2$ distances of $U$, $S$ and $T$ (means; medians for $S$ and $T$); not preregistered endpoints. & \ref{app:payload} \\
$U$ versus $S$ and $T$ (RC fixed path) &
Exact McNemar on the joint pairs ($U$-only/$T$-only $25/34$, $p=0.30$; $U$-only/$S$-only $27/32$, $p=0.60$), computed for this manuscript; not an equivalence test. & \ref{app:payload} \\
E0 full-view difference $D_N^{W0}$ &
Difference of two preregistered descriptive success rates, with its exact bound ($3.42\pp$), derived after data; decides no verdict. & \ref{app:sg-anytouch} \\
Q2-XA first-visit counts &
First-visited cup correct in $88/177$ (reads on) and $88/178$ (reads off) successes, counted after adjudication; no gate uses them. & \ref{app:sg-anytouch} \\
Exact bounds (Q2-XA, E2-W0) &
Exact upper bounds (one-sided 95\% Clopper--Pearson components, coverage $\geq90\%$) added beside the recorded paired-bootstrap bounds: Q2-XA $3.14$, $3.65\pp$ (bootstrap $1.5$, $2.5\pp$); E2-W0 $2.94\pp$ at $b=c=2$ (bootstrap $1.5\pp$, anti-conservative at so few discordant pairs); verdicts unchanged. & \ref{app:readpath} \\
Realized-floor decision rates &
Operating characteristics re-simulated at $\rho_S=0.13$, $n=193{+}195$, $\phi=0.3$. & \ref{app:resolution} \\
E1B $\mathrm{Acc}\cdot\mathrm{SR}^*$ check &
Predicted against observed task success for the three non-oracle conditions; descriptive. & \ref{app:e1b} \\
E1 manifest-rebuild sensitivity &
Deleting the added chains, reinstating the removed chains as failures, an adversarial reassignment, and an enrichment check of the additions; none preregistered. & \ref{app:e1b} \\
Chain-clustered reanalysis (Bridge) &
Chain bootstrap, sign-flip $p$ and intraclass correlations for attempts 4--5; conclusion unchanged. & \ref{app:learned-outcomes} \\
Interval relabeling &
Labels only, after data: a difference bound built from two one-sided 95\% components is labeled coverage $\geq90\%$ (R27-A, E2-W0, Q2-XA, E0); for GRID-1a ($\alpha=0.025$ per component) each one-sided endpoint is labeled $\geq95\%$ and the two-sided interval $\geq90\%$. & \ref{app:statistics} \\
\midrule
\multicolumn{3}{@{}l}{\emph{Amendments}} \\
Ladder amendment 3 (blackout ladder, $G_{\text{sham-valid}}$) &
\emph{After formal data.} The registered sham-validity fraction divided by all reads, not reads with history, so it equalled $22/23$ whatever the outcomes and the check failed as registered; the denominator was corrected, the check merged with the substitution-coverage gate, and the sealed root re-adjudicated without a rerun. The original (INVALID) and first amended adjudications are retained; both split $\theta$ as $b'=c'=6$, and the reported split and its $p$ come from the second amended one ($\hat\theta$ unchanged). & \ref{app:blackout-ladders} \\
Ladder amendment 4 (structured ladder) &
\emph{After structured smoke, before formal data.} Gate criteria selected by memory architecture; identical empty-read positions across conditions replace a predicted count. & \ref{app:blackout-ladders} \\
Ladder amendment 2 &
\emph{After released-ladder data; not applied to it.} Structural change to the protection against reading outcomes before adjudication. & \ref{app:blackout-ladders} \\
Released ladder: precommit vs.\ launch &
\emph{At launch, before data.} The precommit planned $U$ ($n=200$) and $C_{\rm det}$ ($n=50$) in wave~2 and $N_{\rm blk}$ in wave~1; all conditions launched at 400 seeds. Its ``same-latent sham'' and ``different-latent swap'' are implemented as a within-latent permutation and a moment-matched resample. & \ref{app:blackout-ladders} \\
Grid amendment 1 ($G_{\text{seed}}$) &
\emph{After smoke \#7 adjudication.} The registered expected torch seed ($4096$) was voided; smoke \#7 moved from failing to passing through exactly four gates, all $G_{\text{seed}}$; the formal adjudication records this. & \ref{app:resolution} \\
GRID-1a-02 &
\emph{Timing not recorded.} Five negative fixture cases added (64 to 69). & \ref{app:resolution} \\
Grid amendments 4 and 5 &
04: a failed soft gate was not reported in formal mode; the fix must land before any further formal run. 05, \emph{after formal data}: the determinism companion's whole-argv equality became three conjuncts; replicate coverage is $32/96$ pairs. & \ref{app:resolution} \\
Version-conflict amendment 1 &
\emph{After a smoke gate failure, before formal launch.} Adjudicator re-registered with an explicit task list and non-vacuous $G_{\text{det}}$, $G_{\text{empty}}$. & \ref{app:version} \\
Version-conflict amendment 2 (orchestration) &
\emph{During the run; outcomes not read.} GPU-token dispatch failed 9 of 12 cells before they produced data; they were rerun from a static device list with a per-line integrity gate. & \ref{app:version} \\
Version-conflict amendment 3 &
\emph{After data; no number moved.} The first adjudication was INVALID (preflight products sought in the wrong directory); adjudicator unchanged, link added; estimands bit-identical. & \ref{app:version} \\
E0 amendment 2 &
\emph{After smoke data.} The read-side implementation became an ungated agreement diagnostic, removing a passable gate. & \ref{app:sg-anytouch} \\
E1B amendment 2 &
\emph{After attempt-1 data.} Attempt~1 failed on executor infrastructure; its verdict was voided (INVALID); sentinels added (execution success $\geq0.95$, oracle floor $\geq0.50$); scientific criteria unchanged. & \ref{app:e1b} \\
\pihalf{} fragility profile &
\emph{After data.} Preregistered mixture split $55/45$; the measured $9/91$ is a parameter revision, not a confirmed prediction. & \ref{app:executor-floors} \\
\bottomrule
\end{tabular}
\end{table}

\section{Reproducibility and Artifact Binding}
\label{app:reproducibility}

\subsection{Artifact Bindings}
\label{app:artifacts}

Each formal campaign is stored as an immutable evidence unit containing the
preregistered protocol, launch packet, environment and checkpoint binding,
per-episode outcomes, aggregation script, logs, and a SHA256 manifest.
Paper-facing tables are generated from these units. For all paired closed-loop
studies, episode-level files retain seed, task, condition, success, discordance
direction, execution status and the memory diagnostics needed to reconstruct
$b$, $c$, $D$ and $\rho$. Failed episodes are counted and reported rather than
retried. The formal Bridge and trained-plugin campaigns reported zero execution
failures. Exploratory development data, unresolved tests and invalidated
attempts are stored apart from confirmation artifacts
(Appendix~\ref{app:excluded}). Checkpoint and implementation checksums per
campaign are in Appendix~\ref{app:checkpoint-provenance}.

\subsection{Cross-Machine Adjudication and Independence Scope}
\label{app:adjudication}

We do not claim a cross-machine recompute for every campaign. For E0, E2-W0,
Q2-XA, R27-A and Q3-VC, the registered adjudicator was rerun on a second
machine from sealed inputs whose recorded SHA256 checksums were verified first,
and it returned identical verdicts, gates and estimands. For R27-A, a
reimplementation that shares no code with the registered adjudicator also
reproduced its paired counts and per-condition success rates.

For GRID-1a, the
delivered smoke and closeout files were compared byte for byte and the fixture
suite and check battery were re-run, all passing. This is not a recompute of
the formal adjudication. Neither Shell Game blackout ladder
(Appendix~\ref{app:blackout-ladders}) has a cross-machine recompute on record:
each adjudication was run once, triggered by the review session rather than by
the session that ran the conditions, and its statistical self-check agrees with a
separate implementation that shares no code path, which rules out an arithmetic
slip but not an error both implementations share. The post-data criterion
change is listed in Appendix~\ref{app:amendments}.

No cross-machine recompute
is claimed for the MIKASA substitution and configuration contrast, Bridge,
Fractal, the trained plugin, the original version-conflict campaign, ChainBench
E1/E1B or the writer study. Where an independent audit is reported for them, it
is a second implementation on the same records. All recomputes were run by the
AI sessions described in the AI use statement, so ``independent
implementation'' means two code paths, not human--machine independence. The
R27-A and Q3-VC closeouts record theirs as reported by the review session
rather than performed first-hand by the session that wrote the closeout.

\subsection{Writer-Study Checksum Boundary}
\label{app:writer-audit}

The writer closed-loop study (60 chains, five conditions, \pihalf{} executor, bounded
memory budget) compares write-time consolidation with FIFO eviction under two
write keys. Its four predictions were specified before rollout. With the
oracle (perfect entity) key, consolidation gains $+0.183$ in task success over
FIFO (paired bootstrap, 10{,}000 resamples, 95\% CI $[+0.100,+0.283]$); with
the learned perceptual key it gains $+0.033$, with a CI that includes zero.
Both keys are always reported together, and the study enters no headline claim.
Because its protocol and result note share one legacy document, the study was
audited rather than rerun: the audit launched no rollout, read the original run
root only, and its aggregation of the raw rows reproduces the originally
reported condition means and prediction verdicts. No launch-time checksum list
survived, so eight checksums were established at audit time: the four raw-row
shards, the protocol document, two code files and the \pihalf{} checkpoint. The
original bootstrap seed was not archived. The recomputed interval differs from
the first computation, $[+0.083,+0.283]$, by $0.017$ at the lower endpoint,
and exact reproduction of the original interval is not possible without
the missing bootstrap seed.

\section{Memory Maintenance at the Decision Level}
\label{app:memory-maintenance}

\textbf{Scope.} These additional ChainBench comparisons examine memory states
and retrieval rules at the decision level. All results in this section are
non-confirmatory. They concern decision accuracy, not robot task success,
and are not pooled with the E1B execution results in
Table~\ref{tab:chainbench}.

\textbf{Relation to E1B.} The experiment audit links the frozen 202-chain
E1B manifest to a subset of Run~A's 403 chains. The ten decision fields agree
with the corresponding evaluation rows on all applicable chains: 202 for
conditions defined on every chain and 97 for conditions restricted to
rewritten chains. Initial facts, final facts, and rewrite flags also agree.
The full memory configuration below is therefore the source of the sealed
E1B decisions described in Appendix~\ref{app:chainbench}. Run~B contains a
different cohort of 138 chains. The comparisons here use the full evaluation
cohorts, whereas the reported E1B task success comes from the August 25 motor
rerun of the 202-chain subset.

\textbf{Memory states and readout.} On rewritten chains, \emph{superseded
write} retains only the old record. \emph{Latest write} retains only the new
record, selected by episode role rather than by the fact label. This is a
stipulated memory state, not the output of an evaluated eviction policy.
\emph{Read-time gate} retains both records without using their provenance and
selects by query similarity. \emph{Retrieval} applies two-stage selection to
the full memory set, including six same-scene cross-chain distractor writes.
\emph{Full system} combines the latest-write memory set, a calibrated presence
gate, and two-stage retrieval. The retrieval-based conditions use a
leave-one-chain-out nearest-centroid readout whose centroids exclude the
query chain's events. No language-model call is made at decision time.

\textbf{Condition names.} Table~\ref{tab:maintenance} uses readable labels
for the archived keys. Latest write, read-time gate, and superseded write
correspond to \texttt{write\_time\_eviction},
\texttt{read\_time\_gate\_best}, and \texttt{stale\_memory}, respectively.
Full system, retrieval, oracle memory, no memory, and shuffled records
correspond to \texttt{full\_system}, \texttt{retrieval},
\texttt{oracle\_memory}, \texttt{no\_memory}, and
\texttt{shuffled\_memory}. The superseded-write condition differs from the
initial-fact baseline in Table~\ref{tab:chainbench}, which directly selects
the initial answer without invoking retrieval.

\begin{table}[!ht]
\centering
\small
\caption{\textbf{Memory-state and retrieval contrasts (non-confirmatory).}
Each contrast is the first condition minus the second. Differences and
interval endpoints are absolute percentage-point changes in decision
accuracy. Intervals are reported 95\% chain-bootstrap percentile intervals
with 10,000 draws and seed 20260723. ``Rewritten'' restricts the cohort to
chains with a superseded fact. Run~A includes the E1B executed subset;
Run~B is a distinct cohort.}
\label{tab:maintenance}
\setlength{\tabcolsep}{4pt}
\begin{tabular}{@{}lcrrr@{}}
\toprule
Contrast & Cohort & $n$ & Difference (pp) & 95\% CI (pp) \\
\midrule
\multicolumn{5}{@{}l}{\emph{Run A}} \\
Retrieval $-$ no memory & all & 403 & $+5.71$ & $[-0.74, +12.16]$ \\
Retrieval $-$ shuffled records & all & 403 & $+5.21$ & $[-1.49, +12.16]$ \\
Oracle memory $-$ retrieval & all & 403 & $+43.67$ & $[+38.71, +48.64]$ \\
Latest write $-$ read-time gate & rewritten & 204 & $+11.27$ & $[+7.35, +15.69]$ \\
Superseded write $-$ no memory & rewritten & 204 & $-47.06$ & $[-53.92, -40.20]$ \\
Full system $-$ no memory & all & 403 & $+48.88$ & $[+43.92, +53.85]$ \\
\midrule
\multicolumn{5}{@{}l}{\emph{Run B}} \\
Retrieval $-$ no memory & all & 138 & $-0.72$ & $[-12.32, +10.87]$ \\
Retrieval $-$ shuffled records & all & 138 & $-1.45$ & $[-13.04, +10.14]$ \\
Oracle memory $-$ retrieval & all & 138 & $+44.20$ & $[+35.51, +52.17]$ \\
Latest write $-$ read-time gate & rewritten & 69 & $+14.49$ & $[+7.25, +23.19]$ \\
Superseded write $-$ no memory & rewritten & 69 & $-59.42$ & $[-71.01, -47.83]$ \\
Full system $-$ no memory & all & 138 & $+43.48$ & $[+35.51, +51.45]$ \\
\bottomrule
\end{tabular}
\end{table}

\textbf{Interpretation.} In both runs, the intervals for retrieval versus
no memory and shuffled records include zero. These comparisons do not
establish equivalence. Superseded-write memory performs worse than no
memory, and the latest-write state performs better than choosing between
both records at read time. These are comparisons between the specified
configurations. In particular, full system and retrieval differ in both
memory state and presence gating, so their contrast cannot isolate the
effect of a single component. The latest-write result does not measure the
ability of a deployed system to perform maintenance.

\textbf{Evidence status and limits.} The evaluation reports mark these runs
as non-confirmatory and constrain their interpretation to a decision-level
retrieval-to-action rule. The 403- and 138-chain accuracy contrasts are not
robot task-success contrasts and are not differenced against the executed
202-chain rates. This restriction does not imply that no decisions from
these evaluations were subsequently executed: E1B replays a subset of Run~A.
The experiment audit reports reproducing all twelve contrasts and their
bootstrap intervals from the per-row records. The E1B executor is bound by
a frozen digest, but the run artifacts do not record the source digest of
the decision evaluator. Consequently, the exact historical evaluator code
version is unresolved even when archived decisions and derived statistics
agree. These additional comparisons contribute no confirmatory verdict.

\section{Related Work (Extended)}
\label{sec:related}

\textbf{Memory in VLA systems and agentic harnesses.} Memory extends foundation
policies \citep{pi0,pi05,rt2,openvla,octo} through perceptual and cognitive
banks with token merging \citep{memoryvla}, event-triggered keyframes
\citep{kemo}, and keyframe retrieval by a trained planner \citep{memer}.
HAMLET summarizes each timestep into learned moment tokens and aggregates them
with a lightweight temporal memory module \citep{hamlet}. Harness VLA uses
memory-guided primitives to steer frozen policies \citep{harnessvla}, while
dual-system designs separate deliberation from control
\citep{groot,helix,hirobot}. Reflexion, ExpeL, and AgeMem also retain experience
for subsequent agent decisions \citep{reflexion,expel,agemem}. Component
ablations can establish the utility of these complete designs; attributing
that utility to a particular stored fact requires additional controls.
History is not automatically useful evidence: causal confusion in imitation
learning \citep{causalconfusion} and the copycat problem \citep{copycat} show
how policies can exploit historical correlates rather than the information
needed for the next action. These motivate our attribution question without
establishing that the same failure occurs in the systems we evaluate.

\textbf{Memory benchmarks.} MIKASA-Robo \citep{mikasarobo}, RoboMME
\citep{robomme}, and RMBench \citep{rmbench} evaluate memory-dependent robotic
manipulation. RoboMME-Interference examines interference from other sessions
\citep{robommeinterference}. LongMemEval and LoCoMo evaluate long-term
conversational memory \citep{longmemeval,locomo}. MemoryAgentBench supplies
incremental multi-turn evaluations, including FactConsolidation: an earlier
fact is followed by a contradictory update, and answers should reflect the
latest information \citep{memoryagentbench}. This is closely related to
record validity and updating in IMEP. However, evaluating updated answers is
different from comparing same-fact and different-fact replacements at a
specified retrieval interface. ChainBench complements these benchmarks with
a query observation that omits the deciding fact and separate decision and
execution readouts. IMEP specifies comparisons at the retrieval interface, adding paired
outcome-flip accounting and delivery checks to task-level evaluation.

\textbf{Interventions and attribution.} Causal abstraction
\citep{causalabstractions}, causal tracing and activation patching
\citep{rome,zhangnanda}, and causal scrubbing \citep{causalscrubbing} test
behavioral dependence on internal variables. The replacement distribution
limits what such interventions identify \citep{causalscrubbing,zhangnanda}.
Embodied Interpretability connects visual interventions to VLA generalization
\citep{embodiedinterpretability}. Present but Not Remembered examines visual
history in frozen VLAs \citep{presentnotremembered}. Negative controls can
reveal confounding \citep{negativecontrols}, while randomized-parameter and
randomized-label sanity checks test saliency methods' dependence on learned
models and data \citep{sanitychecks}. These checks do not by themselves
establish sensitivity to small memory-content effects. PoPE uses placebo controls to distinguish useful error content from
presentation effects in code-model self-repair \citep{pope}. IMEP brings
these control principles to embodied memory evaluation through paired
success/failure flips and separate decision and execution evidence. Learned-bank studies measure closed-loop outcomes, whereas
ChainBench's shared recorded executions and separate scripted-reader test
support the narrower interpretations stated in the main text.

\textbf{Equivalence and undetected effects.} Nonsignificance does not establish
that an effect is negligible. Equivalence testing instead compares uncertainty
with a specified effect-size interval \citep{lakens2017}. Inference for paired
proportions also depends on the paired design \citep{tango1998}. IMEP
retains study-specific tests and margins, distinguishes original verdicts from
post hoc intervals, and interprets equivalence only for the interventions as
delivered. A narrow interval for one contrast cannot establish complete memory
delivery or rule out an untested mechanism.

\section{Additional Protocol Details and Interpretation}
\label{app:protocol-details}

This appendix details the intervention implementation, validity checks, and
separation of decision and execution readouts used in Section~\ref{sec:method}.

\subsection{Pairing and intervention implementation}
The evaluation unit is a task--seed pair (a scope) or a multi-episode chain.
Comparisons reuse the same units across conditions, but matching seeds alone
does not guarantee identical physical execution. Each study must specify the
memory interface, intervention time, controlled variables, and source of
execution randomness. Blocking reads assesses memory-access utility together
with any computation it changes. Content substitutions instead hold the
retrieval interface fixed at installation. They do not guarantee identical
subsequent computation or semantically identical non-target content.

In cross-episode learned-bank studies, $U$, $S$, and $T$ use an installation
routine that $C$ bypasses. Keys, slots, timestamps, versions, and selected
counts are preserved initially, but subsequent merges can modify the payload.
Thus $U$ versus $C$ does not isolate numerical perturbation alone, and a
same-fact donor is not semantically identical to the recipient. Under Shell
Game blackout, $S_\pi$ permutes and $T_\mu$ resamples retrieved latent values
at each read. These operations preserve the selected record but are not
verified fact-preserving and fact-changing edits. Their contrast measures
resampling versus permutation (Appendix~\ref{app:blackout-ladders}).

\subsection{Readouts and validity checks}
In Eq.~\eqref{eq:decomp}, ten success-to-failure changes and ten
failure-to-success changes among 100 pairs yield $D_X=0$ but
$\rho_X=20\%$. Aggregate success can therefore conceal substantial changes
in individual outcomes. A positive $\theta$ in Eq.~\eqref{eq:discordance}
supports fact specificity only to the extent that replacements control
non-target differences and reach the reader. It concerns the implemented
replacement distributions, not a complete account of internal mechanisms.
For blackout ladders we report $\rho_{T_\mu}-\rho_{S_\pi}$ with the narrower
interpretation above. Decision changes must not be pooled with motor-outcome
changes.

\textbf{Information requirement.} To establish that a task requires history,
the deciding fact must be absent from the current observation, and success
must require one choice among $K$ options, preventing recovery by trying
alternatives. When decisions are scored, we compare no-memory accuracy with
$1/K$ and report $G=\mathrm{Acc}_O-\mathrm{Acc}_N$. An interval containing
chance is a compatibility check, not an equivalence test or proof that model
weights contain no relevant information. The history-required check also
asks whether the lower bound on $G$ exceeds zero. Memory can still help when
partial current-observation cues are available, but that supports a different
claim.

\textbf{Intervention delivery.} Verify scheduled reads in $A$ and empty reads
in $N$. Substitution coverage is the fraction of units whose every decision
query selects at least one unmodified substituted record. Coverage of one
satisfies this operational delivery check without proving that the reader
uses the target fact. Coverage below one limits interpretation to swaps as
delivered. Merged records may still retain substituted fields. Audits are
post hoc unless a survival gate was registered
(Appendix~\ref{app:survival}). Without substitution traces, matching oracle
accuracy is behavioral evidence, not direct verification of the memory path.

\textbf{Record validity.} Vary persistence while holding other components
fixed to test outdated facts. A comparison between whole configurations
cannot attribute an effect to stale information alone.

\textbf{Measurement and decision rules.} Report repeatability controls,
confidence intervals, and study-specific detection limits. Original campaign
verdicts and magnitude criteria remain unchanged. The retrospective Q3 rule
requires the lower bound on $\rho_X$ to exceed the exact-replicate floor;
Q4 requires the lower bound on $\theta$ to exceed both zero and the upper
bound on $\rho_U$, subject to the validity checks. Remember Color's registered
equivalence analysis uses a 90\% interval within $\pm7.5$ percentage points,
following the confidence-interval logic of two one-sided tests
\citep{lakens2017}. This concerns swaps as delivered and does not exclude
smaller effects. Nonsignificance establishes neither equivalence nor absence
of memory use. The framework and its interpretation criteria were formulated
after the campaigns closed (Appendix~\ref{app:amendments}).

\subsection{ChainBench readout and execution}
The 202-chain study compares full memory with no memory, shuffled or stale
records, an initial-fact baseline that ignores updates, and an oracle
decision. Full memory retains the latest write and uses a calibrated presence
gate, two-stage retrieval, and a leave-one-chain-out nearest-centroid readout,
without a decision-time LLM call. The latest-write set is stipulated rather
than produced by an evaluated eviction policy
(Appendix~\ref{app:chainbench}). This evaluates a memory configuration and
readout, not autonomous memory maintenance.

We report decision accuracy $\mathrm{Acc}_X$ and task success
$\mathrm{SR}_X$ separately. Conditions share recorded executions within each
chain. Consequently, making the same correct decisions as the oracle
necessarily yields its observed success rate $\mathrm{SR}^*=\mathrm{SR}_O$,
an execution limit for those recorded trials. The descriptive approximation
$\mathrm{SR}_X\approx\mathrm{Acc}_X\,\mathrm{SR}^*$ additionally assumes
that wrong choices fail and that execution success among a condition's
correct choices is represented by the oracle's overall rate. Agreement within
a success-rate interval is a consistency check, not validation of this
assumption. Repeated physical executions vary, so the $C/U/S/T$ ladder is
evaluated separately on the scripted decision reader. It does not test motor
outcomes or live-harness writing and retrieval
(Appendix~\ref{app:executor-floors}).

\stopcontents[appendix]

\end{document}